\DocumentMetadata{}
\documentclass[manuscript,screen]{acmart}

\makeatletter
\@namedef{r@tocindent4}{0pt}
\@namedef{r@tocindent5}{0pt}
\makeatother

\usepackage{subfigure}
\usepackage{color}
\usepackage{soul}
\usepackage{multicol}
\usepackage{multirow}
\usepackage{tabularx}
\usepackage{booktabs}
\usepackage{xcolor}
\usepackage{xargs}
\usepackage{graphbox}
\usepackage{fontawesome5}
\usepackage{listings}
\usepackage{algorithmic}
\usepackage{tcolorbox}
\usepackage[colorinlistoftodos,prependcaption,textsize=small]{todonotes}
\usepackage{graphicx}
\usepackage{multirow}
\usepackage{longtable}
\usepackage{lscape}
\usepackage{soul}
\usepackage{wrapfig}
\usepackage{caption}
\usepackage{lineno}

\soulregister\cite7
\soulregister\ref7

\definecolor{googler}{rgb}{0.91, 0.25, 0.21}
\definecolor{googleg}{rgb}{0.45, 0.61, 0.12}
\definecolor{googleb}{rgb}{0.26, 0.52, 0.95}
\definecolor{googley}{rgb}{0.98, 0.73, 0.01}

\definecolor{yesColor}{rgb}{0.565, 0.773, 0.482}
\definecolor{noColor}{rgb}{0.796, 0.447, 0.416}
\definecolor{volColor}{rgb}{0.98, 0.73, 0.01}

\newcommandx{\info}[2][1=inline]{\todo[linecolor=googley, backgroundcolor=googley!20, bordercolor=googley, #1, caption={}]{\textcolor{black} {#2} }}
\newcommandx{\help}[2][1=inline]{\todo[linecolor=googleb, backgroundcolor=googleb!20, bordercolor=googleb, #1, caption={}]{\textcolor{black} {#2} }}

\definecolor{code}{rgb}{0.98, 0.98, 0.98}

\colorlet{punct}{red!60!black}
\definecolor{background}{HTML}{EEEEEE}
\definecolor{delim}{RGB}{20,105,176}
\colorlet{numb}{magenta!60!black}

\lstdefinelanguage{json}{
    basicstyle=\scriptsize\ttfamily,
    numbers=none,
    numberstyle=\scriptsize,
    stepnumber=1,
    numbersep=8pt,
    showstringspaces=false,
    breaklines=true,
    frame=lines,
    backgroundcolor=\color{code},
    literate=
     *{0}{{{\color{numb}0}}}{1}
      {1}{{{\color{numb}1}}}{1}
      {2}{{{\color{numb}2}}}{1}
      {3}{{{\color{numb}3}}}{1}
      {4}{{{\color{numb}4}}}{1}
      {5}{{{\color{numb}5}}}{1}
      {6}{{{\color{numb}6}}}{1}
      {7}{{{\color{numb}7}}}{1}
      {8}{{{\color{numb}8}}}{1}
      {9}{{{\color{numb}9}}}{1}
      {:}{{{\color{punct}{:}}}}{1}
      {,}{{{\color{punct}{,}}}}{1}
      {\{}{{{\color{delim}{\{}}}}{1}
      {\}}{{{\color{delim}{\}}}}}{1}
      {[}{{{\color{delim}{[}}}}{1}
      {]}{{{\color{delim}{]}}}}{1},
}

\AtBeginDocument{%
  \providecommand\BibTeX{{%
    \normalfont B\kern-0.5em{\scshape i\kern-0.25em b}\kern-0.8em\TeX}}}

\setcopyright{acmcopyright}
\copyrightyear{2022}
\acmYear{2022}
\acmDOI{XXXXXXX.XXXXXXX}

\acmConference[CSUR '22]{ACM Computing Surveys}{June 03--05, 2022}{Woodstock, NY}
\acmPrice{15.00}
\acmISBN{978-1-4503-XXXX-X/18/06}

\begin{document}

\title{Data Readiness for AI: A 360-Degree Survey}


\author{Kaveen Hiniduma}
\orcid{0009-0006-8516-0215}
\email{hiniduma.1@osu.edu}
\affiliation{%
    \institution{The Ohio State University}
    \city{Columbus}
    \state{Ohio}
    \country{USA}
}
\author{Suren Byna}
\orcid{0000-0003-3048-3448}
\email{byna.1@osu.edu}
\affiliation{%
    \institution{The Ohio State University}
    \city{Columbus}
    \state{Ohio}
    \country{USA}
}

\author{Jean Luca Bez}
\email{jlbez@lbl.gov}
\orcid{0000-0002-3915-1135}
\affiliation{%
    \institution{Lawrence Berkeley National Laboratory}
    \city{Berkeley}
    \state{California}
    \country{USA}
}


\renewcommand{\shortauthors}{Hiniduma et al.}



\begin{abstract}
Artificial Intelligence (AI) applications critically depend on data. Poor quality data produces inaccurate and ineffective AI models that may lead to incorrect or unsafe use. Evaluation of data readiness is a crucial step in improving the quality and appropriateness of data usage for AI. R\&D efforts have been spent on improving data quality. However, standardized metrics for evaluating data readiness for use in AI training are still evolving. In this study, we perform a comprehensive survey of metrics used to verify data readiness for AI training. This survey examines more than $140$ papers published by ACM Digital Library, IEEE Xplore, journals such as Nature, Springer, and Science Direct, and online articles published by prominent AI experts. This survey aims to propose a taxonomy of data readiness for AI (DRAI) metrics for structured and unstructured datasets. We anticipate that this taxonomy will lead to new standards for DRAI metrics that would be used for enhancing the quality, accuracy, and fairness of AI training and inference.     
\end{abstract}






\maketitle

\section{Introduction}
\label{sec:introduction}


Data readiness for artificial intelligence (AI) refers to the critical process of preparing and ensuring the quality, accessibility, and suitability of datasets before using them for AI applications. Readying the data is a fundamental step, which involves collecting, cleaning, organizing, and validating the dataset not only to make them compatible with AI algorithms and models, but also to make certain that the datasets are appropriate and unbiased. By achieving data readiness, organizations can maximize the accuracy, efficiency, and effectiveness of their AI systems, ultimately leading to more informed decision-making and successful AI-driven outcomes. 

Data readiness for AI (DRAI) is an important concern in AI applications, as evidenced by a survey conducted by Scale AI \cite{website:datacenterknowledge}. A significant number of participants encountered challenges related to data readiness within their machine learning (ML) projects. Similarly, a study \cite{website:datanami} involving nearly $2,400$ respondents from over $100$ countries explains the time-intensive nature of data preparation for data scientists working with AI applications. 
It is crucial to recognize that the quality of outcomes generated by an AI system is heavily linked to the readiness of the input data. This connection highlights the significance of addressing the ``garbage in, garbage out'' saying, which emphasizes that flawed or insufficient input data will inevitably lead to inaccurate and unreliable results from AI algorithms \cite{Schmelzer_2019}. Hence, ensuring the availability of well-prepared data for training machine learning (ML) models is critical, as it leads to more precise and dependable predictions.

With growing 
requirements of unbiased data for AI,
the field of quantitative evaluation of data readiness with appropriate metrics is still evolving. 
Data Quality Toolkit (DQT) \cite{IBM_DQT} 
provides 
a suite of tools and functionalities to streamline the preparation and cleaning of data. 
DQT's data quality report 
various dimensions of data quality metrics defined by Sidi et al. \cite{6204995} such as completeness, consistency, accuracy, and timeliness, along with suggestions for improvement. 
Ravi et al. \cite{Ravi_2022} focus on the critical process of making experimental datasets FAIR (Findable, Accessible, Interoperable, Reusable) for AI readiness. 
They emphasize the usage of current data infrastructure to establish a framework suitable for automatic AI-powered exploration. To achieve this objective, they publish FAIR and AI-ready datasets
\cite{blaiszik2016materialdatafacility}. 
This study also illustrates usage of FAIR principle compliance \cite{FAIR:Principles} and AI-ready datasets 
for inference.

Although these separate efforts and tools 
are available to improve quality of data, there is a lack of a comprehensive study on effective metrics and standards for evaluating data readiness for AI.
To address that gap, we perform a comprehensive examination of the existing metrics and tools that could be used for evaluating data readiness, covering both structured and unstructured data dimensions
We also describe the metrics designed to evaluate fairness and privacy related issues in data, which critically 
impact the process of decision-making in AI algorithms. 

This survey refers to a comprehensive set of data readiness dimensions and data preparation techniques targeting data usage in AI.
Metrics targeting data readiness for AI (DRAI) contain a subset of data quality dimensions including completeness, duplicates, correctness, and timeliness. This distinction between DRAI and data quality is critical for understanding our survey's scope. We identify 
the existing metrics and scoring mechanisms (\S\ref{sec:drai_metrics}) and existing frameworks or tools (\S\ref{sec:frameworks}) in the literature that could be used to measure DRAI. Based on the distillation of available literature, we propose a potential comprehensive definition of data readiness for AI using six dimensions (\S\ref{sec:definition}). We discuss gaps and challenges towards developing a DRAI assessment framework (\S\ref{sec:gaps-challenges}).
This survey is particularly aimed at data preparers for future AI use and data scientists who analyze the data
to ensure their datasets are ready for AI applications. 

\vspace{-5pt}

\section{Scope of the Study}
\label{sec:scope}

Our literature review methodology was carefully structured to ensure a comprehensive and unbiased examination of existing research. The search queries used to obtain the sources for this study are presented in Table \ref{tab:search_queries}, categorized into general, structured data, and unstructured data search queries. As a result, we gathered nearly $30$ papers from ACM Digital Library, over $20$ papers from IEEE Xplore, $10$ papers each from Springer and Science Direct, and more than $40$ papers from journals including Nature, Springer and Science Direct, as well as several relevant books, to review. Additionally, we highlight discussions on six web articles and explore the metrics used in six commercially used tools.

\begin{table}[ht]
\centering
\caption{Data Readiness for AI Metrics: Summary of search terms used to identify literature to perform this study}
\label{tab:search_queries}
\vspace{-10px}
\setlength\extrarowheight{2pt}
\resizebox{\textwidth}{!}{%
\begin{tabular}{|ll|l|l|}
\hline
\multicolumn{2}{|c|}{\textbf{General}} &
  \multicolumn{1}{c|}{\textbf{Structured Data Related}} &
  \multicolumn{1}{c|}{\textbf{Unstructured Data Related}} \\ \hline
\multicolumn{1}{|l|}{``data readiness'' AND ``AI''} &
  ``data quality'' AND ``assessment'' &
  \begin{tabular}[c]{@{}l@{}}searched under each data readiness for AI dimension \\ e.g., ``discriminat*'' AND ``metric'' OR ``measure''   OR ``evaluat*''\end{tabular} &
  ``speech quality'' AND ``metric'' OR ``measure'' OR ``evaluat*'' \\ \hline
\multicolumn{1}{|l|}{``data readiness''} &
  ``data quality dimension'' &
   &
  ``audio quality'' AND ``metric'' OR ``measure'' OR ``evaluat*'' \\ \hline
\multicolumn{1}{|l|}{``data readiness'' AND ``machine learning'' OR ``ML''} &
  ``data quality'' AND ``metric'' &
   &
  ``video quality'' AND ``metric'' OR ``measure'' OR ``evaluat*'' \\ \hline
\multicolumn{1}{|l|}{"AI ready"} &
  ``data prepare'' AND ``AI'' &
   &
  ``image quality'' AND ``metric'' OR ``measure'' OR ``evaluat*'' \\ \hline
\multicolumn{1}{|l|}{``data quality'' AND ``machine learning''} &
  ``data read*'' AND ``metric'' &
   &
  ``visual quality'' AND ``metric'' OR ``measure'' OR ``evaluat*'' \\ \hline
\multicolumn{1}{|l|}{``data quality'' AND ``measure''} &
  ``data preprocess'' &
   &
   \\ \hline
\multicolumn{1}{|l|}{``data quality'' AND ``evaluation''} &
  ``data clean'' &
   &
   \\ \hline
\multicolumn{1}{|l|}{``data quality'' and ``AI''} &
  ``data quality'' AND ``survey'' &
   &
   \\ \hline
\end{tabular}%
}
\end{table}

Papers and articles were included if they discussed data readiness metrics for AI or data quality metrics, covered structured or unstructured data dimensions, and addressed fairness and privacy issues related to DRAI. We included sources that provided metrics or tools for evaluating data readiness and those that were peer-reviewed or published in reputable journals. We also used web articles that provided valuable insights and were considered highly relevant to our study. Additionally, if the metrics were related to AI data preprocessing, such as feature relevancy, class imbalance measures, and FAIR compliance, they were also included.

We excluded papers that did not focus on DRAI, those that were not peer-reviewed, and articles that lacked substantial contributions to understanding DRAI metrics. We also excluded duplicates and studies not available in English. Additionally, we excluded papers if the metrics were not quantifiable or unrelated to the pre-data training stage in AI.

Our survey aims to identify and analyze the existing metrics and scoring mechanisms for measuring DRAI, covering both conventional dimensions of data quality (e.g., completeness, outliers, timeliness, and correctness) and AI-specific dimensions (e.g., fairness, feature importance, class imbalance, and mislabeled data). We seek to provide insights into how these dimensions and metrics can be used to assess the preparedness of data for AI applications.

\begin{wrapfigure}{r}{0.50\textwidth}
  \centering
  \includegraphics[width=0.50\textwidth]{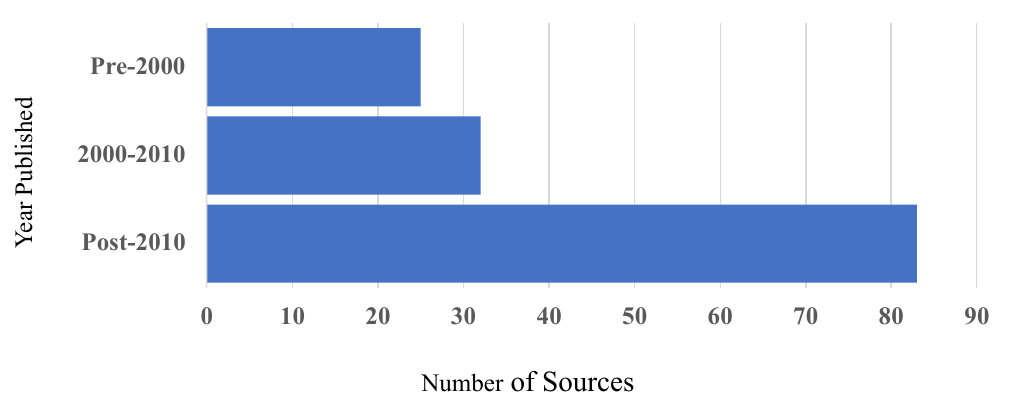}
  \caption{Papers chosen for this survey from different time frames.
  }
  \label{fig:distribution of sources}
  \vspace{-10pt}
\end{wrapfigure}

In forming this study, our focus included published literature across different timeframes -- pre-2000, 2000-2010, and post-2010. In Fig. \ref{fig:distribution of sources}, we show the distribution of sources across different time frames. Paying attention to references prior to 2000 is particularly important because they include some well-established data quality metrics that remain relevant today. These early efforts provide the context and insights into an evolution of data readiness metrics over the years. Before 2000, work primarily focused on general data quality considerations without specific emphasis on applications in AI. From 2000 to 2010, AI research was focused on foundational computational methods and concepts. Post-2010, with the rise of big data and machine learning technologies, a notable shift toward creating metrics to assess data preparedness emerged, specifically for AI applications.

\vspace{-10pt}
\subsection{Existing Surveys and Gaps}


Existing surveys on data quality (\cite{batini2006data,4016511,6204995,5412098,LIN2011297}) primarily focused on traditional dimensions like completeness, correctness, timeliness, and also the quality of textual and multimedia data, providing a strong foundation for understanding the general challenges in the field. However, with the rise of AI, there has been an increasing emphasis on AI-specific concerns like feature relevance, class imbalance, mislabeled data, privacy, and fairness (\cite{10.1145/3168389,ntoutsi2020bias,10.1145/3136625,article:priestley,10.1145/3588433}). These emerging factors are critical alongside conventional quality benchmarks. Recognizing the importance of both aspects, our survey aims to address this shift towards a comprehensive view of data readiness, where traditional and AI-focused dimensions are equally essential.

Toward gaining an understanding of metrics for data readiness, a few studies surveyed data pre-processing stages in preparing data for AI. \citet{article:priestley} conducted a study highlighting the role of decision-makers and practitioners in improving data-focused practices. They highlighted the significance of data cleaning and pre-processing stages, including feature selection, duplicate elimination, outlier removal, consistency assurance, and handling of missing values. Documentation of these pre-processing steps was essential to ensure compatibility and identify potential dependencies and information leakages among features. Their insights, derived from an extensive literature survey, laid the foundation for recognizing the pivotal role of data readiness in AI applications. 
\citet{9678209} presented a survey paper on data quality evaluation and enhancement. They compared existing frameworks, models, and methods for data quality evaluation and enhancement, and identified challenges. 


Batini and Scannapieco's \cite{batini2006data} book offers a comprehensive introduction to a broad range of data quality issues. This book provides a state of the art overview of data quality measurement practices by using probability theory, data mining, statistical data analysis, and machine learning.

A few studies have looked into metrics for unstructured data. \citet{4016511} explored similarity metrics for duplicate record detection, specifically focusing on challenges posed by typographical variations in string data. \citet{10.1145/3136625} concentrated on feature selection metrics, categorizing traditional approaches into wrapper, filter, embedded, and hybrid methods. Furthermore, \citet{10.5555/944919.944974} explored feature selection metrics for text classification, examining their performance. 
The studies on image quality measures \cite{5412098} and perceptual visual quality metrics \cite{LIN2011297} are important in understanding DRAI for unstructured data, particularly in visual data, where assessment of image quality and perceptual metrics are essential for effective AI applications.

\begin{wrapfigure}{r}{0.55\textwidth}
  \centering
  \includegraphics[width=0.55\textwidth]{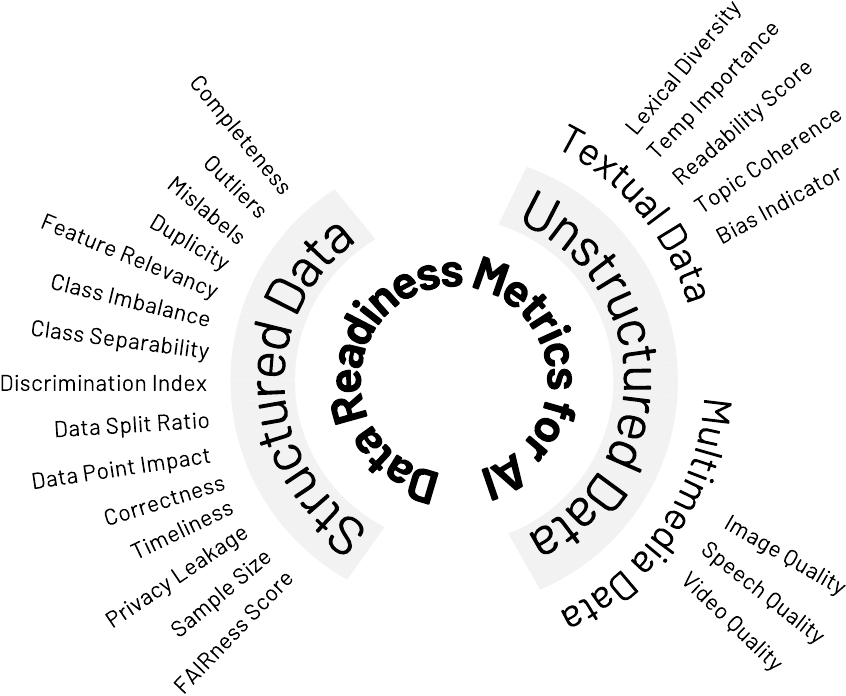}
  \caption{360\textdegree~View of Mapping Data Readiness Dimensions for AI}
  \label{fig:360_plot}
\end{wrapfigure}

In the context of bias and fairness, \citet{10.1145/3588433} provided a survey of techniques focused on identifying and mitigating representation bias across diverse data types, such as structured data, image data, textual data, and graph data. They defined the problem and discussed causes and methods for measuring and quantifying representation bias in structured and unstructured datasets. Addressing the issue of discrimination in DRAI systems, \citet{ntoutsi2020bias} focused on the challenges and implications of biased data on the fairness and accuracy of AI-based decision-making. They emphasized the importance of understanding and mitigating biases in data to prevent the continuation of discriminatory practices. Similarly, in the context of privacy, \citet{10.1145/3168389} reviewed over $80$ privacy metrics across six domains, categorizing them based on the aspect of privacy measured, required inputs, and data types needing protection. They identified research gaps in areas such as metric combination and interdependent privacy, proposing a method for selecting appropriate metrics through nine key questions. Emphasizing the importance of employing multiple metrics to address diverse privacy aspects, the paper serves as a reference guide and toolbox for privacy researchers, aiding informed choices in metric selection for specific scenarios.

Our study aims to contribute to the evolving field of DRAI metrics, specifically focusing on structured and unstructured data metrics. While previous works have explored specific dimensions or concentrated on individual metrics, we address the lack of a comprehensive study including numerous DRAI dimensions. There are broader perspectives in addressing DRAI applications. We identify numerous important factors that define data readiness, such as data preparation, privacy leakage evaluation, data discrimination evaluation, compliance to FAIR principles, mislabeled data detection, feature relevancy analysis, bias-related issues, and quality evaluation of speech and multimedia data. 
To survey existing literature in these dimensions and to identify gaps,
our effort aims to advance the field of data readiness for AI (DRAI) and inform best practices for evaluating the suitability of data for AI applications. Given the growing importance of DRAI, our work fills a critical gap in the published literature and gives insights to practitioners and decision-makers in the field. As illustrated in Fig. ~\ref{fig:360_plot}, the 360\textdegree~plot view of data readiness incorporates a range of metrics that reflect the ongoing discussion in the literature for both structured and unstructured data. Additionally, we also define DRAI by introducing key pillars and the DRAI metrics for each pillar.

\vspace{-10pt}




\section{DRAI Metrics}
\label{sec:drai_metrics}
\setcounter{secnumdepth}{4}

This section provides an extensive summary of the existing metrics found in the literature that are used to measure data readiness for AI. We will explore these metrics for both structured and unstructured data, primarily focusing on structured data for which metrics have evolved more. Nevertheless, we describe many metrics related to assessing readiness of unstructured data, including textual, multimedia, image, speech, and video-related data. 
In Table \ref{tab:dim_cite}, we provide a snapshot of all the dimensions and metrics discussed in this survey.

\vspace{-10pt}

\begin{table}[b]
\centering
\vspace{-10px}
\caption{Dimensions and Metrics for (Structured and Unstructured) DRAI}
\label{tab:dim_cite}
\vspace{-10px}
\setlength\extrarowheight{2pt}
\resizebox{\textwidth}{!}{%
\begin{large}
\begin{tabular}{|cl|cll|}
\hline
\multicolumn{2}{|c|}{\textbf{Structured Data}} &
  \multicolumn{3}{c|}{\textbf{Unstructured Data}} \\ \hline
\multicolumn{1}{|c|}{\textbf{Dimension}} &
  \multicolumn{1}{c|}{\textbf{Metrics}} &
  \multicolumn{1}{c|}{\textbf{Data Type}} &
  \multicolumn{1}{c|}{\textbf{Dimension}} &
  \multicolumn{1}{c|}{\textbf{Metrics}} \\ \hline
\multicolumn{1}{|l|}{\textbf{Completeness}} &
  \begin{tabular}[c]{@{}l@{}}Blake et al. \cite{10.1145/1891879.1891881}, Bors et al. \cite{10.1145/3190578}, \\ Santos et al. \cite{SANTOS2020111}, Pearson \cite{10.1145/1147234.1147247}\end{tabular} &
  \multicolumn{1}{c|}{\multirow{19}{*}{\textbf{Textual Data}}} &
  \multicolumn{1}{l|}{\textbf{Lexical Diversity}} &
  \begin{tabular}[c]{@{}l@{}}Templin \cite{templin1957certain}, McCarthy et al. \cite{mccarthy2005assessment}, \\McCarthy et al.\cite{mccarthy2010mtld}\end{tabular} \\ \cline{1-2} \cline{4-5} 
\multicolumn{1}{|l|}{\textbf{Outliers}} &
  \begin{tabular}[c]{@{}l@{}}Bors et al. \cite{10.1145/3190578}, Li et al. \cite{9458702}, \\ Breunig et al. \cite{breunig2000lof}, Pokraja et al. \cite{pokrajac2007incremental}, \\ Rosner et al. \cite{rosner1983percentage}, Leys et al. \cite{leys2013detecting}, \\ Rousseeuw et al. \cite{rousseeuw2018anomaly}, Degirmenci et al. \cite{9627977}\end{tabular} &
  \multicolumn{1}{c|}{} &
  \multicolumn{1}{l|}{\textbf{Term Importance}} &
  \begin{tabular}[c]{@{}l@{}}Luhn \cite{5392697}, \\ Sparck Jones \cite{sparck1972statistical}\end{tabular} \\ \cline{1-2} \cline{4-5} 
\multicolumn{1}{|l|}{\textbf{Mislabels}} &
  Gupta et al’s \cite{gupta2021data}, Cohen's Kappa \cite{cohen_coefficient_1960} &
  \multicolumn{1}{c|}{} &
  \multicolumn{1}{l|}{\textbf{Readability Score}} &
  \begin{tabular}[c]{@{}l@{}}Rudolf Flesch \cite{flesch1986art}, \\ Coleman and Liau \cite{Coleman1975ACR}, \\ Robert Gunning Associates \cite{readable_gf},\end{tabular} \\ \cline{1-2} \cline{4-5} 
\multicolumn{1}{|l|}{\textbf{Duplicate Values}} &
  \begin{tabular}[c]{@{}l@{}}Bors et al. \cite{10.1145/3190578}, Levenshtein distance metric \cite{Levenshtein1965}, \\ Waterman et al. \cite{Waterman1976}, Jaro's distance metric \cite{jaro1976unimatch}, \\ Monge et al. \cite{monge1996fieldmatching}, Russell et al. \cite{russell1922index}\end{tabular} &
  \multicolumn{1}{c|}{} &
  \multicolumn{1}{l|}{\textbf{Topic Coherence}} &
  \begin{tabular}[c]{@{}l@{}}R\"{o}der et al. \cite{10.1145/2684822.2685324}, \\ Mimno et al. \cite{10.5555/2145432.2145462}, \\ Newman et al. \cite{10.1145/1816123.1816156}\end{tabular} \\ \cline{1-2} \cline{4-5} 
\multicolumn{1}{|l|}{\textbf{Feature Relevancy}} &
  \begin{tabular}[c]{@{}l@{}}Dai et al. \cite{4053044}, He et al. \cite{he2005laplacian}, Zhao et al. \cite{zhao2007spectral}, \\ Duda et al. \cite{duda2012pattern}, Nie et al. \cite{nie2008trace}, Lewis \cite{lewis1992feature}, \\ Robnik-Sikonja et al. \cite{robnik2003theoretical}, Davis et al. \cite{davis1986statistics}, \\ Liu et al. \cite{liu1995chi2}, Gini \cite{gini1912variability}, \\ Hall et al. \cite{hall1999feature}\end{tabular} &
  \multicolumn{1}{c|}{} &
  \multicolumn{1}{l|}{\textbf{Bias Indicator}} &
  Papakyriakopoulos et al. \cite{10.1145/3351095.3372843}, \\ \hline
\multicolumn{1}{|l|}{\textbf{Class Imbalance}} &
  \begin{tabular}[c]{@{}l@{}}Lu et al. \cite{lu2019bayes}, Francisco et al. \cite{alberto2018learning}, \\ Ortigosa-Hernández et al. \cite{ORTIGOSAHERNANDEZ201732}, Zhu et al. \cite{ZHU201836}, \\ Gupta et al. \cite{gupta2021data}\end{tabular} &
  \multicolumn{1}{c|}{\multirow{9}{*}{\textbf{Multimedia Data}}} &
  \multicolumn{1}{l|}{\textbf{Image Quality}} &
  \begin{tabular}[c]{@{}l@{}}MSE and PSNR \cite{psnr}, Wang et al. \cite{995823}, \\ Wang et al. \cite{1284395}, Wang et al. \cite{1292216}, \\ Sarnoff's JND-Metrix \cite{sarnoff_jndmetrix}, Sheikh et al. \cite{1576816}, \\ Chandler et al. \cite{4286985}, Lakhani’s \cite{lakhani2020importance}, \\ Sabottke et al’s. \cite{sabottke2020effect}, Lin et al. \cite{lin2005visual}, \\ Marziliano et al. \cite{marziliano2002no}, PIQ \cite{piq}\end{tabular} \\ \cline{1-2} \cline{4-5} 
\multicolumn{1}{|l|}{\textbf{Class Separability}} &
  \begin{tabular}[c]{@{}l@{}}Gupta et al’s \cite{gupta2021data}, \\ Sejong \cite{OH2011115}, \\ Borsos et al. \cite{borsos2018dealing}\end{tabular} &
  \multicolumn{1}{c|}{} &
  \multicolumn{1}{l|}{\textbf{Speech Quality}} &
  \begin{tabular}[c]{@{}l@{}}Mean Opinion Score \cite{itu_p808}, Rix et al. \cite{941023}, \\ Jayant et al. \cite{jayant1984digital}, Taal et al. \cite{5495701}, \\ Beerends et al. \cite{beerends1994perceptual}, Itakura-Saito Spectral Distortion \cite{itakura1968analysis}, \\ Objective Difference Grade \cite{itu_bs1387},\end{tabular} \\ \cline{1-2} \cline{4-5} 
\multicolumn{1}{|l|}{\textbf{Discrimination Index}} &
  \begin{tabular}[c]{@{}l@{}}Azzalini et al. \cite{10.1145/3552433}, Feldman et al. \cite{10.1145/2783258.2783311}, \\ Celis et al. \cite{celis2020data}, Simonetta et al. \cite{simonetta2021metrics}, \\ Gupta et al. \cite{gupta2021data},Amazon SageMaker Developer Guide \cite{Kemka_2019}\end{tabular} &
  \multicolumn{1}{c|}{} &
  \multicolumn{1}{l|}{\textbf{Video Quality}} &
  \begin{tabular}[c]{@{}l@{}}PSNR \cite{psnr}, Wang et al. \cite{1284395}, Sheikh et al. \cite{1576816}, \\ Chandler et al. \cite{4286985}, Huynh-Thu et al. \cite{Huynh-Thu}, \\  Netflix \cite{Blog_2017}, OPTICOM’s PEVQ \cite{PEVQ}\end{tabular} \\ \hline
\multicolumn{1}{|l|}{\textbf{Data Split Ratio}} &
  Joseph \cite{roshan2022optimal}, Affendras et al. \cite{AFENDRAS2019286}&
   \\ \cline{1-2} 
\multicolumn{1}{|l|}{\textbf{Data Point Impact}} &
  \begin{tabular}[c]{@{}l@{}}Ghorbani et al. \cite{ghorbani2019data}, Wang et al. \cite{wang2023data},\\ Leave-One-Out \cite{cook1982residuals}, Koh et al. \cite{koh2017influence},\\ Bachem et al. \cite{bachem2017coreset}, Ribeiro et al. \cite{ribeiro2016lime}\end{tabular} &
   \\ \cline{1-2} 
\multicolumn{1}{|l|}{\textbf{Correctness}} &
  Kaiser et al. \cite{Kaiser_Klier_Heinrich_1970}, Pipino et al's\cite{10.1145/505248.506010}&
   \\ \cline{1-2} 
\multicolumn{1}{|l|}{\textbf{Timeliness}} &
  \begin{tabular}[c]{@{}l@{}}Kaiser et al. \cite{Kaiser_Klier_Heinrich_1970}, Heinrich et al. \cite{HEINRICH201582}, \\ Blake et al. \cite{10.1145/1891879.1891881}\end{tabular} &
   \\ \cline{1-2} 
\multicolumn{1}{|l|}{\textbf{Privacy Leakage}} &
  \begin{tabular}[c]{@{}l@{}}Vatsalan et al. \cite{https://doi.org/10.1111/bjet.13223}, Duddu et al. \cite{duddu2022shapr},\\Carlini et al. \cite{carlini2022privacy},Song et al. \cite{272134},\\Bezzi \cite{4550303}, Longpr{\ 'e} et al. \cite{Longpr2017EntropyAA},\\Sevgi et al. \cite{9458108}, Aindo AI \cite{aindoprivacy}\end{tabular} &
   \\ \cline{1-2} 
\multicolumn{1}{|l|}{\textbf{Sample Size}} &
  \begin{tabular}[c]{@{}l@{}}Alwosheel et al. \cite{ALWOSHEEL2018167}, Haykin \cite{haykin2009neural}\end{tabular} &
   \\ \cline{1-2} 
\multicolumn{1}{|l|}{\textbf{FAIR Compliance}} &
  Wilkinson et al. \cite{Wilkinson_Sansone_Schultes_Doorn_Bonino_da_Silva_Santos_Dumontier_2018}, Clarke et al. \cite{clarke2019fairshake} &
   \\ \cline{1-2} 
\end{tabular}%
\end{large}
}

\end{table}

\subsection{Structured data}
\label{subsec:structured_data}


Structured data is organized with a consistent format and follows a specific structure or schema. Data stored in spreadsheets, relational database tables, self-describing file formats, etc., are common forms of structured data. This section will discuss metrics related to various dimensions, such as completeness, outliers, labels, etc., as shown on the left half of Figure \ref{fig:360_plot} and their sources listed in the left half of Table \ref{tab:dim_cite}.

\vspace{-10pt}
\subsubsection{Completeness}
\label{subsubsec:completeness}

Completeness refers to the presence or availability of required data and attribute values in a dataset. It indicates whether data points or entries are complete, with all relevant attribute values recorded and available.

\noindent\textbf{Example:} In a dataset containing information about credit card customer demographics, this metric verifies if an attribute (e.g., income) is available for all customers.
Completeness ensures that a dataset is reliable and suitable for analysis, as there is no loss of information due to missing data. 

\noindent\textbf{Metrics in Literature:} \citet{10.1145/1891879.1891881} propose a ``completeness'' metric to measure the presence of missing values in a dataset. This metric quantifies the proportion of null (missing) data records to the total number of data records. 

Using this metric, researchers demonstrate how missing data can impact the results of classification tasks. For example, \citet{jaeger2021benchmark} demonstrates that handling missing values enhances predictive model performance. They observe up to $20\%$ improvement for classification tasks and $15\%$ improvement for regression tasks, emphasizing the importance of addressing missing data in optimizing downstream ML outcomes.
In addressing missing data, \citet{SANTOS2020111} use data imputation, which involves filling in or estimating missing values to maintain data integrity. In particular, the authors use the k-nearest neighbors (KNN) imputation technique in this process. They discuss the significance of choosing appropriate distance metrics, such as Heterogeneous Value Difference Metric (HVDM) and Heterogeneous Euclidean-Overlap Metri (HEOM), which effectively handle both nominal and continuous data while preserving data distribution during imputation. 

\citet{10.1145/3190578} propose a different approach to quantify missing values in a dataset by using indicators to distinguish missing from non-missing values. Their method offers a practical tool for data preparation, allowing easy identification of missing data in AI applications. 

Another type of completeness targets missing data ``disguised'' with default values. 
Pearson \cite{10.1145/1147234.1147247} discusses missing values are encoded or represented in ways that obscure their true nature, such as using ``zero'' values to indicate missing data. Such practices can severely distort analysis results. 
According to \citet{vo2024explainabilitymachinelearningmodels} these issues affect not only the model's predictions but also its explainability, potentially skewing feature importance calculations crucial for interpreting complex AI systems.


\noindent\textbf{Impact on AI:} Complete and accurate data enhances AI systems' accuracy and reliability. 
Identifying explicitly missing data is easy, while disguised missing values pose a greater challenge as they appear valid but are placeholders or incorrect entries. Disguised values can lead to biased outcomes, reduced accuracy, and misinterpretation of AI models. 

\noindent\textbf{Summary:} \emph{Metrics proposed by various researchers quantify the impact of missing values and suggest remedies such as KNN imputation with suitable evaluation metrics. Additionally, using indicators for distinction and incorporating completeness metrics that consider data types and relationships further improve the handling of missing data. 
} 

\vspace{-8pt}
\subsubsection{Outliers}
\label{subsubsection:outliers}

Outliers in a dataset refer to data points that significantly deviate from the typical or expected values within the dataset. They are points that are significantly distant from the majority of the data points and do not follow the general patterns present in the dataset.

\noindent\textbf{Example:} Consider a dataset of housing prices based on factors such as size, number of bedrooms and bathrooms, location, schools, etc. In this dataset, an outlier could be a property with extremely high or low prices that do not align with the average price range of similar properties. This outlier might be an exceptional case, such as a luxury mansion in an otherwise average neighborhood, or it could be a data entry error.

\noindent\textbf{Metrics in Literature:} In their research, \citet{10.1145/3190578} discuss the concept of plausibility as a metric to identify outliers in datasets for AI applications, which can disrupt statistical analyses and modeling. Data analysts employ two main approaches to quantify the number of outliers: robust statistics and non-robust statistics. Robust statistics use the median and the robust interquartile range estimator, which are more resistant to outliers. \citet{9458702} further explore the standard deviation and interquartile range-based outlier detection methods in their study. In contrast, non-robust statistics involve using the mean and standard deviation to identify entries that deviate significantly from the mean.  

In contrast, \citet{breunig2000lof} introduce the Local Outlier Factor (LOF) as a metric for identifying outliers in a dataset. LOF quantifies the level of being an outlier for each data instance by considering the density of the dataset's distribution. Outliers are expected to have lower local densities compared to their surrounding instances. The LOF algorithm computes the LOF value for a specific instance by comparing the density of that instance's neighborhood with its neighboring instances. This neighborhood is a user-defined parameter (e.g., number of nearest neighbors). Higher LOF values indicate a higher probability of an instance being an outlier, which implies a notably lower density in its local neighborhood than in neighboring data points. 
Building on this foundation, \citet{pokrajac2007incremental} introduced the ILOF (Incremental Local Outlier Factor) method, which uses the LOF metric to determine if a new data point is an outlier. By analyzing the computed LOF value, the ILOF method assigns a score to the incoming sample, indicating whether it is classified as an outlier or not. This approach allows for real-time outlier detection and updates the scores of existing points to measure the impact of the new data point.

\citet{9627977} introduce RiLOF, which addresses limitations in existing statistical outlier detection techniques by introducing the MoNNAD (Median of Nearest Neighborhood Absolute Deviation) metric. This metric is calculated as the median of the absolute variances among the LOF values of the $k$-nearest neighboring data points and the LOF value of a given sample. This score indicates how much the sample deviates from its local neighborhood. In the RiLOF method, the MoNNAD score is used to label and score query samples. Samples are categorized as outliers when their MoNNAD scores are equal to or greater than a specific limit. 
The RiLOF method assigns more importance to the query sample, resulting in clearer differentiation between inliers and outliers. The study demonstrates that the MoNNAD metric, incorporated in the RiLOF method, successfully detects outliers, including outlier clusters, that other techniques fail to recognize. 

The GESD (Generalized Extreme Studentized Deviation) technique, as introduced by \citet{rosner1983percentage}, and the MAD (Median Absolute Deviation) technique, proposed by \citet{leys2013detecting}, are both outlier detection methods that share similarities in their underlying principles. Both approaches aim to identify outliers within datasets by using statistical measures to assess the deviation of data points from central tendencies. GESD identifies outliers by evaluating the maximum absolute difference between each sample and the dataset's mean and normalizing it by the standard deviation. MAD identifies outliers by considering the median of absolute differences between data records and the dataset's median. It incorporates a constant associated by assuming that data is normally distributed. Additionally, the Z-score method aligns with this principle by normalizing sample values using the mean and standard deviation or MAD. The robust Z-score version, introduced by \citet{rousseeuw2018anomaly}, substitutes the median and MAD for more robust measures, demonstrating the shared concept of using statistical measures to detect outliers in datasets.


\noindent\textbf{Impact on AI:} Outliers can significantly impact AI systems by skewing data distributions and introducing biases that lead to inaccurate models and unreliable predictions. They can distort statistical measures such as mean and variance, affecting algorithms like linear regression and clustering. They also can reduce the generalization and robustness of classification and neural network models. Outliers can also introduce noise, complicating the ML process and potentially leading to false positives or negatives \cite{infolabs2024outlier}. However, in specific contexts like fraud detection or rare disease diagnosis (e.g., \citet{markham2024ai}), outliers can be critical for identifying anomalies and should not be removed without a thorough analysis. Proper detection and management of outliers maintain the integrity and effectiveness of AI models.

\noindent\textbf{Summary:} \emph{Outliers data points deviate significantly from most of the dataset and do not follow the general patterns. Different metrics and techniques proposed to identify and measure outliers in datasets include measures based on column heterogeneity, statistical measures like median, standard deviation, mean, and interquartile range, and techniques such as Local Outlier Factor (LOF), Generalized Extreme Studentized Deviation (GESD), Z-score.  
}

\vspace{-10pt}
\subsubsection{Mislabeled Data}
\label{subsubsection:mislabels}

Mislabeled data in the context of preparing a dataset for AI refers to instances or data points with inaccurate labels. It represents a form of labeling error or inconsistency within the dataset, where the assigned labels do not align with the true or expected labels.

\noindent\textbf{Example:} Consider a dataset for email spam classification, where each email is labeled as either ``spam'' or ``not spam''. If some emails are mistakenly labeled as ``not spam'' when they should have been labeled as ``spam'', or vice versa, it introduces mislabeled data. In this case, the mislabeled instances create discrepancies between the assigned labels and the actual content or characteristics of the emails.

\noindent\textbf{Metrics in Literature:} Gupta et al.'s \cite{gupta2021data} Data Quality Toolkit (DQT) introduces a label purity metric to measure the impact of adding random noise on the performance of a classifier trained on the dataset. In the example provided in the study, $10\%$ random noise is introduced to $41$ datasets from UCI (\citet{dua2017uci}) and Kaggle (\cite{Kaggle}) repositories, and the performance of an AutoAI classifier (\citet{liu2020admm}) is measured using 3-fold cross-validation. The results show a drop in classifier performance after inducing noise, with varying degrees of decrease observed across the datasets. 

In evaluating the accuracy of labels assigned by multiple annotators, Cohen's Kappa \cite{cohen_coefficient_1960} is widely employed for assessing inter-rater reliability, especially with categorical or binary labels. Cohen's Kappa calculates the agreement beyond chance, ranging from $-1$ to $1$, where values near $1$ indicate substantial agreement. \citet{lavitas2021annotation} contribute a credibility metric, assessing the likelihood of correct annotations based on multiple reviewers' agreement. The metric ranges from $(N/2 + 1)/N$ to $1$, reflecting high credibility with close agreement and lower credibility with less agreement among reviewers, offering insights into the reliability of the annotation process involving multiple reviewers

\noindent\textbf{Impact on AI: }Incorrect labels in training data lead to poor model performance because the AI learns from erroneous examples, resulting in skewed predictions and decisions \cite{cleanlab2024elevating}. This issue can cause substantial financial costs due to the need to retrain models and correct errors. Additionally, mislabeled data can undermine trust in AI systems, as stakeholders lose confidence in their outputs. Ethical implications are also significant, as mislabeled data can introduce or amplify biases that lead to discriminatory outcomes, such as flawed medical diagnoses or biased hiring algorithms. Addressing mislabeled data is crucial for maintaining the integrity and effectiveness of AI models.

\noindent\textbf{Summary: } \emph{Mislabeled data refers to instances in a dataset that have inaccurate labels, creating discrepancies between assigned labels and the true or expected labels. Available approaches include a label purity metric for classifying performance under induced noise, label agreement among annotators, and credibility metric through reviewer consensus.
}

\vspace{-7pt}
\subsubsection{Duplicate Values}
\label{subsubsection:duplicity}
This refers to the presence of duplicate or redundant records within a dataset. Duplicates appear when the same or similar data entries appear multiple times, potentially skewing the ML  process.

\textbf{Example:} Consider a dataset of customer transactions where each entry represents a purchase made by a customer. Multiple entries of the same transaction refer to duplication, which may impact the analysis results. 

\textbf{Metrics in Literature:} \citet{10.1145/3190578} propose a mechanism to identify duplicate entries in a dataset using a scoring system based on uniqueness. A user selects one or more columns in the dataset that are intended to have distinctive combinations of values. The system assigns a score of $1$ (true) to values with a unique combination in the selected columns and a $0$ (false) score to values found multiple times. By incorporating this score, the system effectively flags duplicate entries in the dataset. 
\citet{4016511} conducted a comprehensive survey exploring various similarity metrics for duplicate detection, addressing challenges in managing typographical variations in string data. One of the highlighted character-based similarity metrics in the survey is the Levenshtein distance metric \cite{Levenshtein1965}. This metric measures the number of operations needed to transform one string into another through edit operations (insertion, deletion, and character replacement). 
\citet{Waterman1976} introduced the Affine Gap Distance metric to overcome the limitations of the standard edit distance metric in matching shortened or truncated strings. It introduces two edit operations: open and extend the gap. An open gap in sequence alignment indicates the start of missing or deleted characters in the sequence. In contrast, an extended gap accommodates consecutive missing characters by extending an existing one. This metric allows for smaller penalties for gap mismatches, resulting in more accurate measurements for truncated or shortened strings.
Additionally, Jaro's distance metric \cite{jaro1976unimatch} quantifies the similarity between two strings by identifying common characters that appear at the same positions in both strings and adding up the number of transpositions. The Jaro metric considers the number of shared characters, the lengths of the strings, and the number of transpositions. 
\citet{monge1996fieldmatching} introduced a token-based similarity metric designed to detect duplicates in text fields using atomic strings. Atomic strings are identified by punctuation characters, acting as delimiters, and consist of alphanumeric characters as individual units within the text fields. Two atomic strings are considered duplicates if they are either identical or if one is a prefix of the other. 
This approach helps identify duplicates in text fields by considering matching atomic strings and provides a similarity score to assess the degree of duplication between fields.

The Soundex algorithm, introduced by \citet{russell1922index}, is a phonetic coding scheme used to detect duplicates by comparing the phonetic similarity of character strings, such as names. The algorithm transforms names into codes based on rules of phonetic similarity. It preserves the initial letter of the name as the prefix letter and assigns codes to each remaining letter according to specific phonetic groups. Vowels act as separators between consecutive consonants. Consecutive occurrences of the same code are merged, and if the resulting code has fewer than three characters, zeros are added as padding. By applying the Soundex algorithm, names are encoded into phonetic codes that capture their phonetic similarities. It enables the detection of similar-sounding names indicating potential duplicates. 


\noindent\textbf{Impact on AI: }When training on duplicated data, models may overfit by learning redundant patterns that do not generalize well to unseen data. This over-representation can lead to skewed predictions and unreliable outcomes. Duplicates also increase the dataset size, storage costs, and computational resources required for training. Additionally, they can degrade data quality, making it harder to derive accurate insights and leading to inefficiencies in data processing.

\noindent\textbf{Summary: }\emph{Duplicates refer to the presence of duplicate or redundant instances in a dataset, which can distort analysis and modeling. 
Various similarity metrics, based on Levenshtein distance, Affine Gap Distance, Jaro's distance, Monge et al.'s token-based algorithm, and the Soundex phonetic coding scheme, are available to measure duplicates. 
}

\vspace{-15pt}

\subsubsection{Feature Relevance}
\label{subsubsection:feature_relevance}
This refers to identifying and selecting the most informative features or variables that contribute to an AI model. In a dataset, various features or variables are typically collected for each instance, representing different aspects or characteristics of the data. However, not all features may be equally relevant or valuable for an AI model. Feature relevance metric aims to identify the subset of features that are most influential in making accurate predictions or capturing the underlying patterns in the data.

\textbf{Example:} Consider a dataset for predicting housing prices that includes number of bedrooms, square footage, location, schools, and proximity to amenities. In this case, feature relevance would involve analyzing the relationship between each feature and the target variable (house prices) to determine which ones have the strongest correlation or impact on the predictions. Features that have weak or negligible influence can be excluded. 

\textbf{Metrics in Literature:} The column heterogeneity measure proposed by \citet{4053044} uses soft clustering techniques and mutual information to quantify the relevance of features in a dataset. Soft clustering assigns fractional memberships to data points across multiple clusters.
Mutual information measures the dependence between feature values and soft clustering results, capturing their association. The computed mutual information values are then used to derive column heterogeneity scores for each feature.
Features with higher scores are considered more informative.

A survey by \citet{10.1145/3136625} collectively explores similarity-based feature selection metrics, including Laplacian Score, SPEC, Fisher Score, Trace Ratio, and ReliefF. The Laplacian Score algorithm by \citet{he2005laplacian} constructs affinity and Laplacian matrices to measure similarities and differences among data points, producing scores that prioritize features capturing underlying data structures. 
SPEC, an extension of the Laplacian Score, introduced by \citet{zhao2007spectral}, emphasizes alignment with data structure through spectral analysis.
Duda et al.'s \cite{duda2012pattern} Fisher Score emphasizes comparability within classes and distinctiveness between classes, while the Trace Ratio criterion by \citet{nie2008trace} and ReliefF algorithm by \citet{robnik2003theoretical} emphasize within-class similarity and between-class dissimilarity. Collectively, these algorithms highlight the significance of exploiting data relationships and class structures.

\citet{10.1145/3136625} focus on information-theory-based feature selection methods, including Mutual Information Maximization (MIM) (\citet{lewis1992feature}). MIM relies on the concept of entropy to evaluate the significance of features by measuring the reduction in uncertainty they bring to a classification task. MIM evaluates each feature's significance based on its correlation with class labels.
Features with higher Mutual Information (MI) scores are considered more informative and are selected until the desired number of features is reached. \citet{10.1145/3136625} also provides statistical methods, highlighting their roles and applications in various fields. Among these methods, the Low Variance method measures feature relevance by evaluating variances and removing features with variances below a specified threshold. In binary classification, the T-Score method, proposed by \citet{davis1986statistics}, quantifies a feature's capacity to differentiate classes by calculating T-scores based on class means and standard deviations, with higher scores indicating stronger discriminatory power. Conversely, the Chi-Square Score method, introduced by \citet{liu1995chi2}, assesses feature-class independence through an independence test derived from differences between observed and expected frequencies. Gini Index \cite{gini1912variability} evaluates a feature's partitioning potential across different classes, using class probabilities, considering how effectively its values divide the dataset. The Correlation-based Feature Selection (CFS) by \citet{hall1999feature} evaluates feature subsets worth using a correlation-based heuristic. The CFS score balances predictive power with redundancy using symmetrical uncertainty.


\noindent\textbf{Impact on AI: }By identifying the most important features, AI models can concentrate on key data aspects, which reduces noise and computational demands and enhances performance. This process helps mitigate the ``curse of dimensionality,'' speeds up training, and can prevent over-fitting. It also enhances model interpretability by highlighting key factors driving predictions or decisions. 

\noindent\textbf{Summary: }\emph{Feature relevance helps in identifying and selecting the most informative and significant features that contribute to the predictive power of AI models. 
Existing feature relevance metrics use statistical techniques such as soft clustering, similarity, and information theory. 
} 

\vspace{-10pt}

\subsubsection{Class Imbalance}
\label{subsubsection:class_imbalance}
In the context of data for AI, class imbalance refers to the highly skewed or uneven distribution of instances among different classes (categories) in a dataset. It means that one or more classes appear more than others, leading to an imbalanced representation. 

\textbf{Example:} 
Class imbalance often appears in datasets with rare event detection, such as credit fraud detection, earthquake prediction, network intrusion detection, customer churn prediction, rare disease diagnosis, rare species of animal sightings, etc. In these cases, rare events or anomalies are the focus of detection. 

Class imbalance can pose challenges during AI model training and evaluation. Models trained on imbalanced data tend to prioritize the majority class, resulting in poor prediction performance for the minority class. In an anomaly detection example, an imbalanced dataset may lead to a prediction model that performs well in predicting normal instances but performs poorly in identifying anomalies, which are often the class of interest.

\textbf{Metrics in Literature:} The Individual Bayes Imbalance Impact Index (IBI3), introduced by \citet{lu2019bayes} assesses the impact of class imbalance on individual samples, providing insights into potential biases and dataset limitations. IBI3 quantifies the difference in posterior probabilities between balanced and imbalanced scenarios, revealing how class imbalance influences classification outcomes. It requires trained models and estimation of posterior probabilities to calculate, making it essential to have access to both for an accurate assessment. IBI3 measures the influence of class imbalance on classification outcomes for each minority class sample, with lower values indicating less impact.

Imbalance Ratio (IR), introduced by \citet{alberto2018learning}, is a widely used metric to quantify the level of class imbalance in a dataset, especially in binary classification problems. It provides a numerical representation of the discrepancy between the majority and minority class instances. IR is calculated by dividing the count of instances in the majority class ($N \textunderscore majority$) by the count of instances in the minority class ($N \textunderscore minority$). A higher IR indicates a more imbalance.
\citet{ORTIGOSAHERNANDEZ201732} propose Imbalance Degree (ID) as a metric to measure class imbalance, considering specific characteristics of the class distribution.
Despite its advantages, ID has drawbacks, such as sensitivity to the choice of distance function and potential unreliability in extreme cases. In contrast, \citet{ZHU201836} introduce the Likelihood Ratio Imbalance Degree (LRID) to overcome ID's limitations. LRID uses the likelihood ratio (LR) test, providing a high-resolution measurement of imbalance by comparing empirical class distribution to a balanced distribution.
Gupta et al. \cite{gupta2021data} propose the class parity metric, which considers various data properties, including the imbalance ratio, dataset size, and proportion of difficult samples in the extreme minority class. 


\noindent\textbf{Impact on AI: }Class imbalance in datasets can significantly impact AI systems, particularly in classification tasks. Class imbalance can cause models to overlook important patterns in minority classes, potentially leading to misclassification.

\noindent\textbf{Summary: }\emph{Class imbalance in AI-ready data is assessed using metrics like Imbalance Ratio (IR), with Imbalance Degree (ID) offering nuanced measurements. Likelihood Ratio Imbalance Degree (LRID) provides a high-resolution assessment through the likelihood ratio test. Gupta et al.'s class parity metric considers imbalance ratio, dataset size, and difficult samples.
}

\vspace{-7pt}
\subsubsection{Class Separability}
\label{subsubsection:class_separability}
Class Separability refers to the degree of similarity or shared characteristics between different classes or categories within the dataset. It measures the overlap or sharing of common features among the data points from different classes.

\textbf{Example:} Consider a facial recognition dataset consisting of two classes: ``smiling'' and ``not smiling.'' The overlap in this dataset refers to the extent to which the facial features of individuals in these two classes.
If the dataset contains many instances where individuals in both classes have similar facial expressions or features, it indicates a high overlap. 

\textbf{Metrics in Literature:} 
Gupta et al.'s Data Quality Toolkit (DQT) \cite{IBM_DQT} introduces a class overlap metric, which quantifies overlapping regions among different classes in a dataset by analyzing data points in overlapping regions of the data space. Additionally, the evaluation of class overlap in imbalanced classification settings is addressed through metrics such as the $R$-value and augmented $R$-value. Sejong's \cite{OH2011115} $R$-value assesses the extent of overlap between classes by considering the proportion of instances in a specific class located in regions of the feature space shared with instances from other classes. \citet{borsos2018dealing} enhance this approach with the augmented $R$-value, which considers the dataset's imbalance ratio (IR). It provides a weighted measure that combines class overlap and dataset imbalance for a more comprehensive understanding.

\noindent\textbf{Impact on AI: }Class separability impacts AI systems in classification tasks by influencing a model's ability to accurately distinguish between different categories. The model can establish clear decision boundaries when classes are well-separated in the feature space. This leads to improved accuracy, faster training, better generalization to new data, and increased resilience against noise. This also enhances model interpretability, making the decision-making process more transparent. 
Conversely, low-class separability can lead to more misclassifications.

\noindent\textbf{Summary: }\emph{Class separability is the level of similarity among diverse classes in a dataset. DQT introduces a metric to detect overlapping areas between classes, assessing data points that are close yet belong to different classes. The R-value is a measure proposed to quantify the degree of overlap in imbalanced classification problems. 
}
\vspace{-10pt}

\subsubsection{Discrimination Index}
\label{subsubsection:discrimination_index}
Discrimination in data refers to 
biases 
that may cause discriminatory outcomes in AI systems. It 
measures unfairness or unjust treatment towards individuals or groups that may be encoded in the data.

\textbf{Example:} 
Consider a company that uses an AI system to filter job applicants based on their resumes. The AI model is trained on historical data of successful candidates and their qualifications. 
If the historical data is biased and reflects discriminatory practices, such as favoring candidates from certain demographic groups, the AI model may unknowingly sustain those biases and lead to unfair outcomes in the hiring process. 

\textbf{Metrics in Literature:} The Difference metric, introduced by \citet{10.1145/3552433}, assesses the degree of bias within a dependency by comparing the confidence of that dependency with and without consideration of sensitive attributes. A higher Difference value indicates a stronger indication of unfair behavior. It is further supported by the Approximate Conditional Functional Dependency (ACFD). 
Additionally, the authors propose the P-Difference metric 
to measure the impact on dependency confidence by excluding one sensitive attribute at a time. This highlights the influential attributes contributing to unfairness.

\citet{10.1145/2783258.2783311} introduce the ``Likelihood Ratio'' ($LR_+$) metric to measure disparate impact in a dataset, calculated based on sensitivity and specificity. It assesses the impact of the protected class on classification outcomes, but it requires a model trained on the dataset to generate results. \citet{celis2020data} introduce two metrics for assessing discrimination based on sensitive attributes. The ``representation rate'' measures fairness by checking how well different attribute values are represented compared to a set threshold. The ``statistical rate'' evaluates fairness by analyzing the conditional probabilities of class labels given attribute values, helping to identify potential discrimination. These metrics provide quantitative fairness evaluation, offering flexibility based on specific application requirements.

\citet{simonetta2021metrics} introduce two metrics that contribute to assessing fairness, bias, and completeness. The first metric, a ``combinatorial metric,'' evaluates dataset completeness by focusing on the distinct combinations of categories within specific columns. It quantifies completeness by comparing the total count of unique data points to the expected number of distinct combinations. 
In contrast, the second metric, based on ``frame theory,''\cite{daubechies1992} offers a sophisticated approach to measuring fairness and bias. It treats the dataset as a matrix and applies operations to analyze the distribution of vectors within the matrix. Eigenvalues obtained from this matrix assessment measure the tightness of the frame, with uniform eigenvalue distribution indicating a balanced dataset. The Gini-Simpson index (\cite{simpson1949measurement}) is used to assess balance and homogeneity further. 
The combinatorial metric targets the representation of distinct combinations, while the frame theory-based metric considers the overall distribution and balance of the dataset's vectors. 



The Amazon SageMaker Developer Guide \cite{Kemka_2019} uses various metrics for identifying bias in data.
Class Imbalance (CI) measures sample distribution across the sensitive attributes, and Difference in Proportions of Labels (DPL) assesses outcome disparities. The guide also leverages various divergence metrics like Kullback-Leibler (KL), Jensen-Shannon (JS), and Lp-norm evaluate differences in the outcome distributions across demographic facets. Total Variation Distance (TVD) and Kolmogorov-Smirnov (KS) measure the degree of distribution divergence, and Conditional Demographic Disparity (CDD) assesses outcome disparities within subgroups. In contrast, DQT \cite{IBM_DQT} presents a disparate impact measure to quantify group discrimination, offering a score for assessing fairness. DQT also includes remediation strategies to mitigate bias in data. 

\noindent\textbf{Impact on AI: }Unfair data can significantly impact AI systems by leading to biased and incorrect decisions.
When AI is trained on biased or unrepresentative data, it can produce outcomes that systematically disadvantage certain groups based on race, gender, socioeconomic status, or other factors. This can result in discriminatory practices in critical areas such as hiring, lending, healthcare, and law enforcement. 

\noindent\textbf{Summary: }\emph{The discrimination index allows analysts to quantify and measure biases or discriminatory outcomes encoded in the data used for training and deploying AI models. The metrics, such as the Difference metric, P-Difference metric, Likelihood Ratio ($LR_+$), representation rate, statistical rate, completeness metric, divergence metrics, and frame theory-based metrics provide quantitative measures to detect discriminatory behavior in the dataset.}

\vspace{-7pt}
\subsubsection{Data Split Ratio}
\label{subsubsection:data_split_ratio}
Optimal data splitting in AI involves dividing a dataset into training, validation, and testing subsets to maximize the performance and generalization of the AI model. This metric aims to allocate the appropriate proportions of data for effective model training, hyperparameter tuning, and unbiased evaluation.

\textbf{Example:} 
Datasets are typically split with a ratio of $60$/$20$/$20$ for training, validation, and testing. Split ratios of $70$/$15$/$15$, $80$/$10$/$10$ are also common. By splitting the dataset in this manner, we can ensure that the AI model is trained on diverse and representative data, fine-tuned for optimal performance, and tested on unseen instances, enabling a robust sentiment analysis system.

\textbf{Metrics in Literature:} \citet{AFENDRAS2019286} suggests that irrespective of data distribution or analytic task, the optimal training sample size in cross-validation is identified as half of the total sample size. Similarly, \citet{roshan2022optimal} examines the ideal data splitting ratio for training and validation sets in linear regression models. The authors propose a ratio of $\sqrt{p}:1$, where $p$ is the number of parameters required to estimate a well-fitting linear regression model. The authors also present a strategy for determining $p$ using variable selection methods. It suggests that this approach can be helpful in regression and classification tasks.

\noindent\textbf{Impact on AI: }An optimal split ensures that the model is trained on a sufficient amount of data to learn effectively while being validated and tested on separate, representative subsets to evaluate its generalization capabilities. An inappropriate split ratio can lead to overfitting or underfitting. Additionally, ensuring that the split maintains the statistical distribution of the data is crucial to avoid biases.

\noindent\textbf{Summary: }\emph{The data split ratio, which involves dividing a dataset into training, validation, and testing subsets, is crucial for optimizing AI model performance. Affendras et al. suggest a guideline, proposing the training set to be half of the total dataset. With metrics like the $\sqrt{p}:1$ ratio guide to ideal splits in linear regression models.}

\vspace{-7pt}
\subsubsection{Data Point Impact}
\label{subsubsection:data_point_impact}
It refers to the measure of the influence or significance of individual data points within a dataset. It quantifies the extent to which each data point contributes to an AI model or system's overall performance, accuracy, or behavior.

\textbf{Example:} Consider a patient medical record dataset with the patient's age, medical history, symptoms, and diagnostic outcomes. By analyzing the impact of data points, we can determine which specific patient records have a higher influence on the outcomes of an AI model built for disease diagnosis.

\textbf{Metrics in Literature:} \citet{ghorbani2019data} introduced the Data Shapley metric, based on the Shapley value from cooperative game theory. It assesses the impact of individual data points in supervised machine learning. The metric measures the contribution of each data point to the model's predictions, revealing its importance in model training. Various techniques, such as Monte Carlo and gradient-based methods, estimate a data point's impact by considering its combinations with different subsets of the training data. Similarly, \citet{wang2023data} propose the Banzhaf value, a metric to assess data point value in the presence of noisy model performance scores. The authors investigate the robustness of data valuation in stochastic gradient descent, where randomness can lead to inconsistent value rankings. 

In addition to these, several other methods have been developed to measure data importance, including the Leave-One-Out (LOO) \cite{cook1982residuals} evaluation, which assesses model performance changes when individual data points are removed, influence functions \cite{koh2017influence} estimate a data point's effect based on loss function gradients, and k-nearest neighbors (KNN) approaches analyze proximity to decision boundaries. Core-set selection \cite{bachem2017coreset} identifies impactful subsets that perform similarly to the full dataset. Local Interpretable Model-agnostic Explanations (LIME) \cite{ribeiro2016lime} provide insights by approximating complex models with interpretable ones around specific predictions.

\noindent\textbf{Impact on AI: }Evaluating the influence of specific data points allows for better understanding and optimization of the model, ensuring that the most relevant and informative data is used while minimizing noise and irrelevant information. This process can improve the accuracy, efficiency, and generalization capabilities of AI systems, leading to more reliable predictions and insights. Additionally, understanding the impact of data points helps identify and mitigate biases, ensuring fair and equitable AI outcomes.

\noindent\textbf{Summary: }\emph{Data point impact refers to the measure of the influence or significance of individual data points within a dataset. 
In addition to feature importance metrics, such as Shapley and Banzhaf values and LIME, influence functions and ablation (removing data points) approaches like LOO are used to measure data point impact. 
}
\vspace{-10pt}

\subsubsection{Correctness}
\label{subsubsection:correctness}
In terms of data values for AI, correctness refers to the degree of accuracy and fidelity in representing the information of the system being analyzed. It measures how closely the recorded data values align with the actual values they are supposed to represent. The goal of the metric is to minimize discrepancies between the recorded data and the ground truth.

\textbf{Example:} Consider a dataset containing temperature measurements from weather stations. The correctness of data values would involve ensuring that each recorded temperature value accurately reflects the actual temperature at the corresponding location and time. Inaccuracies in the recorded values compared to the actual temperature values would indicate a lack of correctness in the dataset.

\textbf{Metrics in Literature:} Pipino et al's\cite{10.1145/505248.506010} correctness metric quantifies data accuracy by calculating the complement of the error ratio. It focuses on clear criteria, like precision levels, and recognizes the contextual variations in error tolerance.  This ensures a systematic evaluation of data correctness. Similarly, \citet{Kaiser_Klier_Heinrich_1970} involves comparing attribute values in the dataset ($w_I$) with their corresponding values in the real world ($w_R$). A domain-specific distance function, denoted as $d(w_I,w_R)$, quantifies the difference between these attribute values. The objective of the metric is to ensure normalization within the interval $[0, 1]$ without using a quotient. 

\noindent\textbf{Impact on AI: }Correct and accurate data ensures that AI models can learn true patterns and make precise predictions to reduce the likelihood of errors and biases. Inaccurate data can lead to flawed outcomes, such as false positives or negatives, undermining the effectiveness of the AI application by spoiling user trust. It also enhances the generalization and robustness of models, enabling them to handle diverse inputs and perform well in real-world scenarios.

\noindent\textbf{Summary: }\emph{In the context of data values for AI, correctness refers to how accurately the recorded data represents the ground truth. They involve calculations related to error ratios, precision criteria, contextual error tolerance, and comparisons between dataset attribute values and their real-world values.}
\vspace{-10pt}

\subsubsection{Timeliness}
\label{subsubsection:timeliness}
The timeliness of data refers to the time data collection and its relevance to the phenomenon or domain being studied in an AI application. It measures how closely the data captures the most relevant information available at the time of analysis or model training, ensuring that the data is up-to-date and reflects the present conditions. The metric may vary depending on the question being solved by an AI application. In some applications, the latest data may be needed. In others, data relevant to the pattern an AI application is trying to predict. Existing metrics define timeliness based on the existence of the most recent data.

\textbf{Example:} Consider an AI system that predicts product demand for an e-commerce company. Timeliness of the data used for training the model would involve using the most recent sales data, customer preferences, and market trends. If the dataset contains sales data from several months ago, it may not accurately capture the current demand patterns and consumer behavior related to the current and near-future conditions. By ensuring timely data, such as incorporating daily or weekly sales updates, the AI model can better adapt to the changing market dynamics and provide more accurate demand predictions.

\textbf{Metrics in Literature:} Two studies, one by  \citet{Kaiser_Klier_Heinrich_1970} and the other by  \citet{HEINRICH201582}, introduce metrics for evaluating the timeliness of attribute values. Both use probability-based approaches to assess the freshness and relevance of data. Kaiser et al.'s metric uses an exponential distribution model to calculate attribute decline rates. It indicates the average proportion of outdated attribute values within a specified time frame. It quantifies attribute age based on data quality assessment time and data acquisition time to offer an automated and interpretable measure of timeliness. In contrast, Heinrich et al. propose a probability-based currency metric (PBCM) that assesses data item timeliness using a set of probabilities. These probabilities are derived from diverse data sources and methods, including expert assessments, historical data analysis, and machine learning algorithms. 
While both metrics share a foundation in probability theory, Kaiser et al.'s metric focuses on attribute-level timeliness. In contrast, Heinrich et al.'s PBCM assesses data item timeliness, offering flexibility for various data types and contexts.

\citet{10.1145/1891879.1891881} propose a method to assess data timeliness using a classification model. They introduce a metric called $T$, which measures the impact of introducing new and more current data into the dataset. To evaluate data volatility and timeliness, the authors replace a percentage of old instances with new ones in the training data while assuming a fixed currency. The $T$ metric is computed based on the total quantity of records in the training and test data and the number of replacement records introduced for reclassification. 

\noindent\textbf{Impact on AI: }
Timely data is critical to AI applications trying to understand patterns, although the `time' refers to the question an AI application is trying to answer. Applications trying to understand and predict patterns related to the latest trends require the most recent data.

\noindent\textbf{Summary: }\emph{The timeliness metric assesses the recency and relevance of the data about the current state of the phenomenon or domain it represents. Kaiser et al. and Heinrich et al. introduced metrics for assessing data timeliness. 
Blake et al. evaluate data timeliness through a classification model by assessing data replacement's impact on model performance.}
\vspace{-10pt}

\subsubsection{Privacy Leakage}
\label{subsubsection:privacy_leakage}
Data privacy in the context of AI refers to protecting and preserving sensitive information contained within datasets, particularly concerning the risk of unauthorized disclosure or inference of private details. 

A notable technique used to assess privacy is Membership Inference Attacks (MIA). MIA determines whether a specific data record was included in the training dataset used to build an AI model. By exploiting patterns and characteristics of the model's outputs, one can infer whether or not a particular data point was part of the training set. This raises concerns about the privacy of individual data records and the potential for unauthorized access to such information.

Another dimension of AI-related privacy issues emerges with the use of synthetic data. Synthetic data is generated to mimic real data while preserving privacy by avoiding the use of actual sensitive information. However, suppose the synthetic data is too close to the real data. In that case, it can reveal private details, making it vulnerable to privacy attacks. The balance between creating useful synthetic data and ensuring privacy remains a significant challenge in AI. Evaluating the closeness of the synthetic data to the real data can be useful in determining the potential privacy leaks that can emerge during AI applications.

\textbf{Example:} Consider a healthcare dataset used to train a machine-learning model for disease diagnosis. The dataset contains sensitive medical information about patients, including their symptoms, test results, and diagnoses. 
Without proper setting of privacy for data, one may determine whether a specific patient's data was used during training, which poses a privacy risk as it could reveal a patient's medical condition or other confidential information.

\textbf{Metrics in Literature:} 
\citet{https://doi.org/10.1111/bjet.13223} focus on mitigating re-identification risks in released datasets within the education sector. In contrast to existing approaches that often assume prior knowledge, the proposed method employs a Markov Model to quantify re-identification risks by using all available information in the datasets, including event-level details associating multiple records with the same individual and exploring correlations between attributes.

In a broader context of privacy metrics in literature, works such as SHAP\scalebox{0.8}{R} introduced by \citet{duddu2022shapr} and Song et al.'s \cite{272134} privacy risk metric are noteworthy. SHAP\scalebox{0.8}{R} quantifies the susceptibility of individual training data records to membership inference attacks by calculating Shapley values, emphasizing the influence of specific data points on model predictions. In contrast, Song et al.'s metric assesses the likelihood of a data record being present in a model's training dataset, focusing on evaluating the privacy risk from an adversarial perspective. Both metrics aim to address privacy concerns by assessing the privacy risks associated with data records. However, they depend on the specific AI model used in the context. 

In another study, \citet{carlini2022privacy} introduced the Attack Success Rate (ASR) as a metric for privacy leakage. ASR measures the success of an attack in predicting if a specific example is part of the training dataset. It is calculated by training a model, performing an attack, and evaluating the attack's success in correctly predicting membership. However, ASR can only be measured after training a model and conducting an attack. Alongside these contributions, \citet{4550303}, \citet{Longpr2017EntropyAA}, and \citet{9458108} proposed entropy-based metrics to offer valuable insights into dataset anonymity to measure unpredictability and disorder in released data.

Regarding synthetic data privacy leaks, Aindo AI's \cite{aindoprivacy} privacy score assesses the privacy risk of synthetic data by comparing proximity ratios between real and synthetic datasets. The process involves calculating the Train to Train Proximity Ratio (TTPR) for real data and the Train to Synthetic Proximity Ratio (TSPR) for synthetic data. The score is derived from the ratio of records below a specific threshold in both distributions. A score of $100$ indicates minimal privacy risk, while lower scores reflect higher risk.
 
\noindent\textbf{Impact on AI: }Evaluating privacy in data impacts AI by ensuring that personal information is protected while enabling AI systems to function effectively. Privacy assessments help identify and mitigate risks associated with data breaches, intentional or unintentional or malicious accesses, and misuse, which are critical as AI technologies increasingly process personal data. Additionally, addressing privacy concerns helps avoid ethical and legal repercussions, such as biases in AI models and non-compliance with regulations like GDPR \cite{GDPR2016a}. Privacy evaluations also encourage the adoption of privacy design principles in data, ensuring that AI systems are developed with privacy considerations. This will ultimately lead to more ethical and responsible AI deployment.

\noindent\textbf{Summary: }\emph{The privacy leakage metric in AI aims to safeguard sensitive information from unauthorized disclosure. Privacy metrics range from a Markov Model to address re-identification risks to SHAP\scalebox{0.8}{R} to quantify membership privacy risk using Shapley values. Attack Success Rate (ASR) is a measure of privacy leakage in membership inference attacks. Entropy-based metrics are also explored to assess dataset anonymity. 
}
\vspace{-10pt}

\subsubsection{Sample Size}
\label{subsubsection:sample_size}
Sample size refers to the number of data points or instances selected from a population to be included in a dataset for training an AI model. It represents the subset of data used to make inferences or predictions. 

\textbf{Example:} 
If researchers collected data from $500$ patients to predict the likelihood of a disease based on patient characteristics, such as their age, gender, medical history, and test results, the sample size of the dataset would be $500$. 


\textbf{Metrics in Literature:} \citet{ALWOSHEEL2018167} investigate the sample size requirements for accurate decision-making analysis using Artificial Neural Networks (ANNs). They introduce a new guideline, ``factor $50$,'' recommending that the optimal dataset size for an ANN should be the number of adjustable parameters in the model multiplied by $50$. This guideline is more conservative than the commonly used ``factor $10$'' rule-of-thumb found in the literature \cite{haykin2009neural}. However, determining the appropriate sample size for ANN-based decision-making is complex and depends on the model's complexity, which is difficult to predict. To address this, the authors propose three approaches: evaluating ex-post if the training sample size was sufficient, using prior studies or literature to estimate the optimal number of neurons, or referring to existing literature to calculate the expected number of neurons required for the analysis. 

\noindent\textbf{Impact on AI: }An adequate sample size ensures that AI models can learn effectively from the data, capturing the underlying patterns without overfitting or underfitting. Small sample sizes may lead to overfitting, where the model performs well on training data but poorly on new, unseen data due to learning noise or random variations rather than the true signal. Conversely, excessively large sample sizes can be resource-intensive and may not significantly improve model performance beyond a certain point. \citet{Rajput2023} have shown varying impacts of sample size on model accuracy, with some models performing better with larger datasets, while others may not see significant gains past a certain threshold.

\noindent\textbf{Summary: }\emph{In AI, sample size denotes the quantity of data points chosen from a population for analysis or model training. 
The ``factor $50$'' determines sample size in decision-making with Artificial Neural Networks (ANNs), contrasting with the commonly used ``factor $10$''.}
\vspace{-10pt}

\subsubsection{FAIR Principle Compliance Score}
\label{subsubsection:FAIRness_score}
FAIR compliance of a dataset for AI refers to the degree to which a dataset adheres to Findability, Accessibility, Interoperability, and Reusability (FAIR) principles \cite{FAIR:Principles}. In AI, a dataset should be well-documented, easily discoverable, accessible, formatted in a way that facilitates integration and analysis, and can be effectively reused for AI applications.

\textbf{Example:} Consider a dataset containing information about different types of cars for training an AI model tasked with predicting car prices. The FAIR principles applied to this dataset would be: 
\begin{itemize}
    \item Findability: The dataset should contain a unique identifier to be easily discoverable with standardized metadata containing details about car attributes, including make, model, year, mileage, engine type, and other existing features. Clear information on the dataset's source and data collection methodologies is also essential.
    \item Accessibility: The dataset's accessibility requires 
    a platform with controlled access, considering privacy or licensing constraints. Secure access should be given through proper authentication and authorization methods.
    \item Interoperability: The dataset should be structured and formatted according to established storage standards like CSV or JSON to ensure easier integration with AI systems and tools. A well-defined schema must be included with the dataset, specifying the meaning and format of each attribute. This promotes consistency and compatibility across diverse AI models and applications.
    \item Reusability: The dataset should come with clear usage licenses or permissions that outline how the dataset can be used. Additionally, comprehensive documentation should be included with the dataset, providing details about data collection procedures, preprocessing steps (such as data cleaning or feature engineering), and any potential biases or limitations in the data.
\end{itemize}

By following these FAIR principles, a structured dataset of car information becomes a valuable resource for AI researchers and practitioners. It facilitates the development of accurate car price prediction models and promotes transparent and ethical AI practices.

\textbf{Metrics in Literature:} \citet{Wilkinson_Sansone_Schultes_Doorn_Bonino_da_Silva_Santos_Dumontier_2018} introduced a comprehensive FAIR compliance measurement framework, aligning with the four FAIR sub-principles. The framework includes $14$ universal metrics corresponding to specific sub-principles, covering aspects such as identifier schemes, metadata accessibility, findability, access protocols, metadata longevity, knowledge representation languages, linking, adherence to standards, and provenance. This flexible approach facilitates the objective assessment and improvement of FAIR compliance in various digital resources applicable across scholarly domains. In a similar framework introduced by \citet{clarke2019fairshake}, FAIR metrics and FAIR rubrics play an important role, allowing users to associate digital resources with existing metrics. The authors emphasized the manual or automated quantification of FAIR metrics and contextual assessments using FAIR rubrics. This empowers users to evaluate and enhance data correctness in diverse projects. DataONE (Data Observation Network for Earth) \cite{Jones_Slaughter_2019} is a community-driven initiative that has adopted metrics to measure FAIR principle compliance of research data. Based on the FAIR criteria, the DataONE FAIR suite generates comprehensive assessment scores based on the metadata.

\noindent\textbf{Impact on AI: }A dataset's compliance with the FAIR principles can impact AI by enhancing data management and accessibility. By ensuring data is easily discoverable and accessible, AI systems can be trained on diverse and comprehensive datasets, leading to robust and accurate models. Interoperability allows different AI systems to work together seamlessly, while reusability enables researchers to build upon existing datasets and models, enabling reproducible innovation and efficiency. Moreover, FAIR compliance promotes collaboration and transparency that accelerates AI research and ensures the reproducibility of results. It is essential for building trust in AI technologies.

\noindent\textbf{Summary: }\emph{The FAIR score measures the extent to which a dataset adheres to the principles of Findability, Accessibility, Interoperability, and Reusability. Both Wilkinson et al. and Clarke et al. created frameworks to assess the FAIR compliance of digital resources aligned with FAIR sub-principles. Several other FAIR compliance score evaluation websites use the same $14$ principles that Wilkinson et al. defined. 
}

\begin{tcolorbox}[size=title, top=5px, bottom=5px, boxrule=0.1mm, arc=0.5mm, colback=white, colframe=black, fonttitle=\sffamily\bfseries, title=Data Readiness Metrics for Structured Data]
In evaluating structured data readiness for AI applications, we considered metrics spanning completeness, outliers, mislabels, duplicates, feature relevancy, class imbalance, class separability, discrimination index, data split ratio, data point impact, correctness, timeliness, privacy leakage, sample size, and FAIR compliance score. 
\end{tcolorbox}

\vspace{-9pt}
\subsection{Unstructured Data}
Unstructured data, including textual, image, and audio data, present unique challenges in evaluating and ensuring readiness for AI applications. In this section, we provide a brief overview of the metrics and scoring mechanisms (shown in the right half of Table \ref{tab:dim_cite}) used to assess the suitability of unstructured data for AI. While the evaluation techniques for structured data are well-established, we will highlight the relevant measuring techniques from structured data that can be applied to unstructured data. By examining these evaluation methods, we gain insights into the readiness, relevance, and accuracy of unstructured data. It enables us to make informed decisions when using such data in AI models. 

\subsubsection{Textual Data} 

\vspace{-10pt}
\paragraph{Lexical Diversity}
Lexical diversity measures the richness, variety, and complexity of the vocabulary used within the text. It offers insights into the level of linguistic expression, domain coverage, and potential challenges in understanding and processing textual data. A higher lexical diversity score indicates a broader range of words and linguistic patterns, providing a solid foundation for AI to learn from the data.

\textbf{Example:} Consider two datasets: \textit{A} with a low lexical diversity score and \textit{B} with a high lexical diversity score. Dataset \textit{A} has a repetitive and limited vocabulary, such as a chatbot dialogue focused on a specific topic. The low lexical diversity suggests a constrained range of language, potentially limiting the chatbot's ability to respond effectively to diverse user inputs. In contrast, dataset \textit{B} consists of a collection of news articles from various domains, exhibiting a wide range of vocabulary and language styles. The high lexical diversity in dataset \textit{B} indicates a greater readiness for AI applications. It offers a more comprehensive representation of language usage, enabling models to generalize across different topics and understand a broader range of inputs.

\textbf{Metrics in Literature:} Type-Token Ratio (TTR) introduced by \citet{templin1957certain} measures lexical diversity in textual data. It calculates the ratio of unique word types (vocabulary size) to the total number of tokens (words or other linguistic units) in the text. TTR provides an estimate of the text's richness and variety of vocabulary. A higher TTR indicates greater lexical diversity, suggesting a more comprehensive range of word usage in the text.

McCarthy et al.'s \cite{mccarthy2010mtld} metrics, vocd-D and HD-D, focus on measuring lexical diversity in textual data. vocd-D uses type-token ratios (TTR) from randomly selected text samples to derive a D coefficient to represent lexical diversity. HD-D, on the other hand, employs the hypergeometric distribution to directly calculate the probabilities of word occurrence in randomly selected samples, resulting in the HD-D index. While both metrics assess lexical diversity, vocd-D uses random sampling and the D coefficient, whereas HD-D approximates results for all possible word arrangements. The correlation between HD-D and vocd-D is high, offering alternative methods of measuring lexical diversity. Additionally, McCarthy introduces the Measure of Textual Lexical Diversity (MTLD) \cite{mccarthy2005assessment}, which quantifies lexical diversity by considering unique words and segment length. MTLD assesses the lexical diversity of longer texts, providing insights into vocabulary richness throughout the text.

\noindent\textbf{Impact on AI: }Measuring lexical diversity can significantly impact AI by influencing how language models are developed and evaluated. In AI, particularly in Natural Language Processing (NLP) and machine translation, assessing lexical diversity can help identify biases and limitations in language models. \citet{reviriego2023playingwordscomparingvocabulary} state that AI-generated texts often exhibit lower lexical diversity compared to human-generated texts, leading to a feedback loop where AI models trained on such data become less effective over time. Ensuring high lexical diversity in AI outputs can improve the quality and accuracy of translations and other language-based AI applications. This ultimately contributes to more inclusive and representative AI systems.

\noindent\textbf{Summary: }\emph{Lexical diversity, necessary for assessing linguistic richness in textual data, is measured by metrics like TTR and vocd-D, evaluating the ratio of unique word types to total tokens with standardized scores. HD-D offers a probability-based alternative for assessing diversity. MTLD divides the text into segments and calculates the average segment length where vocabulary richness falls below a threshold.}

\vspace{-10pt}
\paragraph{Term Importance}
Term importance evaluates the significance of individual terms in textual data. It measures the relevance and impact of terms in capturing the essence and meaning. Term importance considers various factors, such as the frequency of a term within the text, its rarity across the entire dataset or corpus, and its discriminative power in distinguishing the text from others. By assigning weights or scores to terms based on their importance, this metric enables AI models to focus on key terms that carry valuable semantic information and discard less informative or common terms, assisting in feature selection, document ranking, or topic extraction. 

\textbf{Example:} In a dataset of news articles, the term ``pandemic'' might be considered highly important due to its relevance in conveying crucial information regarding COVID-19. On the other hand, common words like ``the'' or ``and'' would be assigned lower importance scores as they provide little discriminative power or unique information. AI models can prioritize and focus on the most significant terms by analyzing term importance, enabling better understanding, classification, or summarization of textual data. 

\textbf{Metrics in Literature:} TF-IDF (Term Frequency-Inverse Document Frequency) is a widely used quantitative metric to evaluate the importance of words in a document or collection of documents (\citet{ramos2003using}, \citet{Simha_2021}, \citet{Qaiser_S}). It combines term frequency (TF), measuring word occurrence in a document, and inverse document frequency (IDF), assessing the rarity of a word's appearance across the entire corpus. TF-IDF quantifies a word's significance by considering its frequency in a document and its discriminative power across the corpus. The TF-IDF score is computed by multiplying TF and IDF values for each word. High TF-IDF scores indicate that words that are both frequent in a document and rarely appear in the corpus, which makes them essential for the context. This metric helps identify essential features and characteristics, enabling information retrieval, document classification, and keyword extraction in natural language processing.

\noindent\textbf{Impact on AI: }Measuring term importance is needed for tasks such as information retrieval, text summarization, and sentiment analysis, where distinguishing between relevant and irrelevant terms can greatly impact the accuracy and relevance of the output. By measuring term importance accurately, AI systems can prioritize critical information that leads to precise and contextually relevant results. Furthermore, TF-IDF aids in reducing computational resource requirements in AI training by focusing on significant terms.

\noindent\textbf{Summary: }\emph{Term importance assesses the significance of individual terms in textual data for AI applications. TF-IDF is a widely used metric that combines term frequency and inverse document frequency.}

\vspace{-10pt}
\paragraph{Readability Score}
Readability score is a quantitative metric used to assess textual data complexity and ease of understanding, enabling effective preparation for AI. It measures various linguistic factors, such as the length of sentences, choice of words, and syntactic structure, to determine the readability of a text. By considering these factors, readability scores provide valuable insights into the suitability of text for different target audiences and applications. AI applications can be optimized by selecting appropriate training data using readability scores, ensuring that the content aligns with the desired level of comprehension and avoids potential barriers to understanding. This metric is important in enabling the development of more accessible and contextually appropriate language models.

\textbf{Example:} Consider a scenario where an AI model is trained to generate educational content for an elementary school. In this case, readability scores can be used to assess the complexity of different texts and select appropriate training data. The readability scores can help identify texts that align with the target audience's reading abilities by analyzing factors such as sentence length, vocabulary difficulty, and grammatical complexity. 
This ensures that an AI model is trained on comprehensible and engaging content for young learners, promoting effective knowledge transfer and enhancing the overall learning experience. 

\textbf{Metrics in Literature:} The Flesch-Kincaid Grade Level, introduced by \citet{flesch1986art} estimates the approximate education (grade) level needed to comprehend a given text using the average number of words in sentences and that of syllables in words.
The resulting score represents the education level required to understand the text. Lower grade values indicate higher/easier readability. This metric provides a standardized measure for assessing text comprehension and enables content tailoring to suit specific audience reading abilities.

The Coleman-Liau Index, developed by \citet{Coleman1975ACR}, is another readability scoring method that assesses the reading level of a text based on factors like letter count and sentence length. Unlike the Flesch-Kincaid Grade Level, which considers syllable count, the Coleman-Liau Index calculates the grade level based on the average number of letters and sentences per $100$ words. A score of $5$ on the index indicates that the text is at a reading level equivalent to that of a fifth grader in the US schooling system. It is widely used in schools and provides a quick measure of readability.

Furthermore, the Gunning Fog Index introduced by Robert Gunning Associates \cite{readable_gf} offers an additional perspective on the readability of textual data for AI model training. Unlike the Coleman-Liau Index and the Flesch-Kincaid Grade Level, which focus on sentence and letter count, the Gunning Fog Index considers both the percentage of complex words and the sentence length. It generates a score between $0$ and $20$, with lower scores representing easier readability. 

\noindent\textbf{Impact on AI: }By evaluating readability, developers can adjust the complexity of the training data to match the intended audience's comprehension level, which is particularly important in fields like education and healthcare, where clear communication is essential. By ensuring that training data is readable, AI models can avoid perpetuating biases that arise from overly complex or inaccessible language. This will also reduce disparities in information access.

\noindent\textbf{Summary: }\emph{The readability score is a quantitative metric used to assess the complexity of textual data and ease of understanding. Popular readability metrics calculate the grade level needed to comprehend a text based on average words per sentence and syllables per word 
and by considering the percentage of complex words and sentence length.} 
\vspace{-10pt}

\paragraph{Topic Coherence}
This measure evaluates the readiness of textual data for AI by assessing the logical and semantic connectedness within a set of topics or a document. It quantifies the degree to which words within a topic exhibit meaningful relationships and contribute to a coherent theme. A higher coherence score shows stronger semantic coherence, indicating that the words are closely related and provide a clearer understanding of the topic. By evaluating topic coherence, AI practitioners can ensure that the textual data is well-structured, coherent, and ready for AI model training. This would promote accurate and meaningful text generation and facilitate better comprehension and usage of data by AI algorithms.

\textbf{Example:} Consider a collection of news articles about technology trends. Topic coherence can be measured to evaluate the readiness of this textual data for AI. Data can be segmented into topics like ``Artificial Intelligence,'' ``Blockchain Technology,'' and ``Internet of Things'' by applying topic modeling techniques. Topic coherence analysis assesses the semantic relationships between words within each topic. A high coherence score would indicate that words within a topic, such as ``machine learning,'' ``algorithm,'' and ``predictive analytics'' in the ``Artificial Intelligence'' topic, are closely related and contribute to a coherent theme. This demonstrates that the textual data is well-prepared for AI, as it exhibits clear and meaningful topic structures.

\textbf{Metrics in Literature:} R"{o}der et al. propose the ``CV coherence score'' \cite{10.1145/2684822.2685324} to quantify topic coherence in textual data by assessing the semantic similarity between words within a topic. This is computed using Latent Dirichlet Allocation \cite{blei2003latent} to extract topics from the text and then measure the pairwise word similarity within each topic to evaluate how well the words contribute to a coherent theme. Higher scores indicate stronger semantic relatedness and better topic coherence. Despite its popularity, the CV coherence score has limitations, such as sensitivity to topic size, potential mismatch with human judgment, and inability to capture higher-level coherence aspects.

Mimno et al. introduce ``UMass coherence score'' \cite{10.5555/2145432.2145462} to assess the topic coherence of a set of topics extracted from a text corpus. It evaluates coherence by considering the probability of word co-occurrences within topics. The score is computed by aggregating the logarithm of the co-occurrence probabilities of all word pairs within and across topics. A higher UMass coherence score indicates stronger word co-occurrence patterns and better topic coherence, while a lower score suggests weaker word associations and less coherent topics. Compared to the CV coherence score, the UMass coherence score offers a more reliable evaluation of topic coherence, accounting for topic size, aligning better with human judgment, and directly measuring word co-occurrence probabilities.

Newman et al. proposed ``UCI coherence score'' \cite{10.1145/1816123.1816156} to assess the coherence of topics generated by a topic model. This measures the semantic relatedness and meaningful connections between words within a topic by calculating the semantic association between word pairs based on their co-occurrence in sliding windows. The score is computed using a specific equation considering the probabilities of observing individual words and word pairs in a sliding window. Higher UCI coherence scores indicate stronger associations between word pairs and better topic coherence.

\noindent\textbf{Impact on AI: }By evaluating and optimizing for topic coherence, developers can ensure that the AI models produce topics that align better with human understanding, which will improve the semantic interpretability of the model outputs. This evaluation can lead to accurate and reliable AI systems, as coherent topics are more likely to represent genuine patterns in the data rather than random associations.

\noindent\textbf{Summary: }\emph{Topic coherence is a metric used to evaluate the logical and semantic connectedness within a set of topics or a document. Coherence scores, such as the CV, UMass, and UCI coherence scores, quantify the semantic similarity or word co-occurrence probabilities within topics to assess their coherence.}
\vspace{-10pt}

\paragraph{Bias Indicator}
A bias indicator is a measure used to prepare textual data for AI by quantifying and identifying potential biases in the text. It serves as a tool to assist in detecting and mitigating biased content, ensuring that AI systems can make more informed and fair decisions. The bias indicator analyzes various linguistic and semantic features within the text, such as word choice, sentence structure, and contextual references, to assess the potential presence of biases related to factors like gender, race, religion, or other sensitive attributes. By providing a quantitative assessment of bias, the indicator helps developers and users understand the underlying biases in the data. It enables them to address and mitigate these biases during AI model development, promoting more equitable and unbiased AI systems.

\textbf{Example:} Consider a dataset of job application statements used to train an AI system to evaluate and rank applicants. A bias indicator can be used to analyze the textual data, ensuring fairness and reducing bias. The indicator would examine language and semantic patterns to identify potential biases, such as gender, race, or age-related biases. For instance, if the indicator detects a bias where certain occupations or characteristics are consistently favored or discriminated against, it would flag it for further examination. This enables developers to address and mitigate biases in the dataset before training the AI model. This promotes equal opportunities and minimizes discriminatory outcomes during the evaluation process. The bias indicator plays a crucial role in preparing the textual data, allowing the creation of more objective and unbiased AI systems for job application assessments.

\textbf{Metrics in Literature:} \citet{10.1145/3351095.3372843} propose a robust bias measurement technique using word embeddings and cosine similarity calculations. The method involves defining word pairs to represent different types of discrimination and creating a list of concepts for measuring bias. For variable concepts, which change based on social groups, the bias calculation equation considers the cosine similarity between word pair embeddings and concept embeddings.
A modified bias calculation equation is used for non-variable concepts, which remain the same irrespective of social groups. By quantifying the differences in cosine distances, the proposed method comprehensively analyzes bias in different contexts, accounting for variable and non-variable concepts.

\noindent\textbf{Impact on AI:} Textual data often contains subtle biases related to gender, race, ethnicity, and other social factors, which can be encoded in language patterns, word associations, and context. If these biases are not identified and addressed before training, an AI model may learn to reinforce these biases, leading to unfair outputs in tasks such as language generation, sentiment analysis, or text classification. This preventive approach leads to ethical and unbiased AI systems.

\noindent\textbf{Summary: }\emph{A bias indicator is used to quantify and identify potential biases in textual data for AI applications. A bias indicator that uses word embeddings and cosine similarity calculations is one of the most used metrics.}

\vspace{-10pt}
\subsubsection{Multimedia Data} 
In this subsection, we explore the key aspects of DRAI for image, speech, and video data. We concentrate on quality evaluation metrics for each data type. This exclusive emphasis on quality evaluation metrics is fundamentally crucial in determining the reliability and usability of these types of data for AI. 


\vspace{-10pt}
\paragraph{Image Quality}
\label{paragraph:image_quality}
As a metric to measure the readiness of image data for AI, image quality refers to the degree to which an image accurately represents the visual content it intends to depict. It includes various aspects such as sharpness, color accuracy, clarity, and the absence of artifacts or distortions. Assessing image quality is crucial in AI applications as it directly impacts the reliability and effectiveness of algorithms that rely on visual data. High-quality images provide a solid foundation for AI to extract meaningful features, recognize patterns, and make accurate decisions. 

\textbf{Example:} In an AI system designed for autonomous driving, image quality plays a vital role in ensuring the accuracy and reliability of object detection and recognition. High-quality images with clear details and accurate colors allow the AI system to accurately identify pedestrians, vehicles, and traffic signs, enabling it to make precise decisions in real time. Conversely, low-quality images with blurriness or artifacts may lead to misinterpretation of objects or false detection, compromising the safety and efficiency of autonomous driving systems. 

\textbf{Metrics in Literature:} The field of image quality assessment is a heavily researched area, constantly evolving to meet the demands of various applications. In this discussion, we explore some of the favored and widely used measures developed to evaluate image quality accurately. 

Mean Squared Error (MSE) and Peak Signal-to-Noise Ratio (PSNR) \cite{psnr} are essential metrics for evaluating image quality and are widely used in AI applications. MSE calculates the average squared difference between the pixel values of a reference and processed image, providing a quantifiable representation of distortion. PSNR, derived from MSE, offers a perceptual quality metric expressed in decibels, comparing the maximum signal power with the average squared error. Higher PSNR values indicate high-quality images. While MSE alone may not align with human perception of quality, PSNR provides a standardized measure that accounts for human visual perception.

\citet{995823} present a universal image quality index, distinct from MSE and PSNR, defined mathematically for independence from specific images, viewing conditions, and observers. This index incorporates correlation coefficient, mean luminance difference, and contrast similarity, offering practical advantages in simplicity and universality. In related work, \citet{1284395} introduce the Structural Similarity (SSIM) index as an alternative to traditional error-sensitive methods, emphasizing structural similarities in luminance, contrast, and structure to align more closely with human visual perception. Furthermore, their Multi-Scale Structural Similarity (MSSIM) approach, as proposed in \cite{1292216}, extends SSIM by considering variations in resolution and viewing conditions, providing increased flexibility and demonstrating improved performance.

Similarly, Sarnoff's JND-Metrix \cite{sarnoff_jndmetrix} takes into account Just Noticeable Differences (JND) based on the Human Visual System (HVS) principles. Unlike traditional metrics such as PSNR or MSE, the JND-Metrix considers the sensitivity and perception of the HVS to different types of image distortions. By incorporating knowledge of the HVS, including factors like contrast sensitivity, spatial masking, and visual attention, the perceptual impact of image distortions is quantified more accurately. JND-Metrix measures the visibility of distortions by estimating JND thresholds for various distortions, indicating the level of distortion perceptually distinguishable from the original image. 

Two notable advancements in image quality assessment include Sheikh et al.'s \cite{1576816} Visual Information Fidelity (VIF) criterion and Chandler et al.'s \cite{4286985} Visual Signal-to-Noise Ratio (VSNR) metric. VIF, a full-reference method, evaluates the correlation between image information and visual quality, outperforming traditional metrics in various scenarios. On the other hand, VSNR uniquely considers human visual system properties, including near-threshold and supra-threshold perception. It provides a more accurate representation of perceived quality by addressing contrast sensitivity and global precedence.

The PyTorch Image Quality (PIQ) \cite{piq} \cite{kastryulin2022piq} provides a diverse array of image quality metrics designed for various assessment requirements. Among full-reference metrics, FSIM (Feature Similarity Index Measure) \cite{5705575} is significant for its ability to evaluate image quality by analyzing structural and color features. GMSD (Gradient Magnitude Similarity Deviation) \cite{Xue_2014} is also noteworthy which focuses on the gradient magnitudes to assess image quality. In the no-reference category, BRISQUE (Blind/Referenceless Image Spatial Quality Evaluator) \cite{6272356} stands out as it evaluates quality using natural scene statistics without needing a reference image. This makes it particularly useful in real-world applications. Within the distribution-based category, FID (Frechet Inception Distance) \cite{heusel2018ganstrainedtimescaleupdate} is a key metric that is used for assessing generative models by measuring the similarity between distributions of real and generated images.

\citet{lakhani2020importance} and \citet{sabottke2020effect} demonstrate that resolution is a critical image quality metric when developing deep learning models for medical imaging and radiology applications. Higher image resolutions can lead to improved AI model performance, especially in detecting specific medical conditions. However, it is essential to strike a balance with computational resources to avoid limitations in the training process. The appropriate image resolution for a given task can vary based on several factors, including the image data type, the specific AI model or algorithm used, and the application's requirements. Additionally, blurriness is another factor that can impact the performance of AI models, particularly those tailored for image recognition, object detection, and segmentation tasks. Additionally, blurriness, a significant factor affecting AI model performance, has been extensively studied. Lin et al.'s \cite{lin2005visual} method estimates contrast decrease on edges, while \citet{marziliano2002no} gauge blurriness by analyzing edge spread, providing valuable insights into overall blurriness levels.

\noindent\textbf{Impact on AI: }
Measuring image quality impacts AI training by ensuring that the data used to train models is clear, accurate, and free from distortions, which directly influences the model's ability to learn effectively. High-quality images allow AI models to extract meaningful features and recognize patterns more accurately, leading to better generalization and performance. Conversely, low-quality images with noise, blurriness, or artifacts can introduce errors, reduce the model's learning efficiency, and lead to inaccurate predictions.

\noindent\textbf{Summary: }\emph{Image quality refers to how accurately an image represents its visual content and covers aspects such as sharpness, clarity, color accuracy, and the absence of artifacts. Commonly used metrics for image quality assessment include MSE, PSNR, SSIM, MSSIM, JND-Metrix, VSNR, and VIF. These metrics provide quantitative measurements of image distortion and quality. Additionally, considering factors like image resolution and blurriness is crucial to balance improved model performance and computational resources.}
\vspace{-15pt}

\paragraph{Speech Quality}
\label{paragraph:speech_quality}
Speech quality refers to the overall perceived clarity, intelligibility, and fidelity of speech signals. It captures the extent to which speech data effectively conveys the intended information and is free from distortions, noise, or artifacts that could impact its understandability. Speech quality includes factors such as signal clarity, absence of background noise, the naturalness of speech, and the ability to capture and reproduce various linguistic and acoustic properties accurately. A high level of speech quality in the data ensures that AI models can effectively process and interpret speech inputs, leading to more accurate and reliable performance across speech-related tasks, such as speech recognition, synthesis, or understanding.

\textbf{Example:} Speech quality is important for the optimal functioning of voice-controlled virtual assistants. In a scenario with high speech quality, a user's command is delivered with clarity and minimal background noise. This ensures the virtual assistant accurately interprets and executes the request. With low speech-quality data with distortions and noise, an AI system may struggle to comprehend the command, leading to potential errors in execution.

\textbf{Metrics in Literature:} The ITU-T (International Telecommunication Union -- Telecommunication Standardization Sector) introduced Mean Opinion Score (MOS) \cite{itu_p808}, which involves human listeners rating the perceived speech quality. 
MOS helps to understand how well the speech data aligns with human expectations, enabling improvements based on listener feedback. On the other hand, Perceptual Evaluation of Speech Quality (PESQ), introduced by \citet{941023}, objectively quantifies the quality of processed or degraded speech signals. It uses a model that simulates human auditory perception to calculate prediction scores. Segmental Signal-to-Noise Ratio (SNRseg), introduced by \citet{jayant1984digital}, is another metric that gauges the ratio of the clean speech signal to the background noise within short segments. By providing a localized evaluation, SNRseg assists in identifying segments of speech that may be affected by noise, leading to targeted noise reduction or enhancement techniques.

Widely used objective metrics for evaluating speech quality and intelligibility are Short-Time Objective Intelligibility (STOI), Perceptual Speech Quality Measure (PSQM), Itakura-Saito Spectral Distortion (IS), and Objective Difference Grade (ODG). STOI, introduced by \citet{5495701}, primarily assesses the similarity between clean and degraded speech signals, focusing on how well the degraded speech retains intelligibility compared to the original signal. In contrast, PSQM, introduced by \citet{beerends1994perceptual}, explores various factors affecting speech quality, including distortion, noise, and echo. IS \cite{itakura1968analysis} quantifies spectral distortion in the frequency domain, helping to understand the impact of different operations on speech quality. On the other hand, ODG, specified in \cite{itu_bs1387}, evaluates the perceived difference in quality between two speech signals, aiding in comparing speech processing algorithms or system configurations. These metrics offer distinct perspectives on speech quality, with STOI and PSQM emphasizing intelligibility and overall quality, IS focusing on spectral distortion, and ODG providing a comparative quality assessment.

\noindent\textbf{Impact on AI: }Evaluating speech quality can significantly impact AI systems, particularly in speech recognition, synthesis, and processing. By assessing audio quality, AI models can improve their accuracy in recognizing speech across various conditions, including noisy environments and diverse accents. For speech synthesis, quality evaluation leads to natural-sounding and intelligible artificial voices. In audio processing, it enables the development of better noise cancellation and audio enhancement algorithms. Moreover, speech quality assessment improves AI training by providing more accurate data labeling and helping to filter out low-quality samples.

\noindent\textbf{Summary: }\emph{Speech quality refers to the perceived clarity, intelligibility, and fidelity of speech signals. Metrics such as Mean Opinion Score (MOS), Segmental Signal-to-Noise Ratio (SNRseg), Short-Time Objective Intelligibility (STOI), Perceptual Evaluation of Speech Quality (PESQ), Perceptual Speech Quality Measure (PSQM), Itakura-Saito Spectral Distortion (IS), and Objective Difference Grade (ODG) are used to assess speech quality objectively and subjectively.}
\vspace{-10pt}

\paragraph{Video Quality}
\label{paragraph:video_quality}
As a metric to evaluate the readiness of video data for AI, video quality refers to the overall visual fidelity and perceptual coherence of a video sequence. It assesses the accuracy with which the video represents the original content, ensuring that crucial details, structures, and visual cues are preserved without significant degradation or distortion. Evaluating video quality involves objective and subjective criteria, where objective assessments employ computational algorithms to analyze pixel-wise differences and feature similarities. In contrast, subjective evaluations incorporate human perception and user feedback. By considering video quality as a crucial factor, AI practitioners can determine the suitability of video data for their applications, ensuring that the data meets the necessary standards for achieving accurate and reliable AI-driven results.

\textbf{Example:} In preparation for developing an AI-powered autonomous driving system, a team of engineers collects a vast amount of video data from various in-car cameras and external sensors. To evaluate the readiness of this video data for AI training, they carefully assess its quality. In this context, video quality refers to the video sequences' overall visual fidelity and coherence, ensuring that critical details are preserved without significant degradation. The engineers analyze the videos to detect potential artifacts or distortions that may impact the AI system's perception algorithms. They also involve human evaluators to rate the video quality subjectively based on their perceived visual appeal. 

\textbf{Metrics in Literature:} PSNR, SSIM, VIF, and VSNR are image quality metrics, discussed in Section \ref{paragraph:image_quality}, that can also be effectively applied to measure the quality of video data. When applied to videos, these metrics analyze individual frames' spatial quality and temporal coherence to capture the visual information across consecutive frames. PSNR, SSIM, VIF, and VSNR provide objective insights into video quality and its perceptual fidelity by assessing the pixel-wise differences, structural similarity, information fidelity, and sensitivity to visual distortions. This adaptability makes them valuable tools for researchers and practitioners seeking to quantify and improve the visual experience in video applications. Furthermore, like speech quality assessments discussed in Section \ref{paragraph:speech_quality}, video quality evaluations often employ the Mean Opinion Score (MOS) methodology. MOS in video quality involves presenting video samples to human viewers who rate their subjective perception of the video's quality. The average MOS scores offer valuable insights into human preferences and perceptual experiences, complementing the objective metrics' findings to make more informed decisions regarding video data suitability for various AI applications.

VQM (Video Quality Metric) introduced by \citet{Huynh-Thu}, VMAF (Video Multimethod Assessment Fusion) introduced by Netflix \cite{Blog_2017}, and OPTICOM's PEVQ (Perceptual Evaluation of Video Quality) \cite{PEVQ} are advanced and specialized metrics specifically designed to assess the quality of video data. VQM focuses on replicating human visual perception to evaluate video quality accurately. By analyzing spatial and temporal characteristics of video frames, VQM provides a comprehensive measure of video fidelity, making it invaluable in video compression and transmission applications. VMAF takes a multifaceted approach by combining traditional metrics like PSNR and SSIM with machine learning techniques. VMAF predicts how viewers perceive video quality by leveraging a human-rated dataset, making it highly effective for video streaming, content delivery, and codec research. Meanwhile, PEVQ, standardized by ITU (International Telecommunication Union), offers both objective and subjective evaluations. Its computational model estimates video quality based on visual and temporal features, while human evaluators provide MOS-based subjective assessments. Widely used in video telephony and conferencing systems, PEVQ ensures that video communication meets acceptable quality standards.

\noindent\textbf{Impact on AI: }Evaluating video quality can significantly impact AI training, particularly for models focused on computer vision, video processing, and generation tasks. By assessing video quality, researchers can curate better training datasets, develop more effective data augmentation techniques, create more accurate labels, and improve the assessment of AI-generated content. This process enables the filtering out of low-quality or corrupted videos that could negatively impact model performance while also allowing for the development of better video generation and enhancement algorithms. Furthermore, incorporating human perception metrics into video quality evaluation helps train AI models to produce results that are visually appealing to human viewers.

\noindent\textbf{Summary: }\emph{Video quality refers to overall visual fidelity and perceptual coherence, assessing its accuracy in representing the original content. Objective metrics such as PSNR, SSIM, VIF, and VSNR, commonly used for image quality assessment, can be effectively applied to measure video quality by analyzing spatial and temporal characteristics. Additionally, specialized metrics like VQM, VMAF, and PEVQ are specifically designed to address the challenges unique to video data, incorporating human perceptual aspects and machine learning techniques to predict viewer perception.}

\vspace{-15pt}
\paragraph{Other Metrics for Unstructured Data}
\label{paragraph:unstructured:other}
Many metrics mentioned in the structured data section (\S\ref{subsec:structured_data}) are also applicable to measuring readiness of unstructured data. For instance, preparing unstructured data for AI applications involves adapting key dimensions typically applied to structured data (\ref{subsec:structured_data}). ``Correctness'' is fundamental, ensuring the accuracy and integrity of content within unstructured data like text, images, audio, and video to maintain AI model reliability. ``Feature Relevancy'' is crucial for identifying informative elements within unstructured data, aiding pattern recognition and decision-making. ``Privacy Leakage'' safeguards sensitive information, requiring privacy-preserving techniques. Addressing ``Class Imbalance'' and ``Class Separability'' enhances classification tasks in unstructured data by ensuring balanced representation and category distinguishability. Lastly, ``Timeliness'' is vital, particularly in dynamic data domains, as it ensures AI models stay relevant and up-to-date with evolving data patterns. 
Compliance of unstructured data with FAIR principles is a major requirement to make certain the data is ready for AI applications. All these metrics collectively contribute to unstructured data readiness for AI usage.

\begin{tcolorbox}[size=title, top=5px, bottom=5px, boxrule=0.1mm, arc=0.5mm, colback=white, colframe=black, fonttitle=\sffamily\bfseries, title=DRAI Metrics for Unstructured Data]
In examining unstructured DRAI, this study addresses both textual and multimedia domains. Textual data metrics, including lexical diversity, term importance, readability score, topic coherence, and bias indicators, are discussed with a focus on metrics highlighted in the literature. Simultaneously, multimedia DRAI metrics, such as image, speech, and video quality, are explored in the same context, featuring metrics highlighted in relevant literature. 
\end{tcolorbox}

\vspace{-10pt}


\section{Existing Frameworks}
\label{sec:frameworks}

Although not specifically targeting data readiness for AI, numerous frameworks have been developed to evaluate various aspects of data quality.
Targeting comprehensive AI readiness evaluation, we have recently developed AI Data Readiness Inspector (AIDRIN) \cite{10.1145/3676288.3676296}. We describe here various data quality evaluation frameworks along with AIDRIN. 

General data quality toolkits include frameworks like Informatica's data quality tool \cite{informatica}, an open-source solution for data profiling, cleansing, and monitoring that assesses metrics such as data completeness, accuracy, and reliability. DQLearn \cite{DQLearn-all} is another tool in this category, focusing on systematically addressing data quality issues through detection, correction, and custom rule implementation. Additionally, Deequ \cite{schelter2018automating} stands out as a library that allows unit tests for data, facilitating early data quality checks within data pipelines. Moreover, the PIQ \cite{piq} \cite{kastryulin2022piq} library offers an extensive list of metrics, including both traditional and modern techniques, for evaluating image quality. This library is a valuable resource for researchers and developers working on AI applications, allowing them to select the most appropriate metric based on their specific requirements. PIQ includes metrics categorized as full-reference, no-reference, and distribution-based metrics.

A small number of toolkits target AI data analysis. Among them, the Data Nutrition Project (DNP) Label \cite{holland2018dataset} provides a standardized format for presenting essential dataset information, including metadata, variable descriptions, and correlations, aiding in the preparation of quality training data for AI. The Data Readiness Report \cite{DRReport} focuses on generating detailed documentation to assist in data preprocessing for machine learning, offering insights into data quality and readiness across standardized dimensions. IBM's Data Quality Toolkit (DQT)\cite{gupta2021data} emphasizes automated explanations of data quality across various dimensions, including completeness, feature relevance, label purity, and data fairness, simplifying data preparation and enhancing productivity for AI practitioners.

Domain-specific toolkits are frameworks designed for specific domains or dimensions of DRAI, such as fairness and FAIR compliance. IBM's AI fairness 360 toolkit \cite{bellamy2018ai} is an open-source software aimed at detecting and mitigating biases in machine learning models, providing metrics like representation and statistical rates of sensitive attributes. For FAIR compliance, tools like FAIR Cookbook \cite{rocca-serra2023fair} and FAIRassist \cite{fairassist} guide users in implementing and measuring FAIR principles, while ESS-DIVE \cite{cholia2024essdive} evaluates FAIR compliance in earth sciences data repositories.

Toward a comprehensive evaluation of data readiness for AI, we have recently developed AIDRIN (AI Data Readiness Inspector) \cite{10.1145/3676288.3676296}. AIDRIN encompasses both traditional data quality assessments and metrics specifically designed for AI readiness, such as data bias, privacy, feature relevance, correlation, and FAIR compliance. The framework allows users to select their assessment criteria and generate intuitive visualizations and reports to enhance the interpretation and usability of results in AI-related tasks.

Despite the increasing interest in evaluating data readiness, there is still a lack of comprehensive tools covering a broad range of metrics to evaluate the readiness of structured and unstructured data for a given AI task. This is a challenging endeavor, and this survey paper will serve as a reference for understanding the available metrics and developing strategies for incorporating them into tools to comprehensively assess data readiness. 

\vspace{-10pt}

\section{Pillars of Data Readiness for AI}
\label{sec:definition}

\begin{wrapfigure}{r}{0.50\textwidth}
  \centering
  \vspace{-25pt}
  \includegraphics[width=0.50\textwidth]{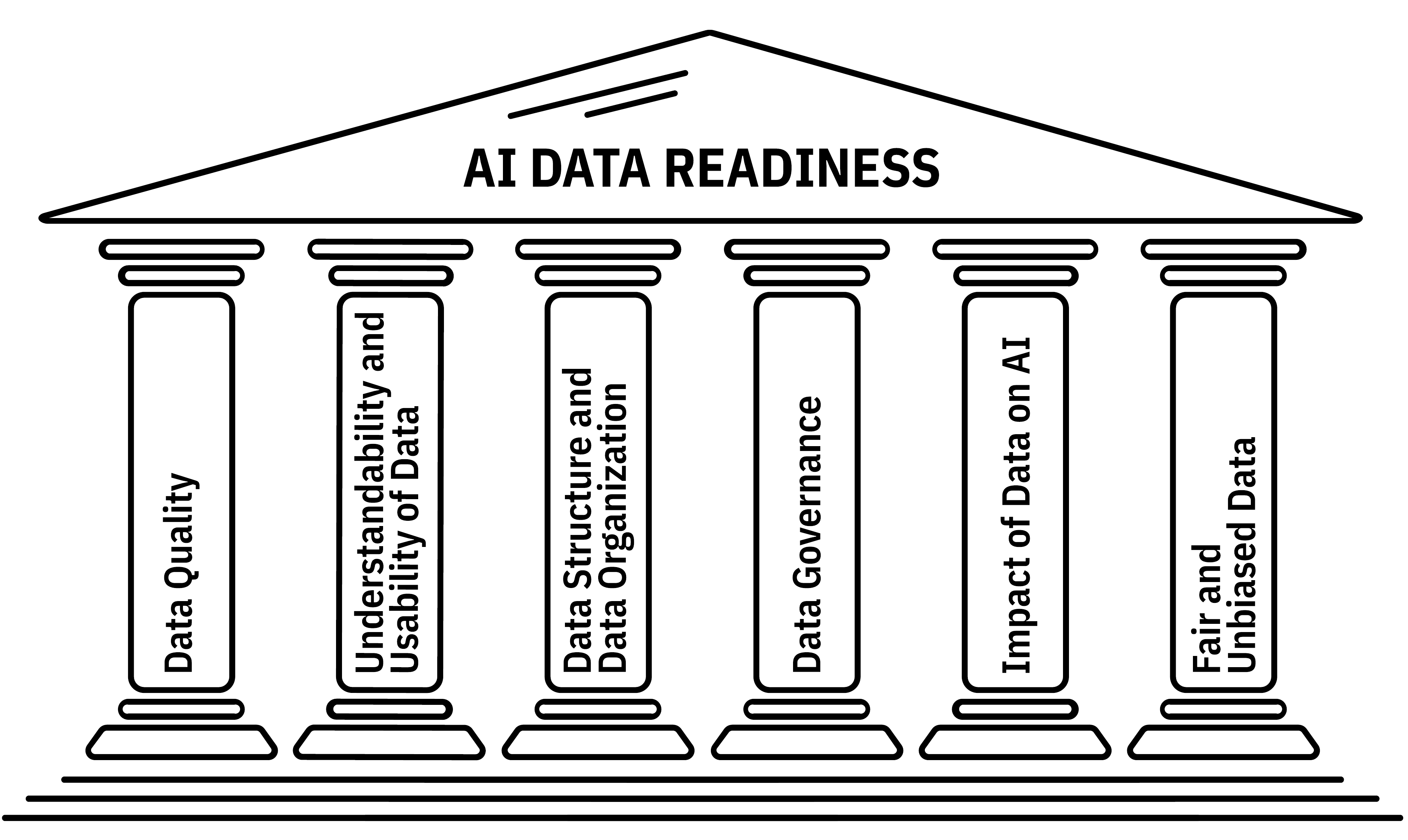}
  \caption{A high-level view of data readiness metric categories for AI.}
  \vspace{-10pt}
  \label{fig:DRAI_pillars}
\end{wrapfigure}

Based on the knowledge gathered in this survey, we propose a high level taxonomy of metrics. We categorize the AI-ready data assessment metrics into six pillars. These are data quality, understandability and usability, data structure and organization, impact of data on AI, fairness and bias, and data governance (as shown in Fig. \ref{fig:DRAI_pillars}). These pillars contain a comprehensive set of aspects in data preparation, ensuring that data is prepared and readied to support AI systems effectively, ethically, and efficiently. Each pillar is supported by specific metrics that provide a structured framework for evaluating data readiness in AI contexts.

Of these categories of metrics, the first four, i.e., data quality, understandability and usability, structure and organization, and data governance, are agnostic to the AI method. They can be applied generically to a broad set of datasets and be used for a wide range of AI applications. In contrast, the last two categories, i.e., the impact of data and fairness, are more specific to the use case and the AI methodology, making them critical for specialized contexts but requiring tailored approaches.

\textbf{Data quality} ensures that the data used to train AI models is accurate, complete, and consistent. High-quality data minimizes the risk of errors in AI outputs, leading to more reliable and trustworthy models. When data quality is compromised, it can result in inaccurate and unstable models. Thus, maintaining rigorous data quality standards is essential for achieving effective and credible AI outcomes. Structured data quality can be evaluated using metrics described in Section \ref{sec:drai_metrics}, such as completeness, correctness, timeliness, mislabeling, multimedia quality. 


\textbf{Data understanding and usability} are important for enabling AI systems to interpret and utilize data effectively. This category of metrics emphasize the importance of clear documentation, comprehensive metadata, reusability, and accessible data interfaces. When data is well-understood and easy to use, it accelerates AI model development.
FAIR principle compliance metrics[\ref{subsubsection:FAIRness_score}] can serve in evaluating data understanding and usability.

\textbf{Data structure and data organization} are important to integrate data into AI workflows seamlessly and efficiently. 
Adequate number of samples in data and proper data partitioning into training, testing, and validation sets allow for accurate model evaluation. In addition, the data model, i.e., the schema of the data, and data organization, i.e., how the data is stored, also play a role in the speed of AI training. also plays a role in the performance of AI applications. 

\textbf{Data Governance} is essential for managing data in a way that is ethical, secure, and compliant with legal standards. This pillar covers key aspects such as data privacy, security, and the ethical use of data, which are necessary for building trust in AI systems. 
Without proper governance, AI systems risk violating privacy regulations, facing security breaches, and engaging in unethical practices, which can harm public trust and lead to significant legal and reputation-related consequences. Metrics such as privacy leakage[\ref{subsubsection:privacy_leakage}] are essential for understanding the extent of potential privacy risks within the data.

The \textbf{impact of data on AI} covers the importance of data content and its relevance to AI applications. Rich and high-impact data that provides meaningful insights is critical for deriving effective AI outcomes by enabling models to make accurate predictions and identify deep patterns. 
Feature relevance[\ref{subsubsection:feature_relevance}] and data point impact[\ref{subsubsection:data_point_impact}] serve as quantitative measures to assess the value of data for a given AI application.

\textbf{Fair and unbiased data} is a fundamental aspect of ensuring that AI systems produce equitable and unbiased outcomes. This pillar focuses on the representativeness of the data and the absence of biases that could lead to discriminatory practices. Fairness in AI models is not only an ethical concern but also crucial for maintaining public trust in AI technologies. When data used in AI models is biased or unrepresentative, it can result in skewed outcomes that may continue existing inequality and injustice. This undermines the societal benefits that AI promises to deliver. Metrics such as the discrimination (bias) index[\ref{subsubsection:discrimination_index}], class imbalance[\ref{subsubsection:class_imbalance}], and class separability[\ref{subsubsection:class_separability}] are crucial for assessing the fairness of data before its use in AI applications.

The six dimensions outlined above can be quantified using specific metrics discussed in this study to evaluate DRAI. While these metrics provide a foundational assessment, additional metrics are required to fully capture the breadth of each dimension. Aggregating the evaluations across the existing and future metrics will result in a comprehensive DRAI assessment that would lead to highly accurate and impactful AI solutions. 

\vspace{-10pt}

\section{Gaps and Challenges}
\label{sec:gaps-challenges}

We discuss the challenges in defining the metrics for assessing data readiness for AI in structured and unstructured data. While structured data poses unique challenges regarding standardization, interpretability, and  sensitivity, unstructured datasets present additional complexities due to their diverse formats, varying modalities, and contextual nuances. 

Assessing data readiness for AI and data science applications, regardless of its structure, presents several  challenges stemming from the absence of a unified framework that complicates the comparison and consistent application of  metrics across diverse dimensions. This absence hampers the development of a cohesive and comprehensive data readiness assessment method explicitly tailored for different types of data. Although IBM DQT \cite{gupta2021data} and AIDRIN \cite{10.1145/3676288.3676296} have taken the initial steps in addressing this concern, their coverage is primarily focused on structured data, leaving a need for further advancements to include a broader range of dimensions related to structured data readiness and extend the toolkit's capabilities to address unstructured data challenges. Additionally, with the rapid growth of large datasets, there is a lack of parallel systems designed to evaluate data readiness effectively at scale. This highlights the need for further advancements to develop robust methods for handling diverse and extensive data environments.

A significant challenge in DRAI assessment is the evolving nature of data quality dimensions. As new data quality metrics emerge, they add complexity to the evaluation process. According to Batini and Scannapieco's \cite{batini2006data}, the development of metrics specific to various domains, such as the condition of archival documents, the integrity of statistical data, and the positional accuracy in geospatial data, further complicates comprehensive assessments of data quality and readiness. This ongoing evolution necessitates continual adaptation and refinement of evaluation frameworks to effectively address established and new metrics.

In the rapidly evolving fields of AI and data science, the emergence of new use cases and diverse data structures constantly challenges the evaluation of data readiness. To ensure these metrics remain effective in assessing data preparedness for the latest AI applications, they must adapt and evolve alongside the technology. Striking the right balance between data readiness and quantity is another crucial challenge. While numerous metrics focus on data readiness, a comprehensive approach should consider the trade-off between data readiness and sufficiency. Sufficient data volumes are essential for meaningful analysis and effective AI model training, making finding the optimal equilibrium between data readiness and quantity a challenge. Additionally, clear interpretability of these readiness metrics is essential for stakeholders to grasp their implications on overall data readiness and their potential impacts on AI model performance. Enhanced visualization and explanation techniques can significantly improve the practical utility of these metrics, facilitating more informed decision-making processes.

Addressing the challenges in the dynamic field of AI and data readiness is important. Specific AI applications often require customized data readiness metrics due to varied data readiness standards, requiring domain-specific expertise for effective navigation. Simultaneously, addressing subjective judgments and human biases, particularly concerning fairness and privacy, adds another layer of complexity. Developing unbiased and ethical metrics that adapt to various data types and applications requires ongoing research and innovation. Furthermore, the ever-evolving nature of real-world datasets requires continuous data readiness assessment throughout an AI system's life cycle, extending beyond the initial data preparation phase.

Data access and ownership concerns can impede data readiness evaluation, particularly when datasets are restricted due to privacy and ownership issues. These limitations can delay a comprehensive data readiness assessment, requiring collaborative efforts and agreements between data providers and users to navigate these challenges effectively. Furthermore, the ongoing challenge lies in establishing meaningful thresholds that define acceptable data readiness levels. Given the context-dependent nature of data and the diversity of AI use cases, universal and context-independent threshold values are challenging to determine. Clear guidelines regarding data readiness thresholds are essential to ensure consistent and effective data preparation practices. Lastly, developing well-established benchmark datasets and evaluation protocols becomes crucial to compare and evaluate various data readiness metrics. Creating representative benchmarks spanning different industries and data structures can facilitate a more comprehensive comparison of diverse metrics and scoring mechanisms.

Data readiness assessment is critical for all AI applications, including large language models (LLMs)\cite{telmai2023}. Ensuring data completeness, accuracy, and consistency, these metrics will build a robust foundation for LLMs. Correct, unbiased, and relevant data enhance the model's ability to generate coherent and reliable outputs. Assessing bias in datasets ensures fairness and mitigates skewed predictions. Therefore, applying comprehensive quality metrics is vital to preparing data that effectively supports the latest requirements of LLMs, leading to more accurate and contextually relevant model performance. These requirements highlight the need for more advanced metrics and tools to effectively prepare data to ensure LLMs are trained on high-quality inputs that meet the requirements of modern AI applications.

While having a comprehensive list of metrics in the toolbox is important, deciding the metrics to be used for a particular AI application is essential. When determining the suitable metrics for an AI application, it is crucial to start by defining the objective of the application and being aware of any constraints including data limits or legal considerations. Once the application is defined, one should explore the available metrics and understand each metric and its suitability to achieve the goals. Thet should consider the trade-offs involved and analyze which metrics best align with your objectives. 
Focusing on metrics that impact the AI application performance and efficiency achieves a balance between simplicity and depth. Ensuring that the AI model's performance is regularly monitored and that the metrics are flexibly adjusted as it evolves also plays a role in using the correct data. Additionally, reviewing relevant literature can provide valuable insights in selecting the most appropriate metrics. For example, \citet{10.1145/3168389} proposed a method for selecting among over $80$ privacy metrics.
The selected metrics assess factors such as data sensitivity, trade-offs, and performance expectations, providing meaningful insights and driving the application toward its intended outcomes.

Addressing these gaps and challenges in data readiness metrics urges collaboration among researchers, practitioners, and industry experts. Advancing the state-of-the-art in these metrics will contribute to more reliable utilization of data in AI applications, unlocking the maximum potential of this valuable resource.

\vspace{-10pt}

\section{Conclusion}
\label{sec:conclusion}

This comprehensive survey underscores the critical role of data preparation toward the effective usage of data by AI applications. We explored the challenges, tools, and metrics associated with data readiness, emphasizing its significance in achieving accurate and dependable AI-driven outcomes. Our study highlights the need for a holistic approach, covering structured and unstructured data, and underscores the importance of incorporating fairness-related metrics to ensure unbiased AI decision-making. We have identified and showcased existing metrics and scoring mechanisms that effectively measure data readiness available in literature by studying more than $140$ publications from reputable journals including ACM, IEEE, and other reputable journals, as well as web articles, spanning the past three decades. By thoroughly exploring these dimensions and metrics, we have summarized numerous data readiness metrics and proposed a taxonomy of them. This will develop a deeper understanding of data readiness for AI applications. As AI advances and data becomes an even more critical asset, staying up to date with the latest research and advancements in data readiness metrics is essential. This survey provides a foundational resource for researchers and practitioners, equipping them with the essential insights needed to navigate the complexities of data preparation for AI effectively and emphasizing that data readiness is not just a preliminary step but an ongoing commitment.

\vspace{-10pt}


\bibliographystyle{ACM-Reference-Format}
\bibliography{main}


\begin{thebibliography}{151}


\ifx \showCODEN    \undefined \def \showCODEN     #1{\unskip}     \fi
\ifx \showDOI      \undefined \def \showDOI       #1{#1}\fi
\ifx \showISBNx    \undefined \def \showISBNx     #1{\unskip}     \fi
\ifx \showISBNxiii \undefined \def \showISBNxiii  #1{\unskip}     \fi
\ifx \showISSN     \undefined \def \showISSN      #1{\unskip}     \fi
\ifx \showLCCN     \undefined \def \showLCCN      #1{\unskip}     \fi
\ifx \shownote     \undefined \def \shownote      #1{#1}          \fi
\ifx \showarticletitle \undefined \def \showarticletitle #1{#1}   \fi
\ifx \showURL      \undefined \def \showURL       {\relax}        \fi
\providecommand\bibfield[2]{#2}
\providecommand\bibinfo[2]{#2}
\providecommand\natexlab[1]{#1}
\providecommand\showeprint[2][]{arXiv:#2}

\bibitem[PEV(2008)]%
        {PEVQ}
 \bibinfo{year}{2008}\natexlab{}.
\newblock \bibinfo{title}{PEVQ – the Standard for Perceptual Evaluation of Video Quality}.
\newblock
\newblock
\urldef\tempurl%
\url{http://www.pevq.com/pevq.html}
\showURL{%
\tempurl}
\newblock
\shownote{Accessed 22 July 2023}.


\bibitem[FAI(2024)]%
        {FAIR:Principles}
 \bibinfo{year}{2024}\natexlab{}.
\newblock \bibinfo{title}{{FAIR Principles}}.
\newblock \bibinfo{howpublished}{\url{https://www.go-fair.org/fair-principles/}}.
\newblock


\bibitem[ain(nd)]%
        {aindoprivacy}
 \bibinfo{year}{n.d.}\natexlab{}.
\newblock
\newblock
\urldef\tempurl%
\url{https://docs.aindo.com/evaluation/privacy/}
\showURL{%
\tempurl}


\bibitem[itu(nd)]%
        {itu_bs1387}
 \bibinfo{year}{n.d.}\natexlab{}.
\newblock \bibinfo{title}{{BS.1387 : Method for Objective Measurements of Perceived Audio Quality}}.
\newblock \bibinfo{howpublished}{\url{https://www.itu.int/rec/R-REC-BS.1387/en}}.
\newblock
\newblock
\shownote{Accessed 17 July 2023}.


\bibitem[rea(nd)]%
        {readable_gf}
 \bibinfo{year}{n.d}\natexlab{}.
\newblock \bibinfo{title}{The Gunning Fog Index}.
\newblock \bibinfo{howpublished}{\url{https://readable.com/readability/gunning-fog-index/}}.
\newblock
\newblock
\shownote{Accessed 12 July 2023}.


\bibitem[sar(nd)]%
        {sarnoff_jndmetrix}
 \bibinfo{year}{n.d.}\natexlab{}.
\newblock \bibinfo{title}{JNDmetrix Technology}.
\newblock \bibinfo{howpublished}{\url{http://www.sarnoff.com/products_services/video_vision/jndmetrix/}}.
\newblock
\newblock
\shownote{Accessed 12 July 2023}.


\bibitem[Kag(nd)]%
        {Kaggle}
 \bibinfo{year}{n.d.}\natexlab{}.
\newblock \bibinfo{title}{Kaggle}.
\newblock
\newblock
\urldef\tempurl%
\url{https://www.kaggle.com/}
\showURL{%
\tempurl}
\newblock
\shownote{Accessed: Sept 2023}.


\bibitem[psn(nd)]%
        {psnr}
 \bibinfo{year}{n.d.}\natexlab{}.
\newblock \bibinfo{title}{PSNR}.
\newblock
\newblock
\urldef\tempurl%
\url{https://www.mathworks.com/help/vision/ref/psnr.html}
\showURL{%
\tempurl}
\newblock
\shownote{Data Accessed 7/26/2023}.


\bibitem[Afendras and Markatou(2019)]%
        {AFENDRAS2019286}
\bibfield{author}{\bibinfo{person}{Georgios Afendras} {and} \bibinfo{person}{Marianthi Markatou}.} \bibinfo{year}{2019}\natexlab{}.
\newblock \showarticletitle{Optimality of training/test size and resampling effectiveness in cross-validation}.
\newblock \bibinfo{journal}{\emph{Journal of Statistical Planning and Inference}}  \bibinfo{volume}{199} (\bibinfo{year}{2019}), \bibinfo{pages}{286--301}.
\newblock
\showISSN{0378-3758}
\urldef\tempurl%
\url{https://doi.org/10.1016/j.jspi.2018.07.005}
\showDOI{\tempurl}


\bibitem[Afzal et~al\mbox{.}(2020)]%
        {DRReport}
\bibfield{author}{\bibinfo{person}{S. Afzal}, \bibinfo{person}{C. Rajmohan}, \bibinfo{person}{M. Kesarwani}, \bibinfo{person}{S. Mehta}, {and} \bibinfo{person}{H. Patel}.} \bibinfo{year}{2020}\natexlab{}.
\newblock \showarticletitle{Data Readiness Report}. In \bibinfo{booktitle}{\emph{{IEEE} Int. Conference on Smart Data Services (SMDS)}}.
\newblock


\bibitem[AI(2023)]%
        {telmai2023}
\bibfield{author}{\bibinfo{person}{Telm AI}.} \bibinfo{year}{2023}\natexlab{}.
\newblock \bibinfo{title}{Demystifying Data Quality's Impact on Large Language Models}.
\newblock \bibinfo{howpublished}{\url{https://www.telm.ai/blog/demystifying-data-qualitys-impact-on-large-language-models/}}.
\newblock
\newblock
\shownote{Accessed: [Your Access Date]}.


\bibitem[Alberto et~al\mbox{.}(2018)]%
        {alberto2018learning}
\bibfield{author}{\bibinfo{person}{Francisco Alberto}, \bibinfo{person}{Salvador García}, \bibinfo{person}{Mikel Galar}, \bibinfo{person}{Ronaldo Prati}, \bibinfo{person}{Bartosz Krawczyk}, {and} \bibinfo{person}{Francisco Herrera}.} \bibinfo{year}{2018}\natexlab{}.
\newblock \bibinfo{booktitle}{\emph{{Learning from Imbalanced Data Sets}}}.
\newblock \bibinfo{publisher}{Springer}.
\newblock


\bibitem[Alwosheel et~al\mbox{.}(2018)]%
        {ALWOSHEEL2018167}
\bibfield{author}{\bibinfo{person}{Ahmad Alwosheel}, \bibinfo{person}{Sander {van Cranenburgh}}, {and} \bibinfo{person}{Caspar~G. Chorus}.} \bibinfo{year}{2018}\natexlab{}.
\newblock \showarticletitle{Is your dataset big enough? Sample size requirements when using artificial neural networks for discrete choice analysis}.
\newblock \bibinfo{journal}{\emph{Journal of Choice Modelling}}  \bibinfo{volume}{28} (\bibinfo{year}{2018}), \bibinfo{pages}{167--182}.
\newblock
\showISSN{1755-5345}
\urldef\tempurl%
\url{https://doi.org/10.1016/j.jocm.2018.07.002}
\showDOI{\tempurl}


\bibitem[Arca and Hewett(2020)]%
        {9458108}
\bibfield{author}{\bibinfo{person}{Sevgi Arca} {and} \bibinfo{person}{Rattikorn Hewett}.} \bibinfo{year}{2020}\natexlab{}.
\newblock \showarticletitle{Is Entropy enough for measuring Privacy?}. In \bibinfo{booktitle}{\emph{2020 International Conference on Computational Science and Computational Intelligence (CSCI)}}. \bibinfo{pages}{1335--1340}.
\newblock
\urldef\tempurl%
\url{https://doi.org/10.1109/CSCI51800.2020.00249}
\showDOI{\tempurl}


\bibitem[Azzalini et~al\mbox{.}(2022)]%
        {10.1145/3552433}
\bibfield{author}{\bibinfo{person}{Fabio Azzalini}, \bibinfo{person}{Chiara Criscuolo}, {and} \bibinfo{person}{Letizia Tanca}.} \bibinfo{year}{2022}\natexlab{}.
\newblock \showarticletitle{E-FAIR-DB: Functional Dependencies to Discover Data Bias and Enhance Data Equity}.
\newblock \bibinfo{journal}{\emph{J. Data and Information Quality}} \bibinfo{volume}{14}, \bibinfo{number}{4}, Article \bibinfo{articleno}{29} (\bibinfo{date}{nov} \bibinfo{year}{2022}), \bibinfo{numpages}{26}~pages.
\newblock
\showISSN{1936-1955}
\urldef\tempurl%
\url{https://doi.org/10.1145/3552433}
\showDOI{\tempurl}


\bibitem[Bachem et~al\mbox{.}(2017)]%
        {bachem2017coreset}
\bibfield{author}{\bibinfo{person}{Olivier Bachem}, \bibinfo{person}{Mario Lucic}, {and} \bibinfo{person}{Andreas Krause}.} \bibinfo{year}{2017}\natexlab{}.
\newblock \showarticletitle{Practical Coreset Constructions for Machine Learning}. In \bibinfo{booktitle}{\emph{Advances in Neural Information Processing Systems}}.
\newblock


\bibitem[Batini and Scannapieco(2006)]%
        {batini2006data}
\bibfield{author}{\bibinfo{person}{Carlo Batini} {and} \bibinfo{person}{Monica Scannapieco}.} \bibinfo{year}{2006}\natexlab{}.
\newblock \bibinfo{booktitle}{\emph{Data Quality: Concepts, Methodologies and Techniques} (\bibinfo{edition}{1} ed.)}.
\newblock \bibinfo{publisher}{Springer-Verlag Berlin Heidelberg}, \bibinfo{address}{Berlin, Heidelberg}. XIX, 262 pages.
\newblock
\showISBNx{978-3-540-33172-8}
\urldef\tempurl%
\url{https://doi.org/10.1007/3-540-33173-5}
\showDOI{\tempurl}


\bibitem[Beerends and Stemerdink(1994)]%
        {beerends1994perceptual}
\bibfield{author}{\bibinfo{person}{J. Beerends} {and} \bibinfo{person}{J. Stemerdink}.} \bibinfo{year}{1994}\natexlab{}.
\newblock \showarticletitle{A Perceptual Speech Quality Measure Based on a Psychoacoustic Sound Representation}.
\newblock \bibinfo{journal}{\emph{Journal of Audio Eng. Soc.}}  \bibinfo{volume}{42} (\bibinfo{date}{December} \bibinfo{year}{1994}), \bibinfo{pages}{115--123}.
\newblock


\bibitem[Bezzi(2007)]%
        {4550303}
\bibfield{author}{\bibinfo{person}{Michele Bezzi}.} \bibinfo{year}{2007}\natexlab{}.
\newblock \showarticletitle{An entropy based method for measuring anonymity}. In \bibinfo{booktitle}{\emph{2007 Third International Conference on Security and Privacy in Communications Networks and the Workshops - SecureComm 2007}}. \bibinfo{pages}{28--32}.
\newblock
\urldef\tempurl%
\url{https://doi.org/10.1109/SECCOM.2007.4550303}
\showDOI{\tempurl}


\bibitem[Blaiszik et~al\mbox{.}(2016)]%
        {blaiszik2016materialdatafacility}
\bibfield{author}{\bibinfo{person}{B. Blaiszik} {et~al\mbox{.}}} \bibinfo{year}{2016}\natexlab{}.
\newblock \showarticletitle{The Materials Data Facility: Data Services to Advance Materials Science Research}.
\newblock \bibinfo{journal}{\emph{JOM}}  \bibinfo{volume}{68} (\bibinfo{year}{2016}).
\newblock
\urldef\tempurl%
\url{https://doi.org/10.1007/s11837-016-2001-3}
\showDOI{\tempurl}


\bibitem[Blake and Mangiameli(2011)]%
        {10.1145/1891879.1891881}
\bibfield{author}{\bibinfo{person}{Roger Blake} {and} \bibinfo{person}{Paul Mangiameli}.} \bibinfo{year}{2011}\natexlab{}.
\newblock \showarticletitle{The Effects and Interactions of Data Quality and Problem Complexity on Classification}.
\newblock \bibinfo{journal}{\emph{J. Data and Information Quality}} \bibinfo{volume}{2}, \bibinfo{number}{2}, Article \bibinfo{articleno}{8} (\bibinfo{date}{feb} \bibinfo{year}{2011}), \bibinfo{numpages}{28}~pages.
\newblock
\showISSN{1936-1955}
\urldef\tempurl%
\url{https://doi.org/10.1145/1891879.1891881}
\showDOI{\tempurl}


\bibitem[Blei et~al\mbox{.}(2003)]%
        {blei2003latent}
\bibfield{author}{\bibinfo{person}{David~M. Blei}, \bibinfo{person}{Andrew~Y. Ng}, {and} \bibinfo{person}{Michael~I. Jordan}.} \bibinfo{year}{2003}\natexlab{}.
\newblock \showarticletitle{Latent Dirichlet Allocation}.
\newblock \bibinfo{journal}{\emph{Journal of Machine Learning Research}}  \bibinfo{volume}{3} (\bibinfo{date}{January} \bibinfo{year}{2003}), \bibinfo{pages}{993--1022}.
\newblock
\newblock
\shownote{Submitted 2/02; Published 1/03}.


\bibitem[Blog(2017)]%
        {Blog_2017}
\bibfield{author}{\bibinfo{person}{Netflix~Technology Blog}.} \bibinfo{year}{2017}\natexlab{}.
\newblock \bibinfo{title}{Toward a practical perceptual video quality metric}.
\newblock
\newblock
\urldef\tempurl%
\url{https://netflixtechblog.com/toward-a-practical-perceptual-video-quality-metric-653f208b9652}
\showURL{%
\tempurl}


\bibitem[Bors et~al\mbox{.}(2018)]%
        {10.1145/3190578}
\bibfield{author}{\bibinfo{person}{Christian Bors}, \bibinfo{person}{Theresia Gschwandtner}, \bibinfo{person}{Simone Kriglstein}, \bibinfo{person}{Silvia Miksch}, {and} \bibinfo{person}{Margit Pohl}.} \bibinfo{year}{2018}\natexlab{}.
\newblock \showarticletitle{Visual Interactive Creation, Customization, and Analysis of Data Quality Metrics}.
\newblock \bibinfo{journal}{\emph{J. Data and Information Quality}} \bibinfo{volume}{10}, \bibinfo{number}{1}, Article \bibinfo{articleno}{3} (\bibinfo{date}{may} \bibinfo{year}{2018}), \bibinfo{numpages}{26}~pages.
\newblock
\showISSN{1936-1955}
\urldef\tempurl%
\url{https://doi.org/10.1145/3190578}
\showDOI{\tempurl}


\bibitem[Borsos et~al\mbox{.}(2018)]%
        {borsos2018dealing}
\bibfield{author}{\bibinfo{person}{Z. Borsos}, \bibinfo{person}{C. Lemnaru}, {and} \bibinfo{person}{R. Potolea}.} \bibinfo{year}{2018}\natexlab{}.
\newblock \showarticletitle{Dealing with overlap and imbalance: a new metric and approach}.
\newblock \bibinfo{journal}{\emph{Pattern Anal Applic}} \bibinfo{volume}{21}, \bibinfo{number}{2} (\bibinfo{year}{2018}), \bibinfo{pages}{381--395}.
\newblock
\urldef\tempurl%
\url{https://doi.org/10.1007/s10044-016-0583-6}
\showDOI{\tempurl}


\bibitem[Breunig et~al\mbox{.}(2000)]%
        {breunig2000lof}
\bibfield{author}{\bibinfo{person}{Markus~M. Breunig}, \bibinfo{person}{Hans-Peter Kriegel}, \bibinfo{person}{Raymond~T. Ng}, {and} \bibinfo{person}{Jörg Sander}.} \bibinfo{year}{2000}\natexlab{}.
\newblock \showarticletitle{{LOF: Identifying density-based local outliers}}. In \bibinfo{booktitle}{\emph{Proc. ACM SIGMOD Int. Conf. Manage. Data}}. \bibinfo{pages}{93--104}.
\newblock


\bibitem[Carlini et~al\mbox{.}(2022)]%
        {carlini2022privacy}
\bibfield{author}{\bibinfo{person}{Nicholas Carlini}, \bibinfo{person}{Matthew Jagielski}, \bibinfo{person}{Chiyuan Zhang}, \bibinfo{person}{Nicolas Papernot}, \bibinfo{person}{Andreas Terzis}, {and} \bibinfo{person}{Florian Tramer}.} \bibinfo{year}{2022}\natexlab{}.
\newblock \bibinfo{title}{The Privacy Onion Effect: Memorization is Relative}.
\newblock
\newblock
\showeprint[arxiv]{2206.10469}~[cs.LG]


\bibitem[Celis et~al\mbox{.}(2020)]%
        {celis2020data}
\bibfield{author}{\bibinfo{person}{L.~Elisa Celis}, \bibinfo{person}{Vijay Keswani}, {and} \bibinfo{person}{Nisheeth~K. Vishnoi}.} \bibinfo{year}{2020}\natexlab{}.
\newblock \bibinfo{title}{Data preprocessing to mitigate bias: A maximum entropy based approach}.
\newblock
\newblock
\showeprint[arxiv]{1906.02164}~[cs.LG]


\bibitem[Chandler and Hemami(2007)]%
        {4286985}
\bibfield{author}{\bibinfo{person}{Damon~M. Chandler} {and} \bibinfo{person}{Sheila~S. Hemami}.} \bibinfo{year}{2007}\natexlab{}.
\newblock \showarticletitle{VSNR: A Wavelet-Based Visual Signal-to-Noise Ratio for Natural Images}.
\newblock \bibinfo{journal}{\emph{IEEE Transactions on Image Processing}} \bibinfo{volume}{16}, \bibinfo{number}{9} (\bibinfo{year}{2007}), \bibinfo{pages}{2284--2298}.
\newblock
\urldef\tempurl%
\url{https://doi.org/10.1109/TIP.2007.901820}
\showDOI{\tempurl}


\bibitem[Clarke~et al.(2019)]%
        {clarke2019fairshake}
\bibfield{author}{\bibinfo{person}{Daniel~JB Clarke~et al.}} \bibinfo{year}{2019}\natexlab{}.
\newblock \showarticletitle{FAIRshake: toolkit to evaluate the FAIRness of research digital resources}.
\newblock \bibinfo{journal}{\emph{Cell systems}} \bibinfo{volume}{9}, \bibinfo{number}{5} (\bibinfo{year}{2019}), \bibinfo{pages}{417--421}.
\newblock


\bibitem[{Cleanlab}(2024)]%
        {cleanlab2024elevating}
\bibfield{author}{\bibinfo{person}{{Cleanlab}}.} \bibinfo{year}{2024}\natexlab{}.
\newblock \bibinfo{title}{Elevating Data Quality: The Crucial Role of Proper Data Annotation}.
\newblock \bibinfo{howpublished}{\url{https://cleanlab.ai/blog/learn/data-annotation/}}.
\newblock
\urldef\tempurl%
\url{https://cleanlab.ai/blog/learn/data-annotation/}
\showURL{%
\tempurl}
\newblock
\shownote{Accessed: 2024-08-06}.


\bibitem[Cohen(1960)]%
        {cohen_coefficient_1960}
\bibfield{author}{\bibinfo{person}{J. Cohen}.} \bibinfo{year}{1960}\natexlab{}.
\newblock \showarticletitle{A Coefficient of Agreement for Nominal Scales}.
\newblock \bibinfo{journal}{\emph{Educational and Psychological Measurement}} \bibinfo{volume}{20}, \bibinfo{number}{1} (\bibinfo{year}{1960}), \bibinfo{pages}{37--46}.
\newblock
\showISSN{0013-1644}
\urldef\tempurl%
\url{https://doi.org/10.1177/001316446002000104}
\showDOI{\tempurl}


\bibitem[Coleman and Liau(1975)]%
        {Coleman1975ACR}
\bibfield{author}{\bibinfo{person}{Meri Coleman} {and} \bibinfo{person}{Ta~Lin Liau}.} \bibinfo{year}{1975}\natexlab{}.
\newblock \showarticletitle{A computer readability formula designed for machine scoring.}
\newblock \bibinfo{journal}{\emph{J. of Applied Psychology}}  \bibinfo{volume}{60} (\bibinfo{year}{1975}), \bibinfo{pages}{283--284}.
\newblock


\bibitem[Cook and Weisberg(1982)]%
        {cook1982residuals}
\bibfield{author}{\bibinfo{person}{R.~Dennis Cook} {and} \bibinfo{person}{Sanford Weisberg}.} \bibinfo{year}{1982}\natexlab{}.
\newblock \bibinfo{booktitle}{\emph{Residuals and Influence in Regression}}.
\newblock \bibinfo{publisher}{Chapman \& Hall}.
\newblock


\bibitem[Dai et~al\mbox{.}(2006)]%
        {4053044}
\bibfield{author}{\bibinfo{person}{Bing~Tian Dai}, \bibinfo{person}{Nick Koudas}, \bibinfo{person}{Beng~Chin Ooi}, \bibinfo{person}{Divesh Srivastava}, {and} \bibinfo{person}{Suresh Venkatasubramanian}.} \bibinfo{year}{2006}\natexlab{}.
\newblock \showarticletitle{Rapid Identification of Column Heterogeneity}. In \bibinfo{booktitle}{\emph{Sixth International Conference on Data Mining (ICDM'06)}}. \bibinfo{pages}{159--170}.
\newblock
\urldef\tempurl%
\url{https://doi.org/10.1109/ICDM.2006.132}
\showDOI{\tempurl}


\bibitem[Daubechies(1992)]%
        {daubechies1992}
\bibfield{author}{\bibinfo{person}{Ingrid Daubechies}.} \bibinfo{year}{1992}\natexlab{}.
\newblock \bibinfo{booktitle}{\emph{Ten Lectures on Wavelets}}.
\newblock \bibinfo{publisher}{Society for Industrial and Applied Mathematics}.
\newblock


\bibitem[Davis and Sampson(1986)]%
        {davis1986statistics}
\bibfield{author}{\bibinfo{person}{John~C. Davis} {and} \bibinfo{person}{Robert~J. Sampson}.} \bibinfo{year}{1986}\natexlab{}.
\newblock \bibinfo{booktitle}{\emph{{Statistics and Data Analysis in Geology}}}. Vol.~\bibinfo{volume}{646}.
\newblock \bibinfo{publisher}{Wiley}, \bibinfo{address}{New York}.
\newblock


\bibitem[Degirmenci and Karal(2021)]%
        {9627977}
\bibfield{author}{\bibinfo{person}{Ali Degirmenci} {and} \bibinfo{person}{Omer Karal}.} \bibinfo{year}{2021}\natexlab{}.
\newblock \showarticletitle{Robust Incremental Outlier Detection Approach Based on a New Metric in Data Streams}.
\newblock \bibinfo{journal}{\emph{IEEE Access}}  \bibinfo{volume}{9} (\bibinfo{year}{2021}), \bibinfo{pages}{160347--160360}.
\newblock
\urldef\tempurl%
\url{https://doi.org/10.1109/ACCESS.2021.3131402}
\showDOI{\tempurl}


\bibitem[Developer(2021)]%
        {IBM_DQT}
\bibfield{author}{\bibinfo{person}{IBM Developer}.} \bibinfo{year}{2021}\natexlab{}.
\newblock \bibinfo{title}{IBM Data Quality AI Toolkit}.
\newblock
\newblock
\urldef\tempurl%
\url{https://developer.ibm.com/learningpaths/data-quality-ai-toolkit/overview/}
\showURL{%
\tempurl}
\newblock
\shownote{Date accessed: June 12, 2023}.


\bibitem[Dua and Graff(2017)]%
        {dua2017uci}
\bibfield{author}{\bibinfo{person}{Dheeru Dua} {and} \bibinfo{person}{Casey Graff}.} \bibinfo{year}{2017}\natexlab{}.
\newblock \bibinfo{title}{{UCI} Machine Learning Repository}.
\newblock \bibinfo{howpublished}{\url{http://archive.ics.uci.edu/ml}}.
\newblock


\bibitem[Duda et~al\mbox{.}(2012)]%
        {duda2012pattern}
\bibfield{author}{\bibinfo{person}{Richard~O. Duda}, \bibinfo{person}{Peter~E. Hart}, {and} \bibinfo{person}{David~G. Stork}.} \bibinfo{year}{2012}\natexlab{}.
\newblock \bibinfo{booktitle}{\emph{{Pattern Classification}}}.
\newblock \bibinfo{publisher}{John Wiley \& Sons}.
\newblock


\bibitem[Duddu et~al\mbox{.}(2022)]%
        {duddu2022shapr}
\bibfield{author}{\bibinfo{person}{Vasisht Duddu}, \bibinfo{person}{Sebastian Szyller}, {and} \bibinfo{person}{N. Asokan}.} \bibinfo{year}{2022}\natexlab{}.
\newblock \bibinfo{title}{SHAPr: An Efficient and Versatile Membership Privacy Risk Metric for Machine Learning}.
\newblock
\newblock
\showeprint[arxiv]{2112.02230}~[cs.CR]


\bibitem[Elmagarmid et~al\mbox{.}(2007)]%
        {4016511}
\bibfield{author}{\bibinfo{person}{Ahmed~K. Elmagarmid}, \bibinfo{person}{Panagiotis~G. Ipeirotis}, {and} \bibinfo{person}{Vassilios~S. Verykios}.} \bibinfo{year}{2007}\natexlab{}.
\newblock \showarticletitle{Duplicate Record Detection: A Survey}.
\newblock \bibinfo{journal}{\emph{IEEE Transactions on Knowledge and Data Engineering}} \bibinfo{volume}{19}, \bibinfo{number}{1} (\bibinfo{year}{2007}), \bibinfo{pages}{1--16}.
\newblock
\urldef\tempurl%
\url{https://doi.org/10.1109/TKDE.2007.250581}
\showDOI{\tempurl}


\bibitem[et~al.(2021)]%
        {gupta2021data}
\bibfield{author}{\bibinfo{person}{Nitin~Gupta et al.}} \bibinfo{year}{2021}\natexlab{}.
\newblock \bibinfo{title}{Data Quality Toolkit: Automatic assessment of data quality and remediation for machine learning datasets}.
\newblock
\newblock
\showeprint[arxiv]{2108.05935}~[cs.LG]


\bibitem[et~al.(2022)]%
        {Ravi_2022}
\bibfield{author}{\bibinfo{person}{Nikil~Ravi et al.}} \bibinfo{year}{2022}\natexlab{}.
\newblock \showarticletitle{{FAIR} principles for {AI} models with a practical application for accelerated high energy diffraction microscopy}.
\newblock \bibinfo{journal}{\emph{Scientific Data}} \bibinfo{volume}{9}, \bibinfo{number}{1} (\bibinfo{date}{nov} \bibinfo{year}{2022}).
\newblock
\urldef\tempurl%
\url{https://doi.org/10.1038/s41597-022-01712-9}
\showDOI{\tempurl}


\bibitem[et~al.(2018)]%
        {bellamy2018ai}
\bibfield{author}{\bibinfo{person}{Rachel K. E.~Bellamy et al.}} \bibinfo{year}{2018}\natexlab{}.
\newblock \bibinfo{title}{AI Fairness 360: An Extensible Toolkit for Detecting, Understanding, and Mitigating Unwanted Algorithmic Bias}.
\newblock
\newblock
\showeprint[arxiv]{1810.01943}~[cs.AI]


\bibitem[et~al.(2024)]%
        {cholia2024essdive}
\bibfield{author}{\bibinfo{person}{S.~Cholia et al.}} \bibinfo{year}{2024}\natexlab{}.
\newblock \bibinfo{title}{{ESS-DIVE} Overview: A Scalable, User-Focused Repository for Earth and Environmental Science Data}.
\newblock \bibinfo{howpublished}{Scientific Data Division, Lawrence Berkeley National Laboratory, Berkeley, CA}.
\newblock
\urldef\tempurl%
\url{https://ess-dive.lbl.gov/}
\showURL{%
\tempurl}


\bibitem[{European Parliament} and {Council of the European Union}(2016)]%
        {GDPR2016a}
\bibfield{author}{\bibinfo{person}{{European Parliament}} {and} \bibinfo{person}{{Council of the European Union}}.} \bibinfo{year}{2016}\natexlab{}.
\newblock \bibinfo{booktitle}{\emph{Regulation ({EU}) 2016/679 of the {European} {Parliament} and of the {Council}}}.
\newblock
\urldef\tempurl%
\url{https://data.europa.eu/eli/reg/2016/679/oj}
\showURL{%
\tempurl}


\bibitem[{FAIRassist.org}(nd)]%
        {fairassist}
\bibfield{author}{\bibinfo{person}{{FAIRassist.org}}.} \bibinfo{year}{n.d.}\natexlab{}.
\newblock \bibinfo{title}{{FAIRassist.Org}}.
\newblock \bibinfo{howpublished}{\url{https://fairassist.org}}.
\newblock
\newblock
\shownote{Jan. 6, 2024}.


\bibitem[Feldman~et al.(2015)]%
        {10.1145/2783258.2783311}
\bibfield{author}{\bibinfo{person}{Michael Feldman~et al.}} \bibinfo{year}{2015}\natexlab{}.
\newblock \showarticletitle{Certifying and Removing Disparate Impact}. In \bibinfo{booktitle}{\emph{Proceedings of the 21th ACM SIGKDD International Conference on Knowledge Discovery and Data Mining}} (Sydney, NSW, Australia) \emph{(\bibinfo{series}{KDD '15})}. \bibinfo{publisher}{Association for Computing Machinery}, \bibinfo{address}{New York, NY, USA}, \bibinfo{pages}{259–268}.
\newblock
\showISBNx{9781450336642}


\bibitem[Flesch(1986)]%
        {flesch1986art}
\bibfield{author}{\bibinfo{person}{Rudolf Flesch}.} \bibinfo{year}{1986}\natexlab{}.
\newblock \bibinfo{booktitle}{\emph{The Art of Readable Writing} (\bibinfo{edition}{19th print.-collier books ed} ed.)}.
\newblock \bibinfo{publisher}{MacMillan}.
\newblock


\bibitem[Forman(2003)]%
        {10.5555/944919.944974}
\bibfield{author}{\bibinfo{person}{George Forman}.} \bibinfo{year}{2003}\natexlab{}.
\newblock \showarticletitle{An Extensive Empirical Study of Feature Selection Metrics for Text Classification}.
\newblock \bibinfo{journal}{\emph{J. Mach. Learn. Res.}}  \bibinfo{volume}{3} (\bibinfo{date}{mar} \bibinfo{year}{2003}), \bibinfo{numpages}{17}~pages.
\newblock
\showISSN{1532-4435}


\bibitem[Ghorbani and Zou(2019)]%
        {ghorbani2019data}
\bibfield{author}{\bibinfo{person}{Amirata Ghorbani} {and} \bibinfo{person}{James Zou}.} \bibinfo{year}{2019}\natexlab{}.
\newblock \bibinfo{title}{Data Shapley: Equitable Valuation of Data for Machine Learning}.
\newblock
\newblock
\showeprint[arxiv]{1904.02868}~[stat.ML]


\bibitem[Gini(1912)]%
        {gini1912variability}
\bibfield{author}{\bibinfo{person}{C. Gini}.} \bibinfo{year}{1912}\natexlab{}.
\newblock \showarticletitle{{Variability and Mutability: Contribution to the Study of Statistical Distribution and Relations}}.
\newblock \bibinfo{journal}{\emph{Studi Economico-Giuridici della R}} (\bibinfo{year}{1912}).
\newblock


\bibitem[Hall and Smith(1999)]%
        {hall1999feature}
\bibfield{author}{\bibinfo{person}{Mark~A. Hall} {and} \bibinfo{person}{Lloyd~A. Smith}.} \bibinfo{year}{1999}\natexlab{}.
\newblock \showarticletitle{{Feature Selection for Machine Learning: Comparing a Correlation-Based Filter Approach to the Wrapper}}. In \bibinfo{booktitle}{\emph{FLAIRS}}. \bibinfo{pages}{235--239}.
\newblock


\bibitem[Haykin(2009)]%
        {haykin2009neural}
\bibfield{author}{\bibinfo{person}{Simon~S. Haykin}.} \bibinfo{year}{2009}\natexlab{}.
\newblock \bibinfo{booktitle}{\emph{Neural networks and learning machines} (\bibinfo{edition}{third} ed.)}.
\newblock \bibinfo{publisher}{Pearson Education}, \bibinfo{address}{Upper Saddle River, NJ}.
\newblock


\bibitem[He et~al\mbox{.}(2005)]%
        {he2005laplacian}
\bibfield{author}{\bibinfo{person}{Xiaofei He}, \bibinfo{person}{Deng Cai}, {and} \bibinfo{person}{Partha Niyogi}.} \bibinfo{year}{2005}\natexlab{}.
\newblock \showarticletitle{{Laplacian Score for Feature Selection}}. In \bibinfo{booktitle}{\emph{NIPS}}. \bibinfo{pages}{507--514}.
\newblock


\bibitem[Heinrich and Klier(2015)]%
        {HEINRICH201582}
\bibfield{author}{\bibinfo{person}{Bernd Heinrich} {and} \bibinfo{person}{Mathias Klier}.} \bibinfo{year}{2015}\natexlab{}.
\newblock \showarticletitle{Metric-based data quality assessment — Developing and evaluating a probability-based currency metric}.
\newblock \bibinfo{journal}{\emph{Decision Support Systems}}  \bibinfo{volume}{72} (\bibinfo{year}{2015}), \bibinfo{pages}{82--96}.
\newblock
\showISSN{0167-9236}
\urldef\tempurl%
\url{https://doi.org/10.1016/j.dss.2015.02.009}
\showDOI{\tempurl}


\bibitem[Heusel et~al\mbox{.}(2018)]%
        {heusel2018ganstrainedtimescaleupdate}
\bibfield{author}{\bibinfo{person}{Martin Heusel}, \bibinfo{person}{Hubert Ramsauer}, \bibinfo{person}{Thomas Unterthiner}, \bibinfo{person}{Bernhard Nessler}, {and} \bibinfo{person}{Sepp Hochreiter}.} \bibinfo{year}{2018}\natexlab{}.
\newblock \bibinfo{title}{GANs Trained by a Two Time-Scale Update Rule Converge to a Local Nash Equilibrium}.
\newblock
\newblock
\showeprint[arxiv]{1706.08500}~[cs.LG]
\urldef\tempurl%
\url{https://arxiv.org/abs/1706.08500}
\showURL{%
\tempurl}


\bibitem[Hiniduma~et al.(2024)]%
        {10.1145/3676288.3676296}
\bibfield{author}{\bibinfo{person}{Kaveen Hiniduma~et al.}} \bibinfo{year}{2024}\natexlab{}.
\newblock \showarticletitle{AI Data Readiness Inspector (AIDRIN) for Quantitative Assessment of Data Readiness for AI}. In \bibinfo{booktitle}{\emph{Proceedings of the 36th International Conference on Scientific and Statistical Database Management}} (Rennes, France) \emph{(\bibinfo{series}{SSDBM '24})}. Article \bibinfo{articleno}{7}, \bibinfo{numpages}{12}~pages.
\newblock
\showISBNx{9798400710209}


\bibitem[Holland et~al\mbox{.}(2018)]%
        {holland2018dataset}
\bibfield{author}{\bibinfo{person}{Sarah Holland}, \bibinfo{person}{Ahmed Hosny}, \bibinfo{person}{Sarah Newman}, \bibinfo{person}{Joshua Joseph}, {and} \bibinfo{person}{Kasia Chmielinski}.} \bibinfo{year}{2018}\natexlab{}.
\newblock \showarticletitle{The Dataset Nutrition Label: A Framework To Drive Higher Data Quality Standards}.
\newblock  (\bibinfo{year}{2018}).
\newblock
\showeprint[arxiv]{arXiv:1805.03677}~[cs.DB]


\bibitem[Huynh-Thu and Ghanbari(2008)]%
        {Huynh-Thu}
\bibfield{author}{\bibinfo{person}{Q. Huynh-Thu} {and} \bibinfo{person}{M. Ghanbari}.} \bibinfo{year}{2008}\natexlab{}.
\newblock \showarticletitle{Scope of validity of PSNR in image/video quality assessment}.
\newblock \bibinfo{journal}{\emph{Electronics Letters}} \bibinfo{volume}{44}, \bibinfo{number}{13} (\bibinfo{date}{Jun 19} \bibinfo{year}{2008}), \bibinfo{pages}{1--2}.
\newblock
\showISBNx{00135194}


\bibitem[Hwang(2022)]%
        {website:datacenterknowledge}
\bibfield{author}{\bibinfo{person}{Helen Hwang}.} \bibinfo{year}{2022}\natexlab{}.
\newblock \bibinfo{title}{{New AI readiness report reveals insights into ML lifecycle}}.
\newblock \bibinfo{howpublished}{\url{https://www.datacenterknowledge.com/machine-learning/new-ai-readiness-report-reveals-insights-ml-lifecycle}}.
\newblock
\newblock
\shownote{Accessed on May 15, 2023}.


\bibitem[Infolabs(2024)]%
        {infolabs2024outlier}
\bibfield{author}{\bibinfo{person}{Espire Infolabs}.} \bibinfo{year}{2024}\natexlab{}.
\newblock \bibinfo{title}{Outlier Detection Redefined: A Deep Dive into AI’s Impact: Espire Blog}.
\newblock
\newblock
\urldef\tempurl%
\url{https://www.espire.com/blog/posts/outlier-detection-redefined-a-deep-dive-into-ai-impact}
\showURL{%
\tempurl}
\newblock
\shownote{Accessed: 2024-08-06}.


\bibitem[Informatica(nd)]%
        {informatica}
\bibfield{author}{\bibinfo{person}{Informatica}.} \bibinfo{year}{n.d.}\natexlab{}.
\newblock \bibinfo{booktitle}{\emph{Data Quality Metrics \& Measures - All You Need to Know}}.
\newblock
\newblock
\shownote{Accessed Aug. 26, 2024}.


\bibitem[{International Telecommunication Union}(2018)]%
        {itu_p808}
\bibfield{author}{\bibinfo{person}{{International Telecommunication Union}}.} \bibinfo{year}{2018}\natexlab{}.
\newblock \bibinfo{booktitle}{\emph{{ITU-T Recommendation P.808: Subjective Evaluation of Speech Quality with a Crowdsourcing Approach}}}.
\newblock \bibinfo{type}{{T}echnical {R}eport}. \bibinfo{institution}{{International Telecommunication Union}}, \bibinfo{address}{Geneva}.
\newblock


\bibitem[Itakura and Saito(1968)]%
        {itakura1968analysis}
\bibfield{author}{\bibinfo{person}{F. Itakura} {and} \bibinfo{person}{S. Saito}.} \bibinfo{year}{1968}\natexlab{}.
\newblock \showarticletitle{Analysis Synthesis Telephony Based on the Maximum Likelihood Method}. In \bibinfo{booktitle}{\emph{Proc. 6th Int. Congr. Acoust.}} \bibinfo{address}{Tokyo, Japan}.
\newblock


\bibitem[Jaro(1976)]%
        {jaro1976unimatch}
\bibfield{author}{\bibinfo{person}{M.A. Jaro}.} \bibinfo{year}{1976}\natexlab{}.
\newblock \bibinfo{booktitle}{\emph{Unimatch: A Record Linkage System: User's Manual}}.
\newblock \bibinfo{type}{{T}echnical {R}eport}. \bibinfo{institution}{US Bureau of the Census}, \bibinfo{address}{Washington, D.C.}
\newblock


\bibitem[Jayant and Noll(1984)]%
        {jayant1984digital}
\bibfield{author}{\bibinfo{person}{N.C. Jayant} {and} \bibinfo{person}{P. Noll}.} \bibinfo{year}{1984}\natexlab{}.
\newblock \bibinfo{booktitle}{\emph{Digital Coding of Waveforms: Principles and Applications to Speech and Video}}.
\newblock \bibinfo{publisher}{Prentice Hall}, \bibinfo{address}{NJ, USA}.
\newblock


\bibitem[Jones and Slaughter(2019)]%
        {Jones_Slaughter_2019}
\bibfield{author}{\bibinfo{person}{Matthew~B Jones} {and} \bibinfo{person}{Peter Slaughter}.} \bibinfo{year}{2019}\natexlab{}.
\newblock
\newblock
\urldef\tempurl%
\url{https://www.dataone.org/uploads/dataonewebinar_jonesslaughter_fairmetadata_190514.pdf}
\showURL{%
\tempurl}


\bibitem[Joseph(2022)]%
        {roshan2022optimal}
\bibfield{author}{\bibinfo{person}{V.~Roshan Joseph}.} \bibinfo{year}{2022}\natexlab{}.
\newblock \showarticletitle{Optimal Ratio for Data Splitting}.
\newblock \bibinfo{journal}{\emph{Statistical Analysis and Data Mining: The ASA Data Science Journal}} \bibinfo{volume}{15}, \bibinfo{number}{4} (\bibinfo{date}{August} \bibinfo{year}{2022}), \bibinfo{pages}{531--538}.
\newblock
\urldef\tempurl%
\url{https://doi.org/10.1002/sam.11583}
\showDOI{\tempurl}


\bibitem[Jäger et~al\mbox{.}(2021)]%
        {jaeger2021benchmark}
\bibfield{author}{\bibinfo{person}{Sven Jäger}, \bibinfo{person}{Anders Allhorn}, {and} \bibinfo{person}{Felix Bießmann}.} \bibinfo{year}{2021}\natexlab{}.
\newblock \showarticletitle{A Benchmark for Data Imputation Methods}.
\newblock \bibinfo{journal}{\emph{Frontiers in Big Data}}  \bibinfo{volume}{4} (\bibinfo{year}{2021}), \bibinfo{pages}{693674}.
\newblock
\urldef\tempurl%
\url{https://doi.org/10.3389/fdata.2021.693674}
\showDOI{\tempurl}


\bibitem[Kaiser et~al\mbox{.}(1970)]%
        {Kaiser_Klier_Heinrich_1970}
\bibfield{author}{\bibinfo{person}{M. Kaiser}, \bibinfo{person}{Mathias Klier}, {and} \bibinfo{person}{Bernd Heinrich}.} \bibinfo{year}{1970}\natexlab{}.
\newblock \bibinfo{title}{[PDF] how to measure data quality? - A metric-based approach}.
\newblock
\newblock
\urldef\tempurl%
\url{https://www.semanticscholar.org/paper/How-to-Measure-Data-Quality-A-Metric-Based-Approach-Kaiser-Klier/afcdf53c5a88f3320c861ad3f09f28237b6744cb}
\showURL{%
\tempurl}


\bibitem[Kastryulin et~al\mbox{.}(2019)]%
        {piq}
\bibfield{author}{\bibinfo{person}{Sergey Kastryulin}, \bibinfo{person}{Dzhamil Zakirov}, {and} \bibinfo{person}{Denis Prokopenko}.} \bibinfo{year}{2019}\natexlab{}.
\newblock \bibinfo{title}{{PyTorch Image Quality}: Metrics and Measure for Image Quality Assessment}.
\newblock
\newblock
\urldef\tempurl%
\url{https://github.com/photosynthesis-team/piq}
\showURL{%
\tempurl}
\newblock
\shownote{Open-source software available at https://github.com/photosynthesis-team/piq}.


\bibitem[Kastryulin et~al\mbox{.}(2022)]%
        {kastryulin2022piq}
\bibfield{author}{\bibinfo{person}{Sergey Kastryulin}, \bibinfo{person}{Jamil Zakirov}, \bibinfo{person}{Denis Prokopenko}, {and} \bibinfo{person}{Dmitry~V. Dylov}.} \bibinfo{year}{2022}\natexlab{}.
\newblock \bibinfo{title}{PyTorch Image Quality: Metrics for Image Quality Assessment}.
\newblock
\newblock
\urldef\tempurl%
\url{https://doi.org/10.48550/ARXIV.2208.14818}
\showDOI{\tempurl}


\bibitem[Kemka(2019)]%
        {Kemka_2019}
\bibfield{author}{\bibinfo{person}{Martin Kemka}.} \bibinfo{year}{2019}\natexlab{}.
\newblock \bibinfo{title}{Learning Amazon Sagemaker}.
\newblock
\newblock
\urldef\tempurl%
\url{https://docs.aws.amazon.com/sagemaker/latest/dg/clarify-data-bias-metric-cddl.html}
\showURL{%
\tempurl}


\bibitem[Koh and Liang(2017)]%
        {koh2017influence}
\bibfield{author}{\bibinfo{person}{Pang~Wei Koh} {and} \bibinfo{person}{Percy Liang}.} \bibinfo{year}{2017}\natexlab{}.
\newblock \showarticletitle{Understanding Black-box Predictions via Influence Functions}. In \bibinfo{booktitle}{\emph{International Conference on Machine Learning}}.
\newblock


\bibitem[Lakhani(2020)]%
        {lakhani2020importance}
\bibfield{author}{\bibinfo{person}{Paras Lakhani}.} \bibinfo{year}{2020}\natexlab{}.
\newblock \showarticletitle{The Importance of Image Resolution in Building Deep Learning Models for Medical Imaging}.
\newblock \bibinfo{journal}{\emph{Radiology: Artificial Intelligence}} \bibinfo{volume}{2}, \bibinfo{number}{1} (\bibinfo{year}{2020}), \bibinfo{pages}{e190177}.
\newblock
\urldef\tempurl%
\url{https://doi.org/10.1148/ryai.2019190177}
\showURL{%
\tempurl}


\bibitem[Lavitas et~al\mbox{.}(2021)]%
        {lavitas2021annotation}
\bibfield{author}{\bibinfo{person}{Liliya Lavitas}, \bibinfo{person}{Olivia Redfield}, \bibinfo{person}{Allen Lee}, \bibinfo{person}{Daniel Fletcher}, \bibinfo{person}{Matthias Eck}, {and} \bibinfo{person}{Sunil Janardhanan}.} \bibinfo{year}{2021}\natexlab{}.
\newblock \showarticletitle{Annotation quality framework-accuracy, credibility, and consistency}. In \bibinfo{booktitle}{\emph{NEURIPS 2021 Workshop for Data Centric AI}}.
\newblock


\bibitem[Levenshtein(1965)]%
        {Levenshtein1965}
\bibfield{author}{\bibinfo{person}{V.I. Levenshtein}.} \bibinfo{year}{1965}\natexlab{}.
\newblock \showarticletitle{Binary Codes Capable of Correcting Deletions, Insertions and Reversals}.
\newblock \bibinfo{journal}{\emph{Doklady Akademii Nauk SSSR}} \bibinfo{volume}{163}, \bibinfo{number}{4} (\bibinfo{year}{1965}), \bibinfo{pages}{845--848}.
\newblock
\newblock
\shownote{Original in Russian---translation in Soviet Physics Doklady, vol. 10, no. 8, pp. 707--710, 1966}.


\bibitem[Lewis(1992)]%
        {lewis1992feature}
\bibfield{author}{\bibinfo{person}{David~D. Lewis}.} \bibinfo{year}{1992}\natexlab{}.
\newblock \showarticletitle{{Feature Selection and Feature Extraction for Text Categorization}}. In \bibinfo{booktitle}{\emph{Workshop on Speech and Natural Language}}. \bibinfo{pages}{212--217}.
\newblock


\bibitem[Leys et~al\mbox{.}(2013)]%
        {leys2013detecting}
\bibfield{author}{\bibinfo{person}{Christophe Leys}, \bibinfo{person}{Christophe Ley}, \bibinfo{person}{Olivier Klein}, \bibinfo{person}{Pierre Bernard}, {and} \bibinfo{person}{Laurent Licata}.} \bibinfo{year}{2013}\natexlab{}.
\newblock \showarticletitle{{Detecting outliers: Do not use standard deviation around the mean, use absolute deviation around the median}}.
\newblock \bibinfo{journal}{\emph{J. Exp. Social Psychol.}} \bibinfo{volume}{49}, \bibinfo{number}{4} (\bibinfo{year}{2013}), \bibinfo{pages}{764--766}.
\newblock


\bibitem[Li et~al\mbox{.}(2017)]%
        {10.1145/3136625}
\bibfield{author}{\bibinfo{person}{Jundong Li}, \bibinfo{person}{Kewei Cheng}, \bibinfo{person}{Suhang Wang}, \bibinfo{person}{Fred Morstatter}, \bibinfo{person}{Robert~P. Trevino}, \bibinfo{person}{Jiliang Tang}, {and} \bibinfo{person}{Huan Liu}.} \bibinfo{year}{2017}\natexlab{}.
\newblock \showarticletitle{Feature Selection: A Data Perspective}.
\newblock \bibinfo{journal}{\emph{ACM Comput. Surv.}} \bibinfo{volume}{50}, \bibinfo{number}{6}, Article \bibinfo{articleno}{94} (\bibinfo{date}{dec} \bibinfo{year}{2017}), \bibinfo{numpages}{45}~pages.
\newblock
\showISSN{0360-0300}
\urldef\tempurl%
\url{https://doi.org/10.1145/3136625}
\showDOI{\tempurl}


\bibitem[Li et~al\mbox{.}(2021)]%
        {9458702}
\bibfield{author}{\bibinfo{person}{Peng Li}, \bibinfo{person}{Xi Rao}, \bibinfo{person}{Jennifer Blase}, \bibinfo{person}{Yue Zhang}, \bibinfo{person}{Xu Chu}, {and} \bibinfo{person}{Ce Zhang}.} \bibinfo{year}{2021}\natexlab{}.
\newblock \showarticletitle{CleanML: A Study for Evaluating the Impact of Data Cleaning on ML Classification Tasks}. In \bibinfo{booktitle}{\emph{2021 IEEE 37th International Conference on Data Engineering (ICDE)}}. \bibinfo{pages}{13--24}.
\newblock
\urldef\tempurl%
\url{https://doi.org/10.1109/ICDE51399.2021.00009}
\showDOI{\tempurl}


\bibitem[Lin et~al\mbox{.}(2005)]%
        {lin2005visual}
\bibfield{author}{\bibinfo{person}{Weisi Lin}, \bibinfo{person}{Li Dong}, {and} \bibinfo{person}{Ping Xue}.} \bibinfo{year}{2005}\natexlab{}.
\newblock \showarticletitle{Visual distortion gauge based on discrimination of noticeable contrast changes}.
\newblock \bibinfo{journal}{\emph{IEEE transactions on circuits and systems for video technology}} \bibinfo{volume}{15}, \bibinfo{number}{7} (\bibinfo{year}{2005}), \bibinfo{pages}{900--909}.
\newblock


\bibitem[Lin and {Jay Kuo}(2011)]%
        {LIN2011297}
\bibfield{author}{\bibinfo{person}{Weisi Lin} {and} \bibinfo{person}{C.-C. {Jay Kuo}}.} \bibinfo{year}{2011}\natexlab{}.
\newblock \showarticletitle{Perceptual visual quality metrics: A survey}.
\newblock \bibinfo{journal}{\emph{Journal of Visual Communication and Image Representation}} \bibinfo{volume}{22}, \bibinfo{number}{4} (\bibinfo{year}{2011}), \bibinfo{pages}{297--312}.
\newblock
\showISSN{1047-3203}
\urldef\tempurl%
\url{https://doi.org/10.1016/j.jvcir.2011.01.005}
\showDOI{\tempurl}


\bibitem[Liu and Setiono(1995)]%
        {liu1995chi2}
\bibfield{author}{\bibinfo{person}{Huan Liu} {and} \bibinfo{person}{Rudy Setiono}.} \bibinfo{year}{1995}\natexlab{}.
\newblock \showarticletitle{{Chi2: Feature Selection and Discretization of Numeric Attributes}}. In \bibinfo{booktitle}{\emph{ICTAI}}. \bibinfo{pages}{388--391}.
\newblock


\bibitem[Liu~et al.(2020)]%
        {liu2020admm}
\bibfield{author}{\bibinfo{person}{Sijia Liu~et al.}} \bibinfo{year}{2020}\natexlab{}.
\newblock \showarticletitle{An ADMM-based Framework for AutoML Pipeline Configuration}. In \bibinfo{booktitle}{\emph{Proceedings of the AAAI Conference on Artificial Intelligence}}, Vol.~\bibinfo{volume}{34}. \bibinfo{pages}{4892--4899}.
\newblock


\bibitem[Longpr{\'e} et~al\mbox{.}(2017)]%
        {Longpr2017EntropyAA}
\bibfield{author}{\bibinfo{person}{Luc Longpr{\'e}}, \bibinfo{person}{Vladik Kreinovich}, {and} \bibinfo{person}{Thongchai Dumrongpokaphan}.} \bibinfo{year}{2017}\natexlab{}.
\newblock \showarticletitle{Entropy as a Measure of Average Loss of Privacy}.
\newblock \bibinfo{journal}{\emph{Thai Journal of Mathematics}} (\bibinfo{year}{2017}), \bibinfo{pages}{7--15}.
\newblock
\urldef\tempurl%
\url{https://api.semanticscholar.org/CorpusID:6672504}
\showURL{%
\tempurl}


\bibitem[Lu et~al\mbox{.}(2019)]%
        {lu2019bayes}
\bibfield{author}{\bibinfo{person}{Yang Lu}, \bibinfo{person}{Yiu ming Cheung}, {and} \bibinfo{person}{Yuan~Yan Tang}.} \bibinfo{year}{2019}\natexlab{}.
\newblock \bibinfo{title}{Bayes Imbalance Impact Index: A Measure of Class Imbalanced Dataset for Classification Problem}.
\newblock
\newblock
\showeprint[arxiv]{1901.10173}~[cs.LG]


\bibitem[Luhn(1957)]%
        {5392697}
\bibfield{author}{\bibinfo{person}{H.~P. Luhn}.} \bibinfo{year}{1957}\natexlab{}.
\newblock \showarticletitle{A Statistical Approach to Mechanized Encoding and Searching of Literary Information}.
\newblock \bibinfo{journal}{\emph{IBM Journal of Research and Development}} \bibinfo{volume}{1}, \bibinfo{number}{4} (\bibinfo{year}{1957}), \bibinfo{pages}{309--317}.
\newblock
\urldef\tempurl%
\url{https://doi.org/10.1147/rd.14.0309}
\showDOI{\tempurl}


\bibitem[Markham(2024)]%
        {markham2024ai}
\bibfield{author}{\bibinfo{person}{Chris Markham}.} \bibinfo{year}{2024}\natexlab{}.
\newblock \bibinfo{title}{How AI Can Uncover Data Outliers and Patterns in Patient Behavior}.
\newblock \bibinfo{howpublished}{\url{https://www.saama.com/how-ai-can-uncover-data-outliers-and-patterns-in-patient-behavior/}}.
\newblock
\urldef\tempurl%
\url{https://www.saama.com/how-ai-can-uncover-data-outliers-and-patterns-in-patient-behavior/}
\showURL{%
\tempurl}


\bibitem[Marziliano et~al\mbox{.}(2002)]%
        {marziliano2002no}
\bibfield{author}{\bibinfo{person}{Pina Marziliano}, \bibinfo{person}{Frederic Dufaux}, \bibinfo{person}{Stefan Winkler}, {and} \bibinfo{person}{Touradj Ebrahimi}.} \bibinfo{year}{2002}\natexlab{}.
\newblock \showarticletitle{A no-reference perceptual blur metric}. In \bibinfo{booktitle}{\emph{Proceedings. International conference on image processing}}, Vol.~\bibinfo{volume}{3}. IEEE, \bibinfo{pages}{III--III}.
\newblock


\bibitem[McCarthy(2005)]%
        {mccarthy2005assessment}
\bibfield{author}{\bibinfo{person}{Philip~M McCarthy}.} \bibinfo{year}{2005}\natexlab{}.
\newblock \emph{\bibinfo{title}{An assessment of the range and usefulness of lexical diversity measures and the potential of the measure of textual, lexical diversity (MTLD)}}.
\newblock \bibinfo{thesistype}{Ph.\,D. Dissertation}. \bibinfo{school}{The University of Memphis}.
\newblock


\bibitem[McCarthy and Jarvis(2010)]%
        {mccarthy2010mtld}
\bibfield{author}{\bibinfo{person}{Peter~M. McCarthy} {and} \bibinfo{person}{Scott Jarvis}.} \bibinfo{year}{2010}\natexlab{}.
\newblock \showarticletitle{{MTLD, VOCD-D, and HD-D: A Validation Study of Sophisticated Approaches to Lexical Diversity Assessment}}.
\newblock \bibinfo{journal}{\emph{Behavior Research Methods}} \bibinfo{volume}{42}, \bibinfo{number}{2} (\bibinfo{year}{2010}), \bibinfo{pages}{381--392}.
\newblock
\urldef\tempurl%
\url{https://doi.org/10.3758/BRM.42.2.381}
\showDOI{\tempurl}


\bibitem[Mimno~et al.(2011)]%
        {10.5555/2145432.2145462}
\bibfield{author}{\bibinfo{person}{David Mimno~et al.}} \bibinfo{year}{2011}\natexlab{}.
\newblock \showarticletitle{Optimizing Semantic Coherence in Topic Models}. In \bibinfo{booktitle}{\emph{Proceedings of the Conference on Empirical Methods in Natural Language Processing}} (Edinburgh, United Kingdom) \emph{(\bibinfo{series}{EMNLP '11})}. \bibinfo{publisher}{Association for Computational Linguistics}, \bibinfo{address}{USA}, \bibinfo{pages}{262–272}.
\newblock
\showISBNx{9781937284114}


\bibitem[Mittal et~al\mbox{.}(2012)]%
        {6272356}
\bibfield{author}{\bibinfo{person}{Anish Mittal}, \bibinfo{person}{Anush~Krishna Moorthy}, {and} \bibinfo{person}{Alan~Conrad Bovik}.} \bibinfo{year}{2012}\natexlab{}.
\newblock \showarticletitle{No-Reference Image Quality Assessment in the Spatial Domain}.
\newblock \bibinfo{journal}{\emph{IEEE Transactions on Image Processing}} \bibinfo{volume}{21}, \bibinfo{number}{12} (\bibinfo{year}{2012}), \bibinfo{pages}{4695--4708}.
\newblock
\urldef\tempurl%
\url{https://doi.org/10.1109/TIP.2012.2214050}
\showDOI{\tempurl}


\bibitem[Monge and Elkan(1996)]%
        {monge1996fieldmatching}
\bibfield{author}{\bibinfo{person}{A.E. Monge} {and} \bibinfo{person}{C.P. Elkan}.} \bibinfo{year}{1996}\natexlab{}.
\newblock \showarticletitle{The Field Matching Problem: Algorithms and Applications}. In \bibinfo{booktitle}{\emph{Proc. Second Int'l Conf. Knowledge Discovery and Data Mining (KDD '96)}}. \bibinfo{pages}{267--270}.
\newblock


\bibitem[Newman~et al.(2010)]%
        {10.1145/1816123.1816156}
\bibfield{author}{\bibinfo{person}{David Newman~et al.}} \bibinfo{year}{2010}\natexlab{}.
\newblock \showarticletitle{Evaluating Topic Models for Digital Libraries}. In \bibinfo{booktitle}{\emph{Proceedings of the 10th Annual Joint Conference on Digital Libraries}} (Gold Coast, Queensland, Australia) \emph{(\bibinfo{series}{JCDL '10})}. \bibinfo{pages}{215–224}.
\newblock
\showISBNx{9781450300858}
\urldef\tempurl%
\url{https://doi.org/10.1145/1816123.1816156}
\showDOI{\tempurl}


\bibitem[Nie et~al\mbox{.}(2008)]%
        {nie2008trace}
\bibfield{author}{\bibinfo{person}{Feiping Nie}, \bibinfo{person}{Shiming Xiang}, \bibinfo{person}{Yangqing Jia}, \bibinfo{person}{Changshui Zhang}, {and} \bibinfo{person}{Shuicheng Yan}.} \bibinfo{year}{2008}\natexlab{}.
\newblock \showarticletitle{{Trace Ratio Criterion for Feature Selection}}. In \bibinfo{booktitle}{\emph{AAAI}}.
\newblock


\bibitem[Ntoutsi~et al.(2020)]%
        {ntoutsi2020bias}
\bibfield{author}{\bibinfo{person}{E. Ntoutsi~et al.}} \bibinfo{year}{2020}\natexlab{}.
\newblock \showarticletitle{{Bias in data-driven artificial intelligence systems—An introductory survey}}.
\newblock \bibinfo{journal}{\emph{Wiley Interdisciplinary Reviews: Data Mining and Knowledge Discovery}} \bibinfo{volume}{10}, \bibinfo{number}{3} (\bibinfo{year}{2020}), \bibinfo{pages}{e1356}.
\newblock
\urldef\tempurl%
\url{https://doi.org/10.1002/widm.1356}
\showDOI{\tempurl}


\bibitem[Oh(2011)]%
        {OH2011115}
\bibfield{author}{\bibinfo{person}{Sejong Oh}.} \bibinfo{year}{2011}\natexlab{}.
\newblock \showarticletitle{A new dataset evaluation method based on category overlap}.
\newblock \bibinfo{journal}{\emph{Computers in Biology and Medicine}} \bibinfo{volume}{41}, \bibinfo{number}{2} (\bibinfo{year}{2011}), \bibinfo{pages}{115--122}.
\newblock
\showISSN{0010-4825}
\urldef\tempurl%
\url{https://doi.org/10.1016/j.compbiomed.2010.12.006}
\showDOI{\tempurl}


\bibitem[Ortigosa-Hernández et~al\mbox{.}(2017)]%
        {ORTIGOSAHERNANDEZ201732}
\bibfield{author}{\bibinfo{person}{Jonathan Ortigosa-Hernández}, \bibinfo{person}{Iñaki Inza}, {and} \bibinfo{person}{Jose~A. Lozano}.} \bibinfo{year}{2017}\natexlab{}.
\newblock \showarticletitle{Measuring the class-imbalance extent of multi-class problems}.
\newblock \bibinfo{journal}{\emph{Pattern Recognition Letters}}  \bibinfo{volume}{98} (\bibinfo{year}{2017}), \bibinfo{pages}{32--38}.
\newblock
\showISSN{0167-8655}
\urldef\tempurl%
\url{https://doi.org/10.1016/j.patrec.2017.08.002}
\showDOI{\tempurl}


\bibitem[Papakyriakopoulos~et al.(2020)]%
        {10.1145/3351095.3372843}
\bibfield{author}{\bibinfo{person}{Orestis Papakyriakopoulos~et al.}} \bibinfo{year}{2020}\natexlab{}.
\newblock \showarticletitle{Bias in Word Embeddings}. In \bibinfo{booktitle}{\emph{Proceedings of the 2020 Conference on Fairness, Accountability, and Transparency}} (Barcelona, Spain) \emph{(\bibinfo{series}{FAT* '20})}. \bibinfo{pages}{446–457}.
\newblock
\showISBNx{9781450369367}
\urldef\tempurl%
\url{https://doi.org/10.1145/3351095.3372843}
\showDOI{\tempurl}


\bibitem[Pearson(2006)]%
        {10.1145/1147234.1147247}
\bibfield{author}{\bibinfo{person}{Ronald~K. Pearson}.} \bibinfo{year}{2006}\natexlab{}.
\newblock \showarticletitle{The problem of disguised missing data}.
\newblock \bibinfo{journal}{\emph{SIGKDD Explor. Newsl.}} \bibinfo{volume}{8}, \bibinfo{number}{1} (\bibinfo{date}{jun} \bibinfo{year}{2006}).
\newblock
\showISSN{1931-0145}
\urldef\tempurl%
\url{https://doi.org/10.1145/1147234.1147247}
\showDOI{\tempurl}


\bibitem[Pipino et~al\mbox{.}(2002)]%
        {10.1145/505248.506010}
\bibfield{author}{\bibinfo{person}{Leo~L. Pipino}, \bibinfo{person}{Yang~W. Lee}, {and} \bibinfo{person}{Richard~Y. Wang}.} \bibinfo{year}{2002}\natexlab{}.
\newblock \showarticletitle{Data Quality Assessment}.
\newblock \bibinfo{journal}{\emph{Commun. ACM}} \bibinfo{volume}{45}, \bibinfo{number}{4} (\bibinfo{date}{apr} \bibinfo{year}{2002}), \bibinfo{pages}{211–218}.
\newblock
\showISSN{0001-0782}
\urldef\tempurl%
\url{https://doi.org/10.1145/505248.506010}
\showDOI{\tempurl}


\bibitem[Pokrajac et~al\mbox{.}(2007)]%
        {pokrajac2007incremental}
\bibfield{author}{\bibinfo{person}{Dragoljub Pokrajac}, \bibinfo{person}{Aleksandar Lazarevic}, {and} \bibinfo{person}{Longin~Jan Latecki}.} \bibinfo{year}{2007}\natexlab{}.
\newblock \showarticletitle{{Incremental local outlier detection for data streams}}. In \bibinfo{booktitle}{\emph{Proc. IEEE Symp. Comput. Intell. Data Mining}}. \bibinfo{pages}{504--515}.
\newblock


\bibitem[Priestley et~al\mbox{.}(2023)]%
        {article:priestley}
\bibfield{author}{\bibinfo{person}{Maria Priestley}, \bibinfo{person}{Fionnt\'{a}n O’Donnell}, {and} \bibinfo{person}{Elena Simperl}.} \bibinfo{year}{2023}\natexlab{}.
\newblock \showarticletitle{A Survey of Data Quality Requirements That Matter in ML Development Pipelines}.
\newblock \bibinfo{journal}{\emph{J. Data and Information Quality}} (\bibinfo{date}{apr} \bibinfo{year}{2023}).
\newblock
\showISSN{1936-1955}
\urldef\tempurl%
\url{https://doi.org/10.1145/3592616}
\showDOI{\tempurl}
\newblock
\shownote{Just Accepted}.


\bibitem[Qaiser and Ali(2018)]%
        {Qaiser_S}
\bibfield{author}{\bibinfo{person}{Shahzad Qaiser} {and} \bibinfo{person}{Ramsha Ali}.} \bibinfo{year}{2018}\natexlab{}.
\newblock \showarticletitle{Text Mining: Use of TF-IDF to Examine the Relevance of Words to Documents}.
\newblock \bibinfo{journal}{\emph{International Journal of Computer Applications}}  \bibinfo{volume}{181} (\bibinfo{date}{07} \bibinfo{year}{2018}).
\newblock
\urldef\tempurl%
\url{https://doi.org/10.5120/ijca2018917395}
\showDOI{\tempurl}


\bibitem[Rajput et~al\mbox{.}(2023)]%
        {Rajput2023}
\bibfield{author}{\bibinfo{person}{Deepak Rajput}, \bibinfo{person}{Wanjun Wang}, {and} \bibinfo{person}{Cheng-Chung Chen}.} \bibinfo{year}{2023}\natexlab{}.
\newblock \showarticletitle{Evaluation of a decided sample size in machine learning applications}.
\newblock \bibinfo{journal}{\emph{BMC Bioinformatics}}  \bibinfo{volume}{24} (\bibinfo{year}{2023}), \bibinfo{pages}{48}.
\newblock
\urldef\tempurl%
\url{https://doi.org/10.1186/s12859-023-05156-9}
\showDOI{\tempurl}


\bibitem[Ramos et~al\mbox{.}(2003)]%
        {ramos2003using}
\bibfield{author}{\bibinfo{person}{Juan Ramos} {et~al\mbox{.}}} \bibinfo{year}{2003}\natexlab{}.
\newblock \showarticletitle{Using tf-idf to determine word relevance in document queries}. In \bibinfo{booktitle}{\emph{Proceedings of the first instructional conference on machine learning}}, Vol.~\bibinfo{volume}{242}. Citeseer, \bibinfo{pages}{29--48}.
\newblock


\bibitem[Reviriego et~al\mbox{.}(2023)]%
        {reviriego2023playingwordscomparingvocabulary}
\bibfield{author}{\bibinfo{person}{Pedro Reviriego}, \bibinfo{person}{Javier Conde}, \bibinfo{person}{Elena Merino-Gómez}, \bibinfo{person}{Gonzalo Martínez}, {and} \bibinfo{person}{José~Alberto Hernández}.} \bibinfo{year}{2023}\natexlab{}.
\newblock \bibinfo{title}{Playing with Words: Comparing the Vocabulary and Lexical Richness of ChatGPT and Humans}.
\newblock
\newblock
\showeprint[arxiv]{2308.07462}~[cs.CL]
\urldef\tempurl%
\url{https://arxiv.org/abs/2308.07462}
\showURL{%
\tempurl}


\bibitem[Ribeiro et~al\mbox{.}(2016)]%
        {ribeiro2016lime}
\bibfield{author}{\bibinfo{person}{Marco~Tulio Ribeiro}, \bibinfo{person}{Sameer Singh}, {and} \bibinfo{person}{Carlos Guestrin}.} \bibinfo{year}{2016}\natexlab{}.
\newblock \showarticletitle{“Why Should I Trust You?” Explaining the Predictions of Any Classifier}. In \bibinfo{booktitle}{\emph{Proceedings of the 22nd ACM SIGKDD International Conference on Knowledge Discovery and Data Mining}}.
\newblock


\bibitem[Rix~et al.(2001)]%
        {941023}
\bibfield{author}{\bibinfo{person}{A.W. Rix~et al.}} \bibinfo{year}{2001}\natexlab{}.
\newblock \showarticletitle{Perceptual evaluation of speech quality (PESQ)-a new method for speech quality assessment of telephone networks and codecs}. In \bibinfo{booktitle}{\emph{2001 IEEE International Conference on Acoustics, Speech, and Signal Processing. Proceedings (Cat. No.01CH37221)}}.
\newblock


\bibitem[Robnik-Šikonja and Kononenko(2003)]%
        {robnik2003theoretical}
\bibfield{author}{\bibinfo{person}{Marko Robnik-Šikonja} {and} \bibinfo{person}{Igor Kononenko}.} \bibinfo{year}{2003}\natexlab{}.
\newblock \showarticletitle{{Theoretical and Empirical Analysis of ReliefF and RReliefF}}.
\newblock \bibinfo{journal}{\emph{Machine Learning}} \bibinfo{volume}{53}, \bibinfo{number}{1-2} (\bibinfo{year}{2003}).
\newblock


\bibitem[Rocca-Serra et~al\mbox{.}(2023)]%
        {rocca-serra2023fair}
\bibfield{author}{\bibinfo{person}{P. Rocca-Serra}, \bibinfo{person}{W. Gu}, \bibinfo{person}{V. Ioannidis}, {et~al\mbox{.}}} \bibinfo{year}{2023}\natexlab{}.
\newblock \showarticletitle{The {FAIR} Cookbook - The Essential Resource for and by {FAIR} Doers}.
\newblock \bibinfo{journal}{\emph{Sci Data}}  \bibinfo{volume}{10} (\bibinfo{year}{2023}).
\newblock


\bibitem[R\"{o}der et~al\mbox{.}(2015)]%
        {10.1145/2684822.2685324}
\bibfield{author}{\bibinfo{person}{Michael R\"{o}der}, \bibinfo{person}{Andreas Both}, {and} \bibinfo{person}{Alexander Hinneburg}.} \bibinfo{year}{2015}\natexlab{}.
\newblock \showarticletitle{Exploring the Space of Topic Coherence Measures}. In \bibinfo{booktitle}{\emph{Proceedings of the Eighth ACM International Conference on Web Search and Data Mining}} (Shanghai, China) \emph{(\bibinfo{series}{WSDM '15})}. \bibinfo{pages}{399–408}.
\newblock
\showISBNx{9781450333177}
\urldef\tempurl%
\url{https://doi.org/10.1145/2684822.2685324}
\showDOI{\tempurl}


\bibitem[Rosner(1983)]%
        {rosner1983percentage}
\bibfield{author}{\bibinfo{person}{Bernard Rosner}.} \bibinfo{year}{1983}\natexlab{}.
\newblock \showarticletitle{{Percentage points for a generalized ESD many-outlier procedure}}.
\newblock \bibinfo{journal}{\emph{Technometrics}} \bibinfo{volume}{25}, \bibinfo{number}{2} (\bibinfo{year}{1983}), \bibinfo{pages}{165--172}.
\newblock


\bibitem[Rousseeuw and Hubert(2018)]%
        {rousseeuw2018anomaly}
\bibfield{author}{\bibinfo{person}{Peter~J. Rousseeuw} {and} \bibinfo{person}{Mia Hubert}.} \bibinfo{year}{2018}\natexlab{}.
\newblock \showarticletitle{{Anomaly detection by robust statistics}}.
\newblock \bibinfo{journal}{\emph{WIREs Data Mining Knowl. Discovery}} \bibinfo{volume}{8}, \bibinfo{number}{2} (\bibinfo{date}{Mar.} \bibinfo{year}{2018}), \bibinfo{pages}{e1236}.
\newblock


\bibitem[Russell(1922)]%
        {russell1922index}
\bibfield{author}{\bibinfo{person}{R.C. Russell}.} \bibinfo{year}{1922}\natexlab{}.
\newblock \bibinfo{title}{Index}.
\newblock
\newblock
\urldef\tempurl%
\url{http://patft.uspto.gov/netahtml/srchnum.htm}
\showURL{%
\tempurl}


\bibitem[Sabottke and Spieler(2020)]%
        {sabottke2020effect}
\bibfield{author}{\bibinfo{person}{Carl~F. Sabottke} {and} \bibinfo{person}{Bradley~M. Spieler}.} \bibinfo{year}{2020}\natexlab{}.
\newblock \showarticletitle{The Effect of Image Resolution on Deep Learning in Radiography}.
\newblock \bibinfo{journal}{\emph{Radiology: Artificial Intelligence}} \bibinfo{volume}{2}, \bibinfo{number}{1} (\bibinfo{year}{2020}), \bibinfo{pages}{e190015}.
\newblock
\urldef\tempurl%
\url{https://doi.org/10.1148/ryai.2019190015}
\showURL{%
\tempurl}


\bibitem[Santos et~al\mbox{.}(2020)]%
        {SANTOS2020111}
\bibfield{author}{\bibinfo{person}{Miriam~Seoane Santos}, \bibinfo{person}{Pedro~Henriques Abreu}, \bibinfo{person}{Szymon Wilk}, {and} \bibinfo{person}{João Santos}.} \bibinfo{year}{2020}\natexlab{}.
\newblock \showarticletitle{How distance metrics influence missing data imputation with k-nearest neighbours}.
\newblock \bibinfo{journal}{\emph{Pattern Recognition Letters}}  \bibinfo{volume}{136} (\bibinfo{year}{2020}), \bibinfo{pages}{111--119}.
\newblock
\showISSN{0167-8655}
\urldef\tempurl%
\url{https://doi.org/10.1016/j.patrec.2020.05.032}
\showDOI{\tempurl}


\bibitem[Schelter et~al\mbox{.}(2018)]%
        {schelter2018automating}
\bibfield{author}{\bibinfo{person}{Sebastian Schelter}, \bibinfo{person}{Dustin Lange}, \bibinfo{person}{Philipp Schmidt}, \bibinfo{person}{Meltem Celikel}, \bibinfo{person}{Felix Biessmann}, {and} \bibinfo{person}{Andreas Grafberger}.} \bibinfo{year}{2018}\natexlab{}.
\newblock \showarticletitle{Automating Large-Scale Data Quality Verification}.
\newblock \bibinfo{journal}{\emph{Proc. VLDB Endow.}} \bibinfo{volume}{11}, \bibinfo{number}{12} (\bibinfo{date}{August} \bibinfo{year}{2018}), \bibinfo{pages}{1781--1794}.
\newblock


\bibitem[Schmelzer(2019)]%
        {Schmelzer_2019}
\bibfield{author}{\bibinfo{person}{Ron Schmelzer}.} \bibinfo{year}{2019}\natexlab{}.
\newblock \bibinfo{title}{The Achilles’ Heel of AI}.
\newblock
\newblock
\urldef\tempurl%
\url{https://www.forbes.com/sites/cognitiveworld/2019/03/07/the-achilles-heel-of-ai/?sh=20e53e4d7be7}
\showURL{%
\tempurl}


\bibitem[Shahbazi et~al\mbox{.}(2023)]%
        {10.1145/3588433}
\bibfield{author}{\bibinfo{person}{Nima Shahbazi}, \bibinfo{person}{Yin Lin}, \bibinfo{person}{Abolfazl Asudeh}, {and} \bibinfo{person}{H.~V. Jagadish}.} \bibinfo{year}{2023}\natexlab{}.
\newblock \showarticletitle{Representation Bias in Data: A Survey on Identification and Resolution Techniques}.
\newblock \bibinfo{journal}{\emph{ACM Comput. Surv.}} (\bibinfo{date}{mar} \bibinfo{year}{2023}).
\newblock
\showISSN{0360-0300}
\urldef\tempurl%
\url{https://doi.org/10.1145/3588433}
\showDOI{\tempurl}
\newblock
\shownote{Just Accepted}.


\bibitem[Sheikh and Bovik(2006)]%
        {1576816}
\bibfield{author}{\bibinfo{person}{H.R. Sheikh} {and} \bibinfo{person}{A.C. Bovik}.} \bibinfo{year}{2006}\natexlab{}.
\newblock \showarticletitle{Image information and visual quality}.
\newblock \bibinfo{journal}{\emph{IEEE Transactions on Image Processing}} \bibinfo{volume}{15}, \bibinfo{number}{2} (\bibinfo{year}{2006}), \bibinfo{pages}{430--444}.
\newblock
\urldef\tempurl%
\url{https://doi.org/10.1109/TIP.2005.859378}
\showDOI{\tempurl}


\bibitem[Shrivastava et~al\mbox{.}(2020)]%
        {DQLearn-all}
\bibfield{author}{\bibinfo{person}{S. Shrivastava}, \bibinfo{person}{D. Patel}, \bibinfo{person}{N. Zhou}, \bibinfo{person}{A. Iyengar}, {and} \bibinfo{person}{A. Bhamidipaty}.} \bibinfo{year}{2020}\natexlab{}.
\newblock \showarticletitle{DQLearn: A Toolkit for Structured Data Quality Learning}. In \bibinfo{booktitle}{\emph{2020 {IEEE} International Conference on Big Data (Big Data)}}. \bibinfo{address}{Atlanta, GA, USA}, \bibinfo{pages}{1644--1653}.
\newblock
\urldef\tempurl%
\url{https://doi.org/10.1109/BigData50022.2020.9378296}
\showDOI{\tempurl}


\bibitem[Sidi~et al.(2012)]%
        {6204995}
\bibfield{author}{\bibinfo{person}{Fatimah Sidi~et al.}} \bibinfo{year}{2012}\natexlab{}.
\newblock \showarticletitle{Data quality: A survey of data quality dimensions}. In \bibinfo{booktitle}{\emph{2012 International Conference on Information Retrieval \& Knowledge Management}}. \bibinfo{pages}{300--304}.
\newblock
\urldef\tempurl%
\url{https://doi.org/10.1109/InfRKM.2012.6204995}
\showDOI{\tempurl}


\bibitem[Simha(2021)]%
        {Simha_2021}
\bibfield{author}{\bibinfo{person}{Simha}.} \bibinfo{year}{2021}\natexlab{}.
\newblock \bibinfo{title}{Understanding TF-IDF for Machine Learning}.
\newblock
\newblock
\urldef\tempurl%
\url{https://www.capitalone.com/tech/machine-learning/understanding-tf-idf/}
\showURL{%
\tempurl}


\bibitem[Simonetta et~al\mbox{.}(2021)]%
        {simonetta2021metrics}
\bibfield{author}{\bibinfo{person}{A. Simonetta}, \bibinfo{person}{A. Trenta}, \bibinfo{person}{M.~C. Paoletti}, {and} \bibinfo{person}{A. Vetrò}.} \bibinfo{year}{2021}\natexlab{}.
\newblock \showarticletitle{{Metrics for Identifying Bias in Datasets}}.
\newblock \bibinfo{journal}{\emph{SYSTEM}} (\bibinfo{year}{2021}).
\newblock


\bibitem[Simpson(1949)]%
        {simpson1949measurement}
\bibfield{author}{\bibinfo{person}{E. Simpson}.} \bibinfo{year}{1949}\natexlab{}.
\newblock \showarticletitle{{Measurement of Diversity}}.
\newblock \bibinfo{journal}{\emph{Nature}} \bibinfo{volume}{163}, \bibinfo{number}{688} (\bibinfo{year}{1949}), \bibinfo{pages}{688}.
\newblock
\urldef\tempurl%
\url{https://doi.org/10.1038/163688a0}
\showDOI{\tempurl}


\bibitem[Song and Mittal(2021)]%
        {272134}
\bibfield{author}{\bibinfo{person}{Liwei Song} {and} \bibinfo{person}{Prateek Mittal}.} \bibinfo{year}{2021}\natexlab{}.
\newblock \showarticletitle{Systematic Evaluation of Privacy Risks of Machine Learning Models}. In \bibinfo{booktitle}{\emph{30th USENIX Security Symposium (USENIX Security 21)}}. \bibinfo{publisher}{USENIX Association}, \bibinfo{pages}{2615--2632}.
\newblock
\showISBNx{978-1-939133-24-3}
\urldef\tempurl%
\url{https://www.usenix.org/conference/usenixsecurity21/presentation/song}
\showURL{%
\tempurl}


\bibitem[Sparck~Jones(1972)]%
        {sparck1972statistical}
\bibfield{author}{\bibinfo{person}{Karen Sparck~Jones}.} \bibinfo{year}{1972}\natexlab{}.
\newblock \showarticletitle{A statistical interpretation of term specificity and its application in retrieval}.
\newblock \bibinfo{journal}{\emph{J. of Documentation}} \bibinfo{volume}{28}, \bibinfo{number}{1} (\bibinfo{year}{1972}), \bibinfo{pages}{11--21}.
\newblock


\bibitem[Taal et~al\mbox{.}(2010)]%
        {5495701}
\bibfield{author}{\bibinfo{person}{Cees~H. Taal}, \bibinfo{person}{Richard~C. Hendriks}, \bibinfo{person}{Richard Heusdens}, {and} \bibinfo{person}{Jesper Jensen}.} \bibinfo{year}{2010}\natexlab{}.
\newblock \showarticletitle{A short-time objective intelligibility measure for time-frequency weighted noisy speech}. In \bibinfo{booktitle}{\emph{2010 IEEE Int. Conf. on Acoustics, Speech and Signal Processing}}. \bibinfo{pages}{4214--4217}.
\newblock
\urldef\tempurl%
\url{https://doi.org/10.1109/ICASSP.2010.5495701}
\showDOI{\tempurl}


\bibitem[Templin(1957)]%
        {templin1957certain}
\bibfield{author}{\bibinfo{person}{Maxine Templin}.} \bibinfo{year}{1957}\natexlab{}.
\newblock \bibinfo{booktitle}{\emph{Certain Language Skills in Children}}.
\newblock \bibinfo{publisher}{University of Minnesota Press}, \bibinfo{address}{Minneapolis}.
\newblock


\bibitem[Thung and Raveendran(2009)]%
        {5412098}
\bibfield{author}{\bibinfo{person}{Kim-Han Thung} {and} \bibinfo{person}{Paramesran Raveendran}.} \bibinfo{year}{2009}\natexlab{}.
\newblock \showarticletitle{A survey of image quality measures}. In \bibinfo{booktitle}{\emph{2009 International Conference for Technical Postgraduates (TECHPOS)}}. \bibinfo{pages}{1--4}.
\newblock
\urldef\tempurl%
\url{https://doi.org/10.1109/TECHPOS.2009.5412098}
\showDOI{\tempurl}


\bibitem[Vatsalan~et al.(2022)]%
        {https://doi.org/10.1111/bjet.13223}
\bibfield{author}{\bibinfo{person}{Dinusha Vatsalan~et al.}} \bibinfo{year}{2022}\natexlab{}.
\newblock \showarticletitle{Privacy risk quantification in education data using Markov model}.
\newblock \bibinfo{journal}{\emph{British Journal of Educational Technology}} \bibinfo{volume}{53}, \bibinfo{number}{4} (\bibinfo{year}{2022}), \bibinfo{pages}{804--821}.
\newblock
\urldef\tempurl%
\url{https://doi.org/10.1111/bjet.13223}
\showDOI{\tempurl}
\showeprint{https://bera-journals.onlinelibrary.wiley.com/doi/pdf/10.1111/bjet.13223}


\bibitem[Vo et~al\mbox{.}(2024)]%
        {vo2024explainabilitymachinelearningmodels}
\bibfield{author}{\bibinfo{person}{Tuan~L. Vo}, \bibinfo{person}{Thu Nguyen}, \bibinfo{person}{Hugo~L. Hammer}, \bibinfo{person}{Michael~A. Riegler}, {and} \bibinfo{person}{Pal Halvorsen}.} \bibinfo{year}{2024}\natexlab{}.
\newblock \bibinfo{title}{Explainability of Machine Learning Models under Missing Data}.
\newblock
\newblock
\showeprint[arxiv]{2407.00411}~[cs.LG]
\urldef\tempurl%
\url{https://arxiv.org/abs/2407.00411}
\showURL{%
\tempurl}


\bibitem[Wagner and Eckhoff(2018)]%
        {10.1145/3168389}
\bibfield{author}{\bibinfo{person}{Isabel Wagner} {and} \bibinfo{person}{David Eckhoff}.} \bibinfo{year}{2018}\natexlab{}.
\newblock \showarticletitle{Technical Privacy Metrics: A Systematic Survey}.
\newblock \bibinfo{journal}{\emph{ACM Comput. Surv.}} \bibinfo{volume}{51}, \bibinfo{number}{3}, Article \bibinfo{articleno}{57} (\bibinfo{date}{jun} \bibinfo{year}{2018}), \bibinfo{numpages}{38}~pages.
\newblock
\showISSN{0360-0300}
\urldef\tempurl%
\url{https://doi.org/10.1145/3168389}
\showDOI{\tempurl}


\bibitem[Wang and Jia(2023)]%
        {wang2023data}
\bibfield{author}{\bibinfo{person}{Jiachen~T. Wang} {and} \bibinfo{person}{Ruoxi Jia}.} \bibinfo{year}{2023}\natexlab{}.
\newblock \bibinfo{title}{Data Banzhaf: A Robust Data Valuation Framework for Machine Learning}.
\newblock
\newblock
\showeprint[arxiv]{2205.15466}~[cs.LG]


\bibitem[Wang and Bovik(2002)]%
        {995823}
\bibfield{author}{\bibinfo{person}{Zhou Wang} {and} \bibinfo{person}{A.C. Bovik}.} \bibinfo{year}{2002}\natexlab{}.
\newblock \showarticletitle{A universal image quality index}.
\newblock \bibinfo{journal}{\emph{IEEE Signal Processing Letters}} \bibinfo{volume}{9}, \bibinfo{number}{3} (\bibinfo{year}{2002}), \bibinfo{pages}{81--84}.
\newblock
\urldef\tempurl%
\url{https://doi.org/10.1109/97.995823}
\showDOI{\tempurl}


\bibitem[Wang et~al\mbox{.}(2004)]%
        {1284395}
\bibfield{author}{\bibinfo{person}{Zhou Wang}, \bibinfo{person}{A.C. Bovik}, \bibinfo{person}{H.R. Sheikh}, {and} \bibinfo{person}{E.P. Simoncelli}.} \bibinfo{year}{2004}\natexlab{}.
\newblock \showarticletitle{Image quality assessment: from error visibility to structural similarity}.
\newblock \bibinfo{journal}{\emph{IEEE Transactions on Image Processing}} \bibinfo{volume}{13}, \bibinfo{number}{4} (\bibinfo{year}{2004}), \bibinfo{pages}{600--612}.
\newblock
\urldef\tempurl%
\url{https://doi.org/10.1109/TIP.2003.819861}
\showDOI{\tempurl}


\bibitem[Wang et~al\mbox{.}(2003)]%
        {1292216}
\bibfield{author}{\bibinfo{person}{Z. Wang}, \bibinfo{person}{E.P. Simoncelli}, {and} \bibinfo{person}{A.C. Bovik}.} \bibinfo{year}{2003}\natexlab{}.
\newblock \showarticletitle{Multiscale structural similarity for image quality assessment}. In \bibinfo{booktitle}{\emph{The Thrity-Seventh Asilomar Conference on Signals, Systems \& Computers, 2003}}, Vol.~\bibinfo{volume}{2}. \bibinfo{pages}{1398--1402 Vol.2}.
\newblock
\urldef\tempurl%
\url{https://doi.org/10.1109/ACSSC.2003.1292216}
\showDOI{\tempurl}


\bibitem[Waterman et~al\mbox{.}(1976)]%
        {Waterman1976}
\bibfield{author}{\bibinfo{person}{M.S. Waterman}, \bibinfo{person}{T.F. Smith}, {and} \bibinfo{person}{W.A. Beyer}.} \bibinfo{year}{1976}\natexlab{}.
\newblock \showarticletitle{Some Biological Sequence Metrics}.
\newblock \bibinfo{journal}{\emph{Advances in Math.}} \bibinfo{volume}{20}, \bibinfo{number}{4} (\bibinfo{year}{1976}), \bibinfo{pages}{367--387}.
\newblock


\bibitem[Wilkinson et~al\mbox{.}(2018)]%
        {Wilkinson_Sansone_Schultes_Doorn_Bonino_da_Silva_Santos_Dumontier_2018}
\bibfield{author}{\bibinfo{person}{Mark~D. Wilkinson}, \bibinfo{person}{Susanna-Assunta Sansone}, \bibinfo{person}{Erik Schultes}, \bibinfo{person}{Peter Doorn}, \bibinfo{person}{Luiz~Olavo Bonino~da Silva~Santos}, {and} \bibinfo{person}{Michel Dumontier}.} \bibinfo{year}{2018}\natexlab{}.
\newblock \bibinfo{title}{A design framework and exemplar metrics for fairness}.
\newblock
\newblock
\urldef\tempurl%
\url{https://www.nature.com/articles/sdata2018118}
\showURL{%
\tempurl}


\bibitem[Woodie(2020)]%
        {website:datanami}
\bibfield{author}{\bibinfo{person}{Alex Woodie}.} \bibinfo{year}{2020}\natexlab{}.
\newblock \bibinfo{title}{{Data Prep Still Dominates Data Scientists’ Time, Survey Finds}}.
\newblock \bibinfo{howpublished}{\url{https://www.datanami.com/2020/07/06/data-prep-still-dominates-data-scientists-time-survey-finds/}}.
\newblock
\newblock
\shownote{Accessed on May 15, 2023}.


\bibitem[Xue et~al\mbox{.}(2014)]%
        {Xue_2014}
\bibfield{author}{\bibinfo{person}{Wufeng Xue}, \bibinfo{person}{Lei Zhang}, \bibinfo{person}{Xuanqin Mou}, {and} \bibinfo{person}{Alan~C. Bovik}.} \bibinfo{year}{2014}\natexlab{}.
\newblock \showarticletitle{Gradient Magnitude Similarity Deviation: A Highly Efficient Perceptual Image Quality Index}.
\newblock \bibinfo{journal}{\emph{IEEE Transactions on Image Processing}} \bibinfo{volume}{23}, \bibinfo{number}{2} (\bibinfo{date}{Feb.} \bibinfo{year}{2014}), \bibinfo{pages}{684–695}.
\newblock
\showISSN{1941-0042}
\urldef\tempurl%
\url{https://doi.org/10.1109/tip.2013.2293423}
\showDOI{\tempurl}


\bibitem[Yalaoui and Boukhedouma(2021)]%
        {9678209}
\bibfield{author}{\bibinfo{person}{Mehdi Yalaoui} {and} \bibinfo{person}{Saida Boukhedouma}.} \bibinfo{year}{2021}\natexlab{}.
\newblock \showarticletitle{A survey on data quality: principles, taxonomies and comparison of approaches}. In \bibinfo{booktitle}{\emph{2021 International Conference on Information Systems and Advanced Technologies (ICISAT)}}. \bibinfo{pages}{1--9}.
\newblock
\urldef\tempurl%
\url{https://doi.org/10.1109/ICISAT54145.2021.9678209}
\showDOI{\tempurl}


\bibitem[Zhang et~al\mbox{.}(2011)]%
        {5705575}
\bibfield{author}{\bibinfo{person}{Lin Zhang}, \bibinfo{person}{Lei Zhang}, \bibinfo{person}{Xuanqin Mou}, {and} \bibinfo{person}{David Zhang}.} \bibinfo{year}{2011}\natexlab{}.
\newblock \showarticletitle{FSIM: A Feature Similarity Index for Image Quality Assessment}.
\newblock \bibinfo{journal}{\emph{IEEE Transactions on Image Processing}} \bibinfo{volume}{20}, \bibinfo{number}{8} (\bibinfo{year}{2011}), \bibinfo{pages}{2378--2386}.
\newblock
\urldef\tempurl%
\url{https://doi.org/10.1109/TIP.2011.2109730}
\showDOI{\tempurl}


\bibitem[Zhao and Liu(2007)]%
        {zhao2007spectral}
\bibfield{author}{\bibinfo{person}{Zheng Zhao} {and} \bibinfo{person}{Huan Liu}.} \bibinfo{year}{2007}\natexlab{}.
\newblock \showarticletitle{{Spectral Feature Selection for Supervised and Unsupervised Learning}}. In \bibinfo{booktitle}{\emph{ICML}}. \bibinfo{pages}{1151--1157}.
\newblock


\bibitem[Zhu et~al\mbox{.}(2018)]%
        {ZHU201836}
\bibfield{author}{\bibinfo{person}{Rui Zhu}, \bibinfo{person}{Ziyu Wang}, \bibinfo{person}{Zhanyu Ma}, \bibinfo{person}{Guijin Wang}, {and} \bibinfo{person}{Jing-Hao Xue}.} \bibinfo{year}{2018}\natexlab{}.
\newblock \showarticletitle{LRID: A new metric of multi-class imbalance degree based on likelihood-ratio test}.
\newblock \bibinfo{journal}{\emph{Pattern Recognition Letters}}  \bibinfo{volume}{116} (\bibinfo{year}{2018}), \bibinfo{pages}{36--42}.
\newblock
\showISSN{0167-8655}
\urldef\tempurl%
\url{https://doi.org/10.1016/j.patrec.2018.09.012}
\showDOI{\tempurl}


\end{thebibliography}

\end{document}